\documentclass[11pt]{article}

\usepackage{arxiv}
\usepackage[utf8]{inputenc}
\usepackage[T1]{fontenc}
\usepackage[numbers,sort&compress]{natbib}
\usepackage{graphicx}
\usepackage{amsmath}
\usepackage[defaultsups=false]{newtxtext}
\usepackage{newtxmath}
\makeatletter
\DeclareRobustCommand*\textsuperscript[1]{\m@th\ensuremath{^{\mbox{\fontsize\sf@size\z@\selectfont#1}}}}
\makeatother

\makeatletter
\renewcommand{\@toptitlebar}{%
  \hrule height 4\p@
  \vskip 0.25in \vskip -\parskip}
\renewcommand{\@bottomtitlebar}{%
  \vskip 0.29in \vskip -\parskip
  \hrule height 1\p@
  \vskip 0.09in}
\renewcommand{\@maketitle}{%
  \vbox{%
    \hsize\textwidth
    \linewidth\hsize
    \vskip 0.1in
    \@toptitlebar
    \centering
    {\LARGE\bfseries \@title\par}
    \@bottomtitlebar
    \vskip 0.06in
    \def\And{\end{tabular}\hfil\linebreak[0]\hfil\begin{tabular}[t]{c}\bf\rule{\z@}{24\p@}\ignorespaces}
    \def\AND{\end{tabular}\hfil\linebreak[4]\hfil\begin{tabular}[t]{c}\bf\rule{\z@}{24\p@}\ignorespaces}
    \begin{tabular}[t]{c}\bf\rule{\z@}{24\p@}\@author\end{tabular}%
    \vskip 0.28in \@minus 0.1in
  }
}
\makeatother
\usepackage{array}
\usepackage{multirow}
\usepackage{makecell}
\usepackage{booktabs}
\usepackage{float}
\usepackage{needspace}
\usepackage{xcolor}
\definecolor{ink}{HTML}{16283F}
\definecolor{accent}{HTML}{15588C}
\definecolor{blockbg}{HTML}{F5F6F8}
\usepackage[ruled,vlined]{algorithm2e}
\usepackage{listings}
\usepackage{microtype}
\usepackage{url}
\usepackage{hyperref}
\usepackage{titlesec}
\titleformat{\section}{\color{ink}\Large\bfseries}{\thesection}{0.85em}{}
\titleformat{\subsection}{\color{ink}\large\bfseries}{\thesubsection}{0.85em}{}
\titleformat{\subsubsection}{\color{ink}\normalsize\bfseries}{\thesubsubsection}{0.85em}{}
\titlespacing*{\section}{0pt}{2.4ex plus .6ex minus .2ex}{1.3ex plus .3ex}
\titlespacing*{\subsection}{0pt}{2.0ex plus .5ex minus .2ex}{0.9ex plus .2ex}
\titlespacing*{\subsubsection}{0pt}{1.7ex plus .4ex minus .2ex}{0.6ex plus .2ex}
\usepackage[font=small,labelfont={bf,color=ink},skip=8pt]{caption}
\renewcommand{\arraystretch}{1.18}
\SetAlCapFnt{\color{ink}\small\bfseries}
\SetAlCapNameFnt{\small}

\title{Evaluating Rollout-Verified Reinforcement Fine-Tuning of a Reasoning Model for Multi-Zone Variable-Air-Volume Control}

\renewcommand{\headeright}{}

\makeatletter
\def\keywordname{{\color{ink}\bfseries\scshape Keywords}}
\def\keywords#1{\par\addvspace\medskipamount{\small\rightskip=0pt plus1cm
\def\and{\ifhmode\unskip\nobreak\fi\ {\color{accent}\textbullet}\ }%
\noindent\keywordname\quad\ignorespaces#1\par}}
\makeatother
\renewcommand{\undertitle}{}
\renewcommand{\shorttitle}{Rollout-Verified RFT for Multi-Zone VAV Control}

\author{%
  Takumi Shioda\thanks{Corresponding author. ORCID: \href{https://orcid.org/0009-0007-7335-4394}{0009-0007-7335-4394}.} \\
  The University of Tokyo \\
  \texttt{tkm-0211@iis.u-tokyo.ac.jp} \\
  \And
  Kohei Terashima \\
  Tokyo University of Science \\
  \texttt{k.terashima@rs.tus.ac.jp} \\
  \And
  Tatsuo Nagai \\
  Tokyo University of Science \\
  \texttt{nagai@rs.tus.ac.jp} \\
}

\date{}

\hypersetup{
  colorlinks=true,
  linkcolor=accent,
  citecolor=accent,
  urlcolor=accent,
  pdftitle={Evaluating Rollout-Verified Reinforcement Fine-Tuning of a Reasoning Model for Multi-Zone Variable-Air-Volume Control},
  pdfauthor={Takumi Shioda, Kohei Terashima, Tatsuo Nagai},
  pdfkeywords={reasoning models, reinforcement fine-tuning, verifiable rewards, HVAC control, variable air volume, building emulator, world models}
}

\lstdefinestyle{promptblock}{
  basicstyle=\ttfamily\footnotesize,
  breaklines=true,
  breakatwhitespace=false,
  breakindent=0pt,
  breakautoindent=false,
  columns=fullflexible,
  keepspaces=true,
  showstringspaces=false,
  backgroundcolor=\color{blockbg},
  frame=leftline,
  framerule=1.5pt,
  rulecolor=\color{accent},
  xleftmargin=12pt,
  framexleftmargin=8pt,
  framextopmargin=4pt,
  framexbottommargin=4pt,
  aboveskip=9pt,
  belowskip=9pt
}

\newcommand{\promptrole}[1]{\par\Needspace*{6\baselineskip}\noindent\textbf{#1}\par\nopagebreak}

\newcommand{\castblcaption}{}
\lstdefinestyle{promptblockcompact}{style=promptblock,basicstyle=\ttfamily\scriptsize,framextopmargin=3pt,framexbottommargin=3pt,aboveskip=6pt,belowskip=6pt}

\begin{document}
\maketitle

\begin{abstract}
Heating, ventilation, and air-conditioning systems account for a large share of building energy use. High-performance control often requires building-specific modeling or training, limiting scalability. We investigate whether a frontier reasoning model---a large language model trained to use additional inference-time computation---can coordinate multi-zone variable-air-volume control without building-specific training, and whether rollout-verified reinforcement fine-tuning can improve a smaller, locally deployable model. Five controllers were evaluated over five summer days in a physics-based four-zone emulator. Across three runs, GPT-5 used 6.8--7.4\% less electricity than a Guideline~36-based baseline, but used less outdoor air and had lower CO$_2$ compliance. GPT-5 provided a capability reference for fine-tuning the smaller model. A trained reinforcement learning controller used 4.5\% less electricity than the baseline while improving temperature and CO$_2$ compliance. To assess this controller's critic as a lower-cost verifier, we compared its estimates with deterministic rollout returns. Across 470 candidates, their correlation was 0.9995, yet the critic selected the best action in only 7 of 30 states. High overall correlation did not ensure reliable ranking at the same state. We fine-tuned \texttt{gpt-oss-20b} using rollout returns as rewards. One 200-step run produced no sustained improvement in the returns of sampled actions and no improvement in closed-loop performance. Analysis of the model's reasoning traces and quantitative tests showed that its next-state predictions of temperature and CO$_2$ were unreliable both before and after fine-tuning. Because next-state prediction is important for decision-making, these findings motivate testing whether supervised fine-tuning on next-state prediction improves subsequent rollout-verified reinforcement fine-tuning.
\end{abstract}


\section{Introduction}\label{sec:introduction}
Buildings account for about 30\% of global final energy consumption \cite{iea2026_buildings}. Heating, ventilation, and air-conditioning (HVAC) systems are a major contributor to this demand \cite{jung2019_human_hvac_review}. Reducing their energy use while maintaining thermal comfort and indoor air quality is therefore an important control objective. A multi-zone variable-air-volume (VAV) system serves several rooms, or \emph{zones}, from one central air-handling unit. The unit mixes outdoor and return air, cools the mixture, and pushes it through a duct network. A terminal damper at each zone varies the amount of supply air delivered to that zone; this variable airflow gives the system its name. The control problem is to coordinate those dampers with the outdoor-air damper, cooling-coil valve, and supply fan as weather and occupancy change.

Increasing airflow to a warm zone may cool it, but it also changes duct pressure, fan power, the flow available to the other zones, and the load on the cooling coil. VAV systems can reduce fan energy by matching supply airflow to zone loads, but the same airflow determines both thermal conditioning and ventilation \cite{shim2014_fan_control_vav}. Outdoor air dilutes occupant-generated CO$_2$, which is used here as an indicator of ventilation adequacy, but hot and humid outdoor air also requires additional cooling and dehumidification. Because outdoor-air requirements vary with occupancy, contaminant sources, and their distribution across zones \cite{lu2022_co2_dcv_critical_review}, a single ventilation command can over-ventilate some zones while the most occupied, least supplied \emph{critical zone} approaches or exceeds its CO$_2$ limit \cite{xu2007_adaptive_dcv}. Recent distributed VAV control therefore optimizes temperature, CO$_2$, and energy together under uneven occupancy \cite{ref:shi2025_hybrid_mas}. An action that helps one zone or device can worsen another.

The air and water systems are also coupled. The same sensible cooling load can be met with more air at a moderate supply temperature or less air at a lower supply temperature. The first option tends to increase fan power; the second requires colder water or more water flow and can increase pump and chiller power. The air handler and cooling plant must therefore be controlled as one energy system \cite{ahn2001_chw_quadratic_control}. In the system studied here, fan speed, cooling-coil valve position, outdoor-air intake, and four zone dampers jointly determine temperatures and flows. The controller does not directly set a zone temperature: it changes equipment commands, the physical system evolves, and temperature and CO$_2$ respond with delay.

Conventional VAV systems handle these interactions with local proportional--integral (PI) feedback loops and higher-level reset rules. A PI loop reacts to both the current error and its accumulated history. A reset rule changes a shared setpoint, such as duct pressure, when many local loops request more capacity. These controllers are effective and interpretable. In multiloop HVAC systems, PI-loop interactions make controller gains interdependent \cite{jette1998_pi_dual_duct}, while manual tuning remains labor-intensive \cite{fiducioso2019_safe_pid_tuning}. Nonlinearity and changing operating conditions further complicate system-level tuning \cite{behrooz2018_fcm_hvac_review}; in one detailed study of VAV, variable-water-volume, and CO$_2$ control, settings tuned for a high-load month gave worse control in other months \cite{yamamoto2021_shase_vav_vwv_co2_part1}. ASHRAE Guideline~36 standardizes high-performance control sequences for energy-efficient and stable operation \cite{ref:ashrae_g36_2024}. It is a strong rule-based baseline, but its coordination logic is fixed at design time. Good local tracking does not by itself guarantee minimum system-wide electricity use.

Model predictive control (MPC) and reinforcement learning (RL) provide two optimization-based alternatives. MPC repeatedly solves a finite-horizon optimization problem over predicted building responses and applies the first command. It needs a control model, state estimation, an optimizer, and integration with the building automation system \cite{ref:drgona2020_mpc_buildings}, and field deployment remains uncommon despite extensive research \cite{ref:saloux2025_mpc_field_implementations}. RL instead learns a \emph{policy}, a mapping from observations to actions, by interacting with an environment and maximizing cumulative reward. It can learn in a building emulator without solving an optimization problem online, but reward design, training data, safety, robustness, and generalization remain practical concerns \cite{ref:wang2020_rl_building_controls}. Recent work has compared deep RL with Guideline~36 in a multi-zone model with central air and water loops \cite{ref:savino2025_drl_g36}.

Even when MPC and RL perform well on energy and comfort metrics, three barriers still limit routine deployment. First, the control knowledge they produce is building-specific: a model or policy developed for one system is costly to obtain and difficult to transfer to another controller or building. Second, a neural policy is opaque at its interface: it can produce commands that perform well on test metrics while giving an operator little basis for judging why it acted as it did, whether the response is physically plausible, or how it will behave outside the tested range. Concerns about trial-and-error actions in a university heating system motivated action masking with engineering constraints \cite{ref:heidari2025_trustworthy_heating}, and other HVAC work extracts inspectable decision-tree policies from black-box controllers \cite{ref:an2024_verifiable_hvac}. Third, the experts who could bridge these gaps are scarce. NIST identifies workforce shortages as a barrier to HVAC maintenance and repair \cite{ref:nist_ai_optimized_controls}. A U.S. Department of Energy roundtable highlights the need for skills spanning HVAC, controls, and information technology \cite{ref:doe_emcs_workforce_roundtable}. A candidate control technology should therefore be judged not only on energy and comfort results but also on whether it improves transferability and inspectability while reducing reliance on scarce specialists.

\subsection{Large language models and reasoning models}\label{subsec:background}
A large language model (LLM) learns patterns from large collections of text by predicting the next token---a word, part of a word, or symbol---from the text that comes before it. Modern LLMs are usually based on the Transformer architecture \cite{vaswani2017_attention} and acquire broad linguistic and task knowledge through large-scale pretraining \cite{brown2020_language_models}. At inference time, a user supplies a \emph{prompt}; the model then generates an output token by token. Structured numerical observations can be written into that prompt, and the generated text can encode an action. An LLM is nevertheless not, in itself, a physical simulator or a control algorithm. Pretraining does not guarantee that it understands a particular building, respects actuator limits, or predicts the consequences of a command correctly.

In this paper, a \emph{reasoning model} means an LLM that has been post-trained and configured to allocate additional inference-time computation to multi-step problem solving before returning its final answer. It remains an LLM; ``reasoning model'' describes its training and inference behavior, not a wholly separate model family. Reinforcement learning is one way to elicit and strengthen this extended deliberation \cite{ref:deepseekr1_2025}. The intended advantage for control is that the model can compare interacting constraints, anticipate delayed effects, and revise a candidate command before producing the final structured action.

General pretraining may allow an LLM to reuse control-relevant patterns across sites, while its text interface can pair actions with human-readable explanations \cite{ref:zhang2024_llm_interpretable_control}. Together, these features could reduce repeated site-specific engineering.

Used as a controller, the LLM implements the same abstract loop as any policy. At decision time $k$, building energy management system (BEMS) measurements and operating constraints are serialized as text; the LLM maps that context to a JSON action; a parser checks the fields and actuator bounds; and the emulator advances to the next observation. Repeating this loop turns a text generator into a closed-loop controller. Unlike a fixed rule set, the model can use heterogeneous text as context without every relationship being hand-coded. Unlike MPC, it does not solve a numerical optimization problem online. Unlike a building-specific RL policy, it begins with knowledge acquired during general pretraining.

Our comparison covers three sources of control knowledge: fixed engineering rules in Guideline~36, experience learned by RL in an emulator, and knowledge acquired during language-model pretraining and post-training. We evaluate an \emph{open-weight} model, meaning that its parameters are available for local inference and adaptation. Local use can keep BEMS data on site, avoid a hosted API during operation, and retain outputs for audit. The controllers also differ in what they expose for diagnosis: a conventional neural policy does not natively return a readable account alongside its action; the hosted GPT-5 interface used here exposes a reasoning summary, whereas the locally served open-weight model allows its generated analysis channel to be logged in full. GPT-5 is included as a \emph{capability reference}: its unadapted performance probes the potential of a frontier LLM for VAV control and measures what is currently attainable without building-specific weight updates; it is not the model to be adapted or deployed locally.

\subsection{Rollout-verified reinforcement fine-tuning}\label{subsec:transfer}
Table~\ref{tab:concept-map} summarizes the terms used in the rest of this paper. The key distinction is between a \emph{policy}, which generates actions, and a \emph{verifier}, which scores generated actions. Reinforcement fine-tuning (RFT) updates the former using feedback from the latter. In this study, a twin-delayed deep deterministic policy gradient (TD3) policy serves as the teacher.

\begin{table}[H]
\centering
\caption{Terms used in the controller comparison and rollout-verified fine-tuning, with definitions and roles for the policy, critic, verifier, fine-tuning procedure, and transition model.}
\label{tab:concept-map}
\footnotesize
\setlength{\tabcolsep}{4pt}
\begin{tabular*}{\textwidth}{@{\extracolsep{\fill}}p{0.18\textwidth}p{0.36\textwidth}p{0.39\textwidth}@{}}
\toprule
Term & Plain-language meaning & Role in this study \\
\midrule
Policy / actor & A mapping from the current observation to an action. & TD3 is the teacher policy, \texttt{gpt-oss-20b} the student policy, and GPT-5 the capability reference. \\
Critic (action-value estimator) & A model that predicts the cumulative future reward after taking an action and then following a policy. & The TD3 critic is audited as a possible lower-cost verifier. \\
Verifier & A procedure that checks or scores a model output without prescribing its wording. & Scores sampled LLM actions and supplies the RFT feedback. \\
RFT & Reinforcement fine-tuning: updating an LLM so that higher-reward sampled outputs become more likely. & Adapts the open-weight student from verifier feedback. \\
Transition model & A predictor of the next state from the current state and action. & Evaluated separately to test whether the model knows the local physical effects needed to propose actions. \\
\bottomrule
\end{tabular*}
\end{table}

RFT is a form of post-training. Supervised fine-tuning (SFT) would provide a target output and train the model to imitate it. RFT instead samples several outputs, assigns each a scalar reward, and changes the policy so that higher-reward outputs become more probable. When the reward is computed by an automatic, objective check, the setting is often called reinforcement learning with verifiable rewards (RLVR). Prominent results include DeepSeekMath, which introduced group-relative policy optimization for mathematical reasoning \cite{deepseekmath2024}, and DeepSeek-R1, which used reinforcement learning with verifiable rewards for reasoning tasks, including mathematics and coding \cite{ref:deepseekr1_2025}. Continuous physical control offers no comparably direct answer checker: the quality of a command depends on state-dependent and delayed consequences that must be assessed through a predictive model, simulation rollout, or measured outcome. In this study, the verifier therefore scores each sampled command from its simulated physical consequences. The LLM is not trained to imitate TD3's actions or copy its network weights; instead, the experiment asks whether these teacher-derived scalar scores can improve the LLM policy.

Recent studies have applied LLMs to several energy-system tasks \cite{ref:mirshekali2025_llm_energy_systems_review}. One role is to estimate information used by a controller. Qaisar et al. used few-shot chain-of-thought prompting to infer office occupancy from sensor data without model fine-tuning \cite{qaisar2026_llm_occupancy}. Related work on occupant-centric control shows how such information can support HVAC operation: measured occupancy was used to compare operating schedules and their effects on energy and comfort \cite{qaisar2025_occupancy_centric_control}, while OCC-Mamba predicted occupancy numbers to adjust terminal operation and minimum airflow \cite{qaisar2026_occmamba}. Both control evaluations used EnergyPlus simulations.

A second role is to select control actions directly. Examples include prompting GPT-4 with demonstrations and the current observation to select each action \cite{ref:song2023_pretrained_llm_industrial_control}; connecting ChatGPT to an EnergyPlus office model \cite{ref:ahn2023_chatgpt_hvac}; combining building data with foundation models in a real office \cite{ref:sawada2025_agentic_ai_dce}; grounding a five-zone EnergyPlus controller in a spatial-semantic knowledge graph \cite{ref:bhatt2026_thermollm}; and generating cooling setpoints with human-readable rationales from retrieved domain guidance \cite{ref:ko2025_darlin}. These studies evaluate models without updating their weights for the control task. Zhong et al. instead fine-tuned an LLM on historical building-operation data to generate feasible action masks for a downstream RL controller in a discrete multi-zone HVAC problem \cite{ref:zhong2026_hierarchical_llm_rl}. That approach uses supervised feasibility labels and leaves final action selection to RL. To our knowledge, no previous study has built a fine-tuning reward from a continuous building-control problem, checked that reward against direct emulator returns, and then measured what control knowledge the reward transfers.

Our previous study tested this transfer in a simpler thermal energy storage (TES) scheduling problem. Dynamic programming enumerated the future consequences of a small discrete action set and provided an exact long-horizon value for every action. Those values were converted into dense rewards. RFT used group-relative policy optimization (GRPO), which compares several outputs generated for the same prompt \cite{deepseekmath2024}, with the Dr.~GRPO objective \cite{liu2025r1zero}. With 30 training prompts, it moved an open-weight reasoning model from well below a deep RL baseline to within 0.7\% of the dynamic-programming optimum \cite{shioda2026_tes_rft}. VAV control is harder in two ways. First, its seven actuator commands are continuous, so the full action set cannot be enumerated and scored: the exhaustive dynamic programming used for TES is unavailable, and the verifier evaluates only the actions sampled by the LLM. Second, action effects are harder to predict. In the TES benchmark, an explicit additive storage balance made each state transition directly calculable; the effects of a VAV command instead propagate through coupled zone, air-loop, and water-loop dynamics over time. VAV therefore forms a boundary test of whether this action-value-based transfer approach extends beyond the discrete TES setting.

The central methodological question is therefore how to score generated actions. RFT needs a reliable ranking of actions generated for the \emph{same} state, but it can reinforce only actions that the model already samples \cite{mroueh2025rlvr}. SayCan uses a related idea: value functions for pretrained robot skills score high-level actions proposed by an LLM \cite{ichter2023_saycan}. Here, a deterministic emulator rollout verifier scores continuous actuator vectors, with TD3 serving as the teacher. The recipe is not specific to VAV: only the verifier---a trusted optimizer, a deterministic rollout, or a measured outcome---changes with the domain. Tasks such as storage scheduling and central-plant sequencing are therefore candidate extensions, and the same construction could in principle apply to district or multi-building energy systems. The open question is whether exact scores for sparsely sampled continuous actions are sufficient to improve the policy when the model may lack the action-conditioned transition knowledge---how each command changes the next state---needed to construct better candidates. Section~\ref{sec:methodology} defines the score and training protocol.

\subsection{Study objectives and contributions}\label{subsec:gap-objectives}
This study addresses two questions for artificial intelligence (AI) in energy systems: whether foundation models can coordinate energy equipment through text, and whether verified action values provide an effective training signal for a smaller model. We examine these questions in one physics-based multi-zone VAV testbed under common evaluation conditions. The first objective is to establish whether a frontier reasoning model, given only the equipment description and BEMS observations as text and no building-specific weight updates, can perform closed-loop control, and how its energy, temperature, and CO$_2$ performance compare with a Guideline~36 rule-based controller and a TD3 policy trained in the emulator. This capability evaluation determines whether the shared textual interface can support coordinated VAV control. With that prerequisite demonstrated, the study then tests whether TD3-guided, rollout-verified RFT can transfer building-specific control knowledge into a locally deployable open-weight reasoning model. GPT-5 therefore serves as the capability reference, TD3 as the teacher, and \texttt{gpt-oss-20b} as the transfer target. Establishing capability does not guarantee successful transfer: a frontier model may already contain useful physical and planning priors, while a smaller model may fail to sample actions that its verifier could reinforce.

This study makes three contributions. (i)~It shows that a learned critic can track returns almost perfectly across states yet rank candidate actions unreliably within a state, which is the comparison that group-relative RFT requires. A direct rollout verifier for continuous LLM actions provides the reference for this audit. (ii)~It compares fixed control rules (Guideline~36), a policy learned in an emulator (TD3), and LLM controllers (GPT-5 and \texttt{gpt-oss-20b}) in the same physics-based VAV emulator, establishing the capability reference for the transfer experiment. (iii)~It tracks sampled-action quality across the 200-step RFT run and, when exact scores produce no sustained improvement, formalizes the missing directional content: the transition-mediated component of $\partial Q/\partial a$ combines a preference over next states with how each actuator locally changes them. A machine-scored five-minute transition test then probes these local transition effects and finds \texttt{gpt-oss-20b} unable to express them in the direct-answer setting before and after RFT, whereas GPT-5 recovers them for three of the four tested pairs.

\section{Methodology}\label{sec:methodology}
\begin{figure}[H]
\centering
\includegraphics[width=\textwidth,pagebox=cropbox]{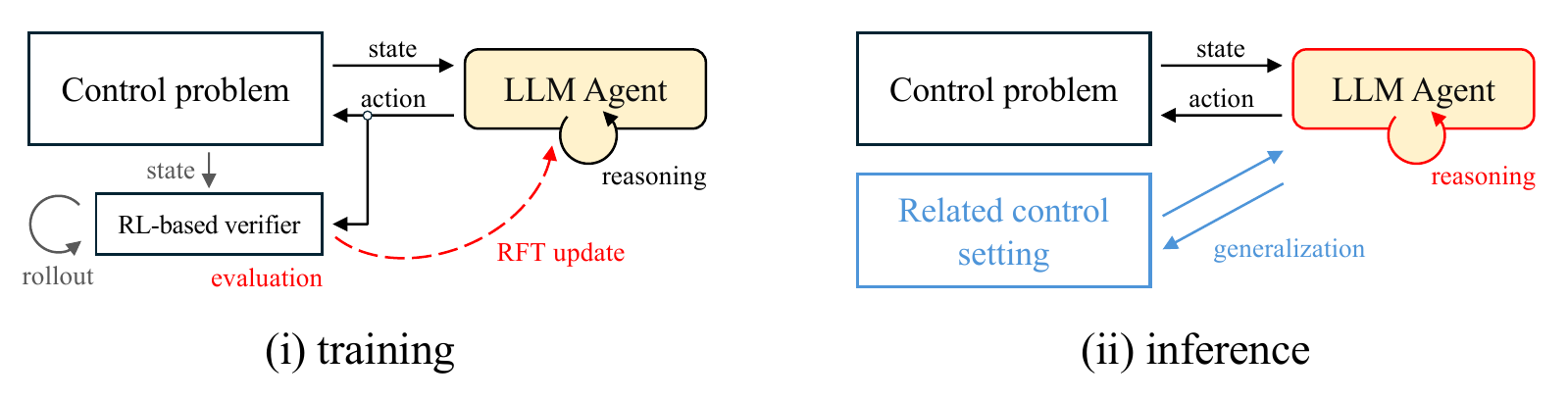}
\caption{Overview of rollout-verified RFT for VAV control. During training, each candidate LLM action is applied for one interval and TD3 controls the remaining intervals. The rollout return provides the RFT reward. During inference, TD3 and the verifier are removed, and the fine-tuned LLM acts directly. The blue path shows the intended use in a related control setting.}
\label{fig:rft_td3_framework}
\end{figure}

This section presents the two parts of the method: the common LLM control interface and the rollout-verified RFT framework. During offline training, we first train and freeze TD3 and save states from its control trajectories. For each saved state, the LLM proposes several actions. Each candidate is applied for one five-minute interval, after which TD3 controls the emulator to the end of the day. The resulting return is used as the verifier score. During closed-loop evaluation, TD3 and the verifier are removed, and the LLM acts directly from the current BEMS observation. Figure~\ref{fig:rft_td3_framework} summarizes these two phases.

Section~\ref{subsec:controller-design} defines what the LLM observes and controls. Section~\ref{subsec:verifier-selection} audits the learned critic against direct rollouts, selects the verifier, and defines its score. Section~\ref{subsec:rollout-rft} then gives the RFT update. Section~\ref{sec:experimental-setup} describes the VAV system, controller settings, and training configuration. Appendix~\ref{app:configurations} provides the remaining implementation details.

\subsection{LLM controller design}\label{subsec:controller-design}
The LLM receives BEMS observations and selects actuator commands every five minutes. It commands seven actuators: four zone dampers, one outdoor-air damper, one chilled-water valve, and the supply fan. Opening a zone damper tends to increase that zone's airflow; opening the outdoor-air damper tends to improve dilution while increasing the outdoor cooling load; opening the chilled-water valve tends to lower the supply-air temperature; and increasing fan speed raises duct pressure and airflow at an electrical cost. Because these effects interact, a useful command is a coordinated seven-dimensional vector rather than seven independent choices. The emulator's equipment models and local loops translate the normalized commands into physical flows and temperatures while enforcing equipment limits.

The LLM and TD3 receive the same types of observations. The LLM receives them as text: current zone temperatures and CO$_2$ concentrations, their latest changes, outdoor conditions and trend, the previous action, and the actuator bounds. The changes and previous action provide a short history, but the controller does not observe internal duct pressures, coil states, wall temperatures, or controller integrators. The problem is therefore \emph{partially observed}: different hidden physical states can produce the same visible measurements but respond differently to the next command. TD3 uses the same fields (Section~\ref{subsec:td3-setup}), so the two learned controller types receive aligned information even though their numerical and textual representations differ.

Each control request contains three messages. The system message defines the zones, comfort band, CO$_2$ limit, and main physical trade-offs. These include the cubic dependence of fan and pump power on speed, the energy cost of outdoor air, and the static-pressure goal of keeping at least one zone damper fully open. The developer message requires one JSON object with a step-by-step \texttt{thought\_process} and an \texttt{action} object with one numeric value per actuator. The user message provides the current observation. This separation makes the interface reproducible: the task definition is fixed, while only the state data change at each decision.

The generated text is not applied to the emulator directly. A parser first checks that all seven fields exist and are numeric. Values are then clipped to the actuator bounds, and only this checked and clipped action vector affects the emulator. The reasoning text is stored for qualitative analysis but has no control authority. Clipping prevents out-of-range commands; it does not guarantee comfort, ventilation, stability, or low energy use. Those properties must be evaluated from the resulting trajectory. Appendix~\ref{app:control-prompt} gives the full prompt.

In the capability evaluation, the model weights remain frozen. The prompt supplies the building description, actuator ranges, comfort targets, and physical trade-offs. This test therefore asks whether the model can use the description at inference time, not whether it learned the building through training. The GPT-5 results reflect the model with this prompt; we did not isolate the contribution of the engineering guidance. In the subsequent transfer experiment, RFT updates low-rank adaptation (LoRA) weights while keeping the base model fixed \cite{lora2022}. The trained adapter can then be loaded or removed without changing the base model.

\subsection{Verifier design and selection}\label{subsec:verifier-selection}
RFT needs a verifier for the LLM's continuous actions. Because the TD3 action is never supplied as a target response, the method rewards action quality rather than imitation. The first candidate was the learned TD3 critic, which predicts cumulative future reward for an observation--action pair in one network evaluation. However, it sees only the controller's partial observation and is trained on replay data near TD3 behavior; ranking same-state LLM actions may therefore require extrapolation. Related offline RL methods limit this problem by keeping actions near the data distribution \cite{ref:fujimoto2019_bcq} or learning conservative values under distribution shift \cite{ref:kumar2020_cql}.

RFT prompts are observations from deterministic TD3 evaluation trajectories. The prompts therefore remain on the teacher's state distribution, although actions sampled within the actuator limits may differ from the teacher's actions.

\begin{sloppypar}
The teacher-derived training observations were held fixed to evaluate action-value supervision without simultaneously changing the state-collection policy. Candidate actions were scored using the reference-policy action value $Q^{\pi_{\mathrm{TD3}}}(s,a)$, with each comparison starting from the same restored state under the same future disturbances.
\end{sloppypar}

We audited the critic against direct emulator returns using candidate actions from the base LLM. The audit used 30 states: ten hourly branch times from 08:20 to 17:20 on each of 21--23 July 2025, fixed before scoring. These states come from three daily TD3 trajectories, not from 30 independent days. A saved \emph{full state} $s$ contains all variables needed to restart the emulator, whereas $o(s)$ contains only the measurements available to TD3 and the LLM and forms the text prompt $x_s$. Conditional on fixed weather and occupancy, the full state, continuous action, interval reward, and deterministic transition define a finite-horizon Markov decision process (MDP) at the five-minute decision scale; because the controllers receive only $o(s)$, they face a partially observable Markov decision process (POMDP). The verifier restores the full state and scores all candidates in a group from that state, so the within-group comparison is consistent conditional on that state and does not require the observation to be Markov. TD3, which receives the same observable information as the LLM, achieved the best temperature and CO$_2$ performance in Section~\ref{subsec:control-results}. This provides empirical evidence that these observations support effective control under the evaluated conditions. Whether longer observation histories improve RFT or LLM control performance remains an open question.

The frozen base \texttt{gpt-oss-20b} model generated 16 actions per state with temperature 0.6 and medium reasoning effort. Of 480 generations, 476 were valid; the four invalid generations were excluded without replacement. Clipping and deduplication within each state yielded 470 unique candidates. Each candidate was applied for five minutes, after which the frozen deterministic TD3 actor controlled the remaining intervals until 18:00 with future disturbances fixed; returns were discounted with $\gamma=0.995$. One such restored branch---a \emph{direct rollout}---provides the reference return. All 470 branches passed the source-trajectory replay check. The critic received the 27-dimensional observation captured at the branch point and the executed action mapped to $[-1,1]$.

We report two kinds of agreement. Pooled agreement is the Pearson correlation between critic estimates and rollout returns across all states. Within-state agreement is the comparison that group-relative RFT requires. For each state, we computed the Spearman correlation and pairwise accuracy among comparable pairs, and recorded whether the critic selected the candidate with the highest rollout return. We report the mean Spearman correlation and pairwise accuracy across the 30 states and the number of states with a matching selection.

\begin{figure}[H]
\centering
\includegraphics[width=\textwidth]{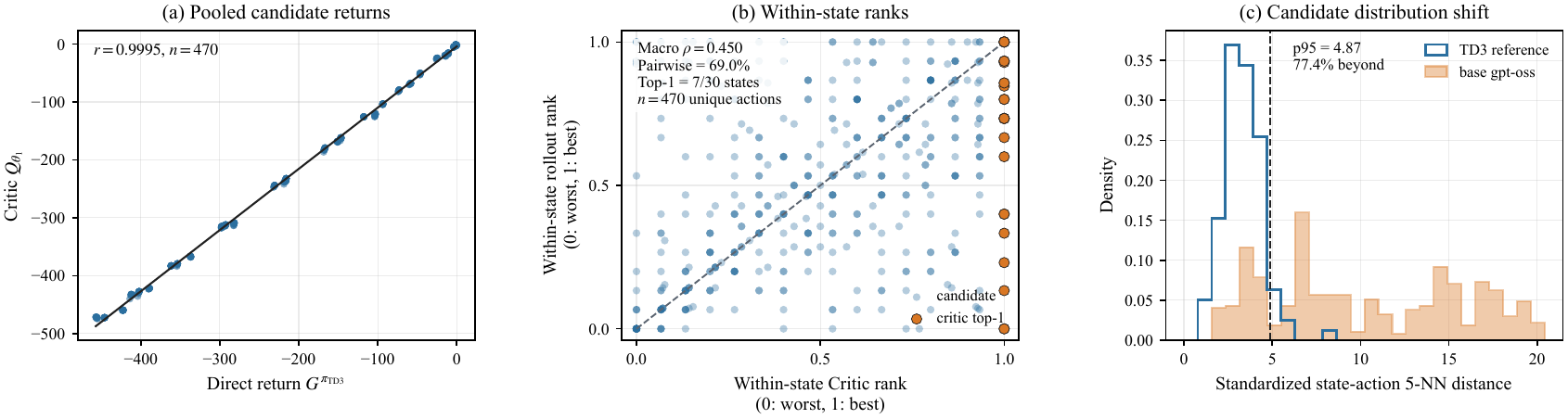}
\caption{Critic audit (30 states). (a) Pooled critic estimates and direct returns for 470 unique base-model actions at 30 TD3-trajectory states. Points cluster by state, with differences in return level between states dominating the pooled correlation. (b) Within-state ranks for the same candidates; orange points mark critic-selected actions. (c) Standardized five-nearest-neighbor distances of these state--action pairs from 100 reference TD3 state--action pairs; the dashed line marks the reference 95th percentile. All three panels use the same 470 candidates; panel (c) additionally shows the TD3 reference distribution.}
\label{fig:critic-rollout-audit}
\end{figure}

Pooled over the 470 candidates, critic estimates and rollout returns had a Pearson correlation of $r=0.9995$ (Figure~\ref{fig:critic-rollout-audit}(a)). Within states, however, the mean Spearman correlation was $0.450$, pairwise accuracy was $69.0\%$, and the critic selected the rollout-best candidate in only 7 of 30 states (Figure~\ref{fig:critic-rollout-audit}(b)). Differences in return level between states dominate the pooled correlation, but these differences play no role when candidates at one state are ranked. Selecting the critic's top candidate instead of the rollout-best candidate incurred a mean discounted-return regret of $0.3013$ (Appendix~\ref{app:critic-selection-loss}).

For the distance diagnostic, observations were normalized with the TD3 normalizer, actions were mapped to $[-1,1]$, and the combined features were standardized using the 100 reference pairs. Of the 470 candidates, 364 ($77.4\%$) exceeded the reference 95th-percentile distance of $4.87$ (Figure~\ref{fig:critic-rollout-audit}(c)), so many LLM actions required the critic to extrapolate. This diagnostic uses a finite reference set and does not establish the cause of the ranking errors.

The critic remained useful for TD3 training but did not reliably rank candidates within a state. Because group-relative RFT depends on this ranking, we rejected the critic as the RFT verifier and used direct rollout. This choice was made before RFT from a 10-state pilot audit, in which the critic selected the rollout-best candidate in 5 of 10 states. A 30-state audit conducted after the RFT experiments, with different state selection and generation settings, confirmed this result.

At a stored state $s$, direct rollout applies candidate action $a$ during interval $t$ and follows the frozen TD3 actor $\mu=\pi_{\mathrm{TD3}}$ from $t+1$ to the last decision $T$, with future disturbances $d_{t:T}$ fixed. With $r_t$ denoting the interval reward, its finite-horizon action value is
\begin{equation}\label{eq:state-q-definition}
Q_{t,d}^{\mu}(s,a)
:=
\mathbb{E}_{\mu}\!\left[
\left.\sum_{k=0}^{T-t}\gamma^k r_{t+k}\,\right|\,
s_t=s,\ a_t=a,\ d_{t:T}
\right].
\end{equation}
This is the quantity that the critic was intended to approximate. The emulator, reward calculation, disturbances, and TD3 actor are deterministic, so Eq.~\eqref{eq:state-q-definition} has no sampling variance:
\begin{equation}\label{eq:branch-return}
\begin{aligned}
Q_{t,d}^{\mu}(s,a)
&=G_{t,d}^{\mu}(s,a)
:=\sum_{k=0}^{T-t}\gamma^k r_{t+k},\\
\operatorname{Var}\!\left[G_{t,d}^{\mu}\mid s,a,d_{t:T}\right]
&=0.
\end{aligned}
\end{equation}
Classical rollout policy improvement evaluates a candidate action by simulating a base policy after that action; stochastic tasks typically require averaging multiple trajectories, whereas one rollout suffices here because each branch is deterministic \cite{tesauro1997_rollout}. When time and disturbances are clear from context, we write this return as $G^{\pi_{\mathrm{TD3}}}(s,a)$. Direct rollout removes critic approximation and extrapolation error from the assigned scores, but exactness applies only to this criterion: $Q_{t,d}^{\mu}$ need not equal $Q_{t,d}^{*}$.

TD3 continuation also gives a one-step policy-improvement interpretation. Define the TD3 baseline value as
\begin{equation}\label{eq:policy-improvement-baseline}
V_{t,d}^{\mu}(s)
:=Q_{t,d}^{\mu}\!\left(s,\mu(o(s))\right).
\end{equation}
For a candidate policy $\pi$, if
\begin{equation}\label{eq:policy-improvement-condition}
\mathbb{E}_{a\sim\pi(\cdot\mid o(s))}
\!\left[Q_{t,d}^{\mu}(s,a)\right]
\ge V_{t,d}^{\mu}(s)
\end{equation}
at every state and decision time, standard policy-improvement reasoning implies that repeatedly using $\pi$ is no worse than repeatedly using $\mu$ for the same finite-horizon problem \cite{sutton2018rl}. The rollout score is thus a baseline-policy action value, not the return under continued LLM control. This sufficient condition is formulated over full emulator states. The observation-based controller can encounter distinct full states with the same observation, and training on 30 prompts with finitely many candidate actions does not establish the condition at all states encountered during deployment. We therefore use policy improvement as the rationale for the scoring target, without claiming its guarantee for the trained LLM.

\subsection{Rollout-verified RFT procedure}\label{subsec:rollout-rft}

\begin{figure}[H]
\centering
\includegraphics[width=\textwidth,pagebox=cropbox]{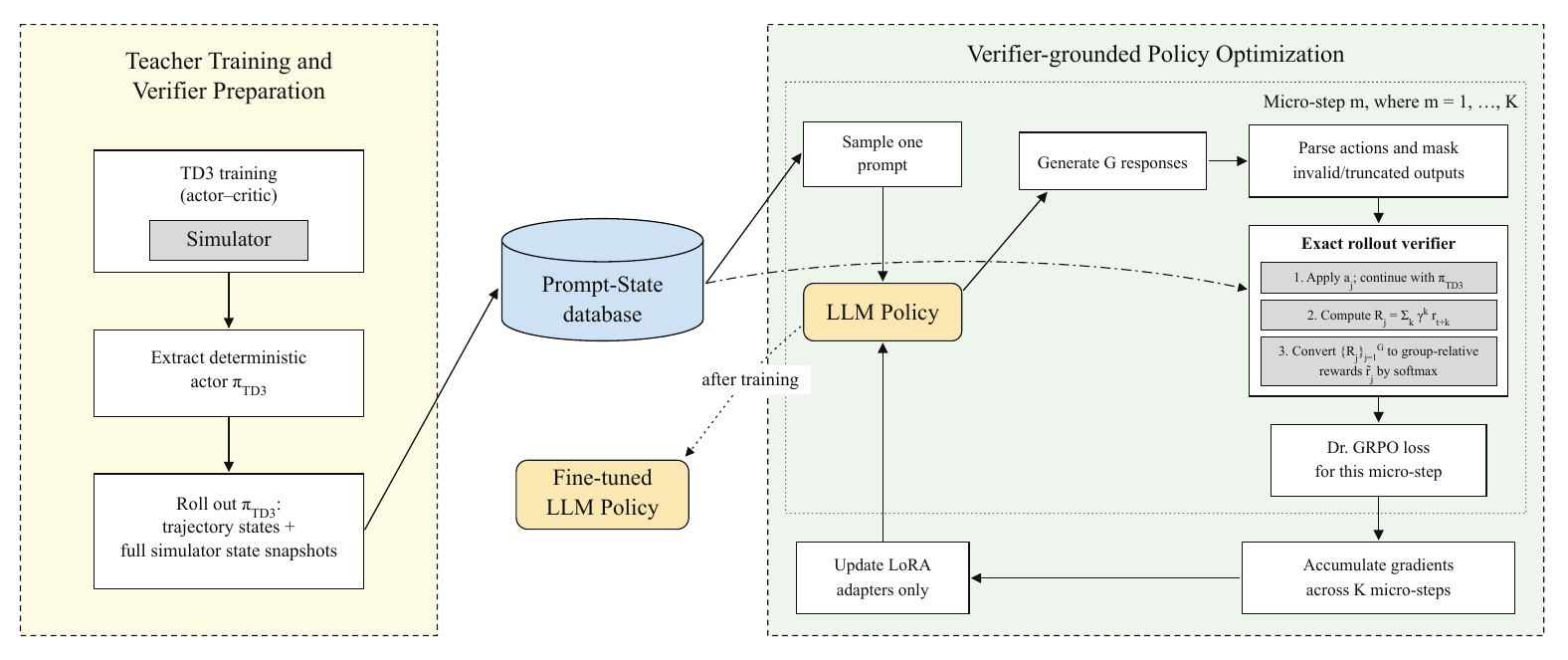}
\caption{Teacher preparation and rollout-verified RFT. The left panel shows TD3 training and the collection of prompt observations with full emulator states. In each RFT micro-step, the LLM generates $G$ responses for one prompt. Invalid responses are masked, and one restored rollout evaluates each valid action. Returns are converted to group-relative rewards, gradients are accumulated over $K$ micro-steps, and only the LoRA parameters are updated.}
\label{fig:rollout_rft_flow}
\end{figure}

With direct rollout selected, the remaining step is to turn its scores into a model update. This is an RLVR procedure because each valid sampled response receives an automatic score from its emulator rollout. For one prompt, the LLM samples a group of $G$ complete responses. The parser defined in Section~\ref{subsec:controller-design} excludes invalid responses from training; let $\mathcal{V}\subseteq\{1,\ldots,G\}$ index the valid responses, with $a^{(j)}$ denoting response $j$'s clipped action. Candidate $j$ receives the direct rollout score $R_j=G_{t,d}^{\mu}(s,a^{(j)})$. The raw returns are then converted to group-relative rewards:
\begin{equation}\label{eq:softmax-reward}
\tilde r_j
=\frac{\exp(R_j/\tau)}{\sum_{\ell\in\mathcal{V}}\exp(R_\ell/\tau)},
\qquad j\in\mathcal{V}.
\end{equation}
The return scale can change with the time of day, whereas Eq.~\eqref{eq:softmax-reward} produces nonnegative weights that sum to one within each prompt group. The reward temperature $\tau$---which is separate from the LLM sampling temperature---controls how strongly the best return is favored. RFT therefore depends mainly on the verifier's within-state ranking, not on the absolute return level. Figure~\ref{fig:rollout_rft_flow} shows the complete training loop.

We optimize the policy using the Dr.~GRPO objective over complete responses. Let $\pi_{\theta_{\mathrm{old}}}$ denote the policy that generated a response, let $u$ index its tokens, and let $L_{\max}$ be the fixed maximum completion length. The centered advantage $\hat A_j$ is positive for an above-average response and negative for a below-average response. The probability ratio $\rho_{j,u}$ measures how much the updated model changes the probability of token $u$ relative to the policy that generated it. For each valid response $j$, we use
{\allowdisplaybreaks
\begin{align}
\bar r_{\mathcal V}
&=\frac{1}{|\mathcal V|}\sum_{\ell\in\mathcal V}\tilde r_\ell,\nonumber\\
\hat A_j
&=\tilde r_j-\bar r_{\mathcal V},\nonumber\\
\rho_{j,u}(\theta)
&=\frac{\pi_\theta(y^{(j)}_u\mid x_s,y^{(j)}_{<u})}
{\pi_{\theta_{\mathrm{old}}}(y^{(j)}_u\mid x_s,y^{(j)}_{<u})},\nonumber\\
\ell_{j,u}(\theta)
&=\min\!\left\{
\rho_{j,u}(\theta)\hat A_j,\right.\nonumber\\*[-2pt]
&\qquad\left.
\operatorname{clip}\!\left(
\rho_{j,u}(\theta),1-\epsilon,1+\epsilon
\right)\hat A_j
\right\},\nonumber\\
\mathcal J_{\mathrm{DrGRPO}}(\theta)
&=\frac{1}{|\mathcal V|L_{\max}}
\sum_{j\in\mathcal V}\sum_{u=1}^{|y^{(j)}|}\ell_{j,u}(\theta).
\label{eq:drgrpo-objective}
\end{align}}
The clipped objective discourages large changes to a sampled token probability in one update. The fixed $L_{\max}$ normalization prevents short responses from receiving extra weight simply because they contain fewer tokens. Rewards are centered within each prompt group but are not divided by the group standard deviation. Only the LoRA parameters receive gradients; the base-model weights remain fixed. The optimizer therefore raises the probability of above-average responses and lowers the probability of below-average responses. Appendix~\ref{app:rft-configuration} lists the full settings, and Algorithm~\ref{alg:vav_rollout_rft} in Appendix~\ref{app:rft-algorithm} gives the complete procedure.

All candidates in one group share the same full initial state, future disturbances, and TD3 continuation. Only the first five-minute action differs, so their returns are directly comparable under Eq.~\eqref{eq:state-q-definition}. Each uncached action requires a rollout to the end of the day, which makes training expensive. After training, this cost disappears: the verifier and TD3 are removed, and the adapted LLM selects actions directly.

\section{Experimental setup}\label{sec:experimental-setup}
\subsection{VAV control problem}\label{subsec:problem-formulation}
\begin{figure}[H]
\centering
\includegraphics[width=\textwidth]{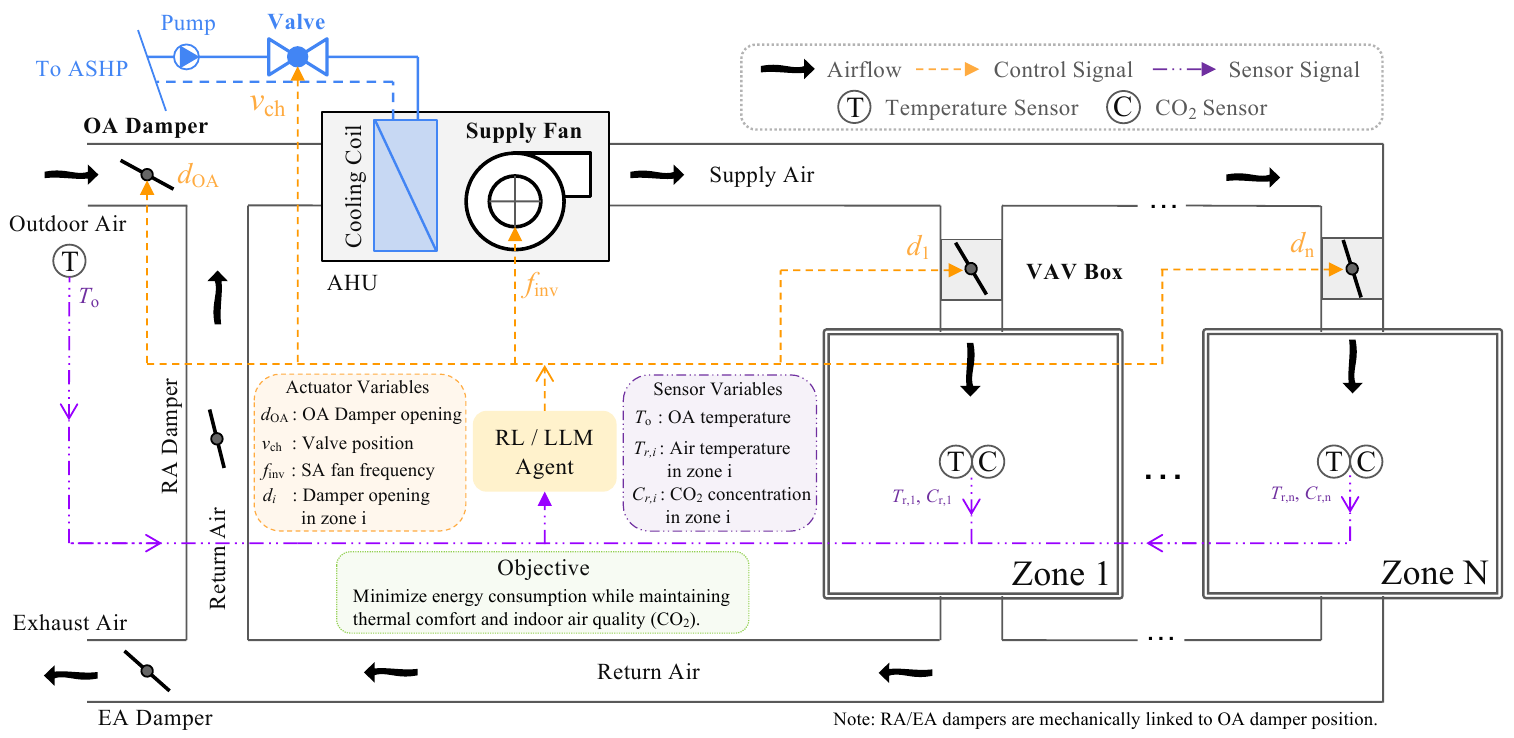}
\caption{Four-zone VAV system used in this study. The controller observes outdoor and zone conditions and sets the zone dampers $d_i$, outdoor-air damper $d_{\mathrm{OA}}$, chilled-water valve $v_{\mathrm{ch}}$, and supply-fan speed $f_{\mathrm{inv}}$. The return- and exhaust-air dampers are mechanically linked to the outdoor-air damper.}
\label{fig:vav_system_overview}
\end{figure}

The testbed represents a VAV system that serves four office zones in Tokyo. In Figure~\ref{fig:vav_system_overview}, the air path runs from left to right: outdoor air and return air from the zones first enter a mixing box. The cooling coil then removes sensible heat and moisture, the supply fan pressurizes the conditioned air, and the duct network distributes it to four VAV terminals. Each terminal damper determines its zone's share of the supply airflow. Air leaving the zones is partly exhausted and partly returned to the air-handling unit. The outdoor-air damper is mechanically linked to the return- and exhaust-air dampers so that changing ventilation also changes the recirculated fraction.

On the water side, a chiller produces chilled water, a variable-speed pump circulates it, and the cooling-coil valve regulates flow through the air-handling unit. Opening that valve generally cools and dehumidifies the supply air more strongly, but it can increase both pumping and chiller electricity. The system therefore contains two interacting transport networks: air carries cooling and ventilation to the zones, while water carries cooling from the chiller to the air stream.

The objective is to minimize total electricity use while keeping every zone comfortable and adequately ventilated during occupied hours. This is a constrained objective rather than an energy-only objective: a controller cannot claim a useful saving by simply turning down cooling or outdoor air until occupants become uncomfortable. Let $\mathcal{T}_{\mathrm{on}}$ be the set of occupied time steps, $P_t$ the total fan, chilled-water pump, and chiller power, and $a_k$ the control action. The constrained problem is
\begin{equation}\label{eq:control-problem}
\begin{aligned}
\min_{\{a_k\}} \quad & \sum_{t\in\mathcal{T}_{\mathrm{on}}} P_t\,\Delta t \\
\mathrm{s.t.}\quad
& x_{t+1}=f(x_t,\,a_{k(t)},\,d_t), \\
& T_\ell \le T_{i,t} \le T_h, \qquad C_{i,t}\le C_{\mathrm{vio}} \quad (\forall i,\ \forall t\in\mathcal{T}_{\mathrm{on}}), \\
& a_k\in\mathcal{A}.
\end{aligned}
\end{equation}
Here, the first line is the energy objective, the second line is the physical state transition, and the remaining lines impose indoor-environment and actuator constraints. The internal emulator state $x_t$ evolves according to nonlinear dynamics $f$; $d_t$ contains uncontrollable weather and occupancy; $k(t)$ maps each one-minute emulator step to the latest five-minute decision; and $\mathcal{A}$ contains the actuator limits. The comfort band is $T_\ell=25\,^{\circ}$C to $T_h=27\,^{\circ}$C around a $26.0\,^{\circ}$C setpoint. The zone CO$_2$ limit is $C_{\mathrm{vio}}=1000$~ppm, consistent with Japan's building environmental hygiene management standard \cite{mhlw_building_environment_criteria}. Because $f(\cdot)$ is nonlinear and high-dimensional and each controller observes only part of $x_t$, Eq.~\eqref{eq:control-problem} is not solved exactly. Instead, it defines the common performance goal. TD3 approximates it with a shaped reward (Section~\ref{subsec:td3-setup}), and the LLMs receive the limits and trade-offs in their prompt (Section~\ref{subsec:llm-setup}).

The seven actuator commands affect temperature and CO$_2$ only through the coupled system dynamics, so these measurements are delayed outcomes rather than direct command variables.

The control action collects all actuator commands into one vector,
\begin{equation}
a_k=\left(a^{\mathrm{vav}}_{1,k},\dots,a^{\mathrm{vav}}_{n,k},\ a^{\mathrm{oa}}_k,\ v^{\mathrm{coil}}_k,\ u^{\mathrm{fan}}_k\right).
\end{equation}
Commands are clipped to the ranges in Table~\ref{tab:observation-action-spaces} before application. In the branch pressure--flow calculation, zone-damper commands below 0.40 are raised to the terminal's minimum effective opening of 0.40 while the air handler is operating. Commands are updated every five minutes and held constant between updates.

Table~\ref{tab:observation-action-spaces} summarizes the partial observation and action spaces used by the learned controllers for $n$ zones. TD3 receives the numerical encoding shown; Section~\ref{subsec:llm-setup} describes the corresponding text encoding for the LLMs.

\begin{table}[H]
\centering
\caption{Observation and action spaces for the learned controllers with $n$ zones ($n=4$ in this study). TD3 uses the numerical observation encoding shown; the LLMs receive the corresponding physical information as text.}
\label{tab:observation-action-spaces}
\footnotesize
\setlength{\tabcolsep}{4pt}
\begin{tabular*}{\textwidth}{@{\extracolsep{\fill}}llcc@{}}
\toprule
Space & Component & Dimension & Unit or admissible range \\
\midrule
\multirow{8}{*}{\makecell[l]{Observation\\$o_k\in\mathbb{R}^{5n+7}$\\($\mathbb{R}^{27}$ for $n=4$)}}
& Zone-temperature error $T_{i,k}-26$ & $n$ & $^{\circ}$C \\
& Latest one-minute zone-temperature change & $n$ & $^{\circ}$C/min \\
& Zone CO$_2$ error $C_{i,k}-1000$ & $n$ & ppm \\
& Latest one-minute zone CO$_2$ change & $n$ & ppm/min \\
& Outdoor-air temperature & 1 & $^{\circ}$C \\
& Outdoor-temperature trend & 1 & $^{\circ}$C/h \\
& Time of day (sine and cosine) & 2 & $[-1,1]$ \\
& Previous action $a_{k-1}$ & $n+3$ & Ranges below \\
\midrule
\multirow{4}{*}{\makecell[l]{Action\\$a_k\in\mathbb{R}^{n+3}$\\($\mathbb{R}^{7}$ for $n=4$)}}
& Zone-damper commands $a^{\mathrm{vav}}_{i,k}$ & $n$ & $[0.05,1.00]$ \\
& Outdoor-air damper $a^{\mathrm{oa}}_k$ & 1 & $[0.05,0.60]$ \\
& Chilled-water valve $v^{\mathrm{coil}}_k$ & 1 & $[0.06,1.00]$ \\
& Supply-fan speed ratio $u^{\mathrm{fan}}_k$ & 1 & $[0.45,1.40]$ \\
\bottomrule
\end{tabular*}
\end{table}

The action combines four local airflow allocations with three shared air-handler commands. Changing a shared command affects all zones, while changing one terminal can redistribute flow through the duct network. Even a coarse grid with ten values per actuator would contain $10^7$ combinations at one decision.

\subsection{VAV system emulator}\label{subsec:simulator}
The VAV emulator couples and extends component models from the open-source HVAC library \textit{phyvac} \cite{miyata2023_phyvac}. Here, \emph{physics-based} means that pressure losses, mass and energy balances, air mixing, coil heat transfer, equipment power, and zone temperature and CO$_2$ balances are computed from explicit equations rather than learned end to end from controller data. The library provides equipment and flow-balance models. For this study, we added the four-zone layout, air--water coupling, zone dynamics, operating logic, and state save-and-restore interface. The emulator includes an air network, outdoor-air mixing box, cooling and dehumidifying coil, chilled-water circuit, and four zones. It is a controlled research testbed, not a calibrated digital twin of a particular building. Appendix~\ref{app:emulator-parameters} gives the component equations, pressure--flow curves, solution order, and numerical settings.

The emulator advances in one-minute steps. At each step, it solves the air network, mixes return and outdoor air, solves the coil and chilled-water circuit, and updates zone temperature, humidity, and CO$_2$. Its save-and-restore interface provides reproducible checkpoints. The saved state contains all zone, equipment, filter, and local-controller variables needed to restart a branch:
\begin{equation}\label{eq:emulator-state}
x_t=\left(\mathbf{T}_t,\mathbf{w}_t,\mathbf{C}_t,
x_t^{\mathrm{eqp}},x_t^{\mathrm{ctrl}}\right).
\end{equation}
Outdoor temperature and relative humidity are taken from JMA 10-minute observations for Tokyo station 47662 and linearly interpolated to the simulation grid \cite{jma_tokyo_10min_2025}.

Both the evaluation objective and the TD3 reward use the following equipment-power model. Moist-air enthalpy $h$ accounts for both dry-air temperature and water vapor. The positive enthalpy drop across the coil gives the total cooling load, including sensible cooling and latent dehumidification:
\begin{equation}\label{eq:coil-load}
\begin{gathered}
h(T,w)=c_{pa}T+\left(r_0+c_{pv}T\right)w, \\
Q_{\mathrm{tot}}
=\dot m_{\mathrm{a}}[h_{\mathrm{ma}}-h_{\mathrm{sa}}]_+
=Q_{\mathrm{sens}}+Q_{\mathrm{lat}}, \\
[y]_+=\max(y,0).
\end{gathered}
\end{equation}
With $G$ the total supply flow in m$^3$/min, $\dot V_{\mathrm{chw}}$ the chilled-water flow in m$^3$/s, pressure in Pa, and power in kW, the equipment and total powers are
\begin{equation}\label{eq:system-power}
\begin{aligned}
P_{\mathrm{fan}}
&=\frac{10^{-3}(G/60)\,\Delta P_{\mathrm{fan}}}
{\eta_{\mathrm{fan}}(G)\eta_{\mathrm{mot}}}, \\
P_{\mathrm{pump}}
&=\frac{10^{-3}\dot V_{\mathrm{chw}}\Delta P_{\mathrm{pump}}}
{\eta_{\mathrm{p}}},
&
P_{\mathrm{ch}}
&=\frac{Q_{\mathrm{tot}}}{\mathrm{COP}}, \\
P_t
&=P_{\mathrm{fan}}+P_{\mathrm{pump}}+P_{\mathrm{ch}}.
\end{aligned}
\end{equation}
Fan and pump power equal hydraulic or air power divided by efficiency, while chiller power is cooling load divided by its coefficient of performance (COP). Here $Q_{\mathrm{tot}}$ is in kW because $h$ is in kJ/kg and $\dot m_{\mathrm{a}}$ is in kg/s; $10^{-3}$ converts W to kW. The emulator uses $\mathrm{COP}=4.0$, meaning that one unit of chiller electricity supplies four units of cooling under the assumed model. Equation~\eqref{eq:system-power} supplies the total power used in Eq.~\eqref{eq:control-problem} and in the energy term of Eq.~\eqref{eq:reward}.

The emulated system operates from 8:00 to 18:00. Outside these hours, the dampers close, outdoor-air intake is minimized, and controller integrators reset. The emulator is deterministic: restoring the state in Eq.~\eqref{eq:emulator-state}, fixing future disturbances, and applying the same commands reproduces the same trajectory to floating-point precision. This property enables the direct rollout verifier in Sections~\ref{subsec:verifier-selection} and \ref{subsec:rollout-rft}.

\subsection{Evaluation scenario}\label{subsec:scenario}
\begin{figure}[H]
\centering
\begin{minipage}[t]{0.495\textwidth}
\centering
\includegraphics[width=\textwidth]{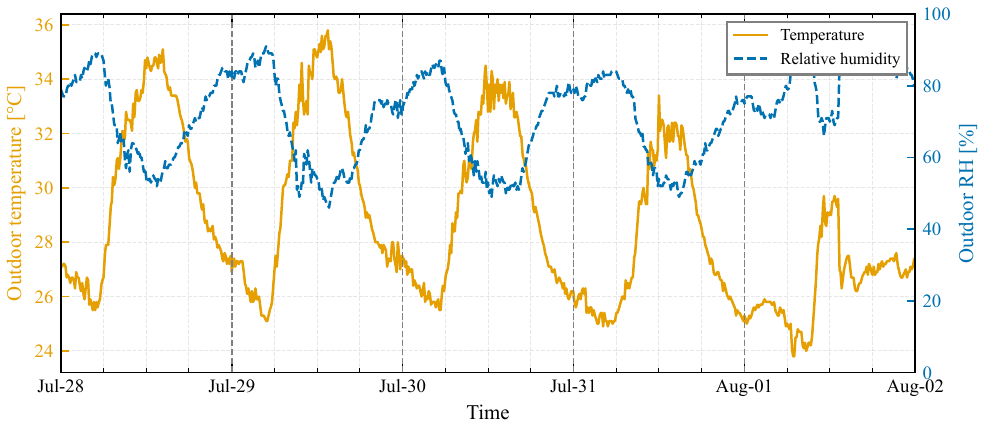}

{\small (a) Outdoor temperature and relative humidity}
\end{minipage}\hfill
\begin{minipage}[t]{0.495\textwidth}
\centering
\includegraphics[width=\textwidth]{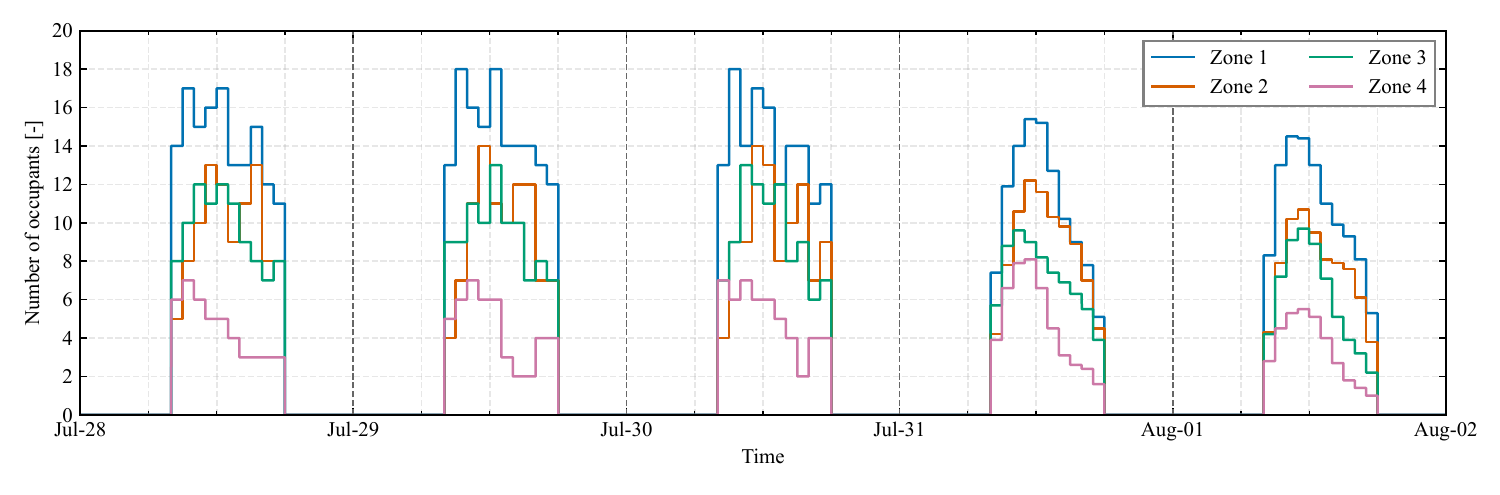}

{\small (b) Zone occupancy schedules}
\end{minipage}
\caption{Evaluation scenario, 28 July--1 August 2025 (Tokyo): (a) outdoor temperature and relative humidity supplied to the emulator; (b) occupancy schedules of the four zones. Occupancy schedules differ by day. Vertical dashed lines mark day boundaries. Occupancy is zero outside each zone's occupied period.}
\label{fig:scenario}
\end{figure}

\begin{sloppypar}
The controllers are evaluated on five consecutive summer weekdays, 28 July--1 August 2025, with the Tokyo weather data described in Section~\ref{subsec:simulator}. Each day has its own occupancy schedule; the schedules for 31 July and 1 August have lower, smoother peaks than those for 28--30 July. Internal heat gains follow the occupancy-dependent model, and all controllers use the same weather and occupancy conditions. Figure~\ref{fig:scenario} shows the outdoor conditions and occupancy over all five days. During occupied hours, the maximum outdoor temperatures were 35.1, 35.8, 34.5, 33.4, and 29.7$\,^{\circ}$C, respectively. The final day was cooler and more humid, with mean occupied-hour outdoor relative humidity of 84.6\%. Staggered occupancy creates unequal loads among the zones throughout the day. GPT-5 was evaluated in three independent runs with the same prompt, model settings, weather, occupancy, and initial state. \texttt{gpt-oss-20b} was evaluated once before and once after RFT.
\end{sloppypar}

\subsection{Rule-based baseline (Guideline 36)}\label{subsec:g36-setup}
The rule-based baseline, called \emph{G36-based} below, combines local PI feedback with duct-static-pressure and supply-air-temperature Trim \& Respond sequences adapted from ASHRAE Guideline~36 \cite{ref:ashrae_g36_2024}. In Trim \& Respond, a shared setpoint is gradually \emph{trimmed} toward lower-energy operation when no zone needs more capacity and \emph{responds} in the opposite direction when one or more zones issue requests. This allows many local loops to coordinate a central fan or coil without solving a numerical optimization problem. The CO$_2$ loop and all numerical settings below are specific to this study.

The baseline operates through four linked loops:
\begin{enumerate}
  \item \textbf{Zone airflow.} An incremental PI controller adjusts each VAV damper from its zone-temperature error ($K_p=1.1667$, $T_i=100$~min, per-step change limited to $\pm 0.10$, and integral reset after 30 minutes of one-sided error).
  \item \textbf{Duct pressure.} Dampers above 95\% generate pressure requests. After a 10-minute startup hold, Trim \& Respond changes the duct static-pressure setpoint every 2 minutes by $-12$~Pa when capacity can be trimmed or by $+12$~Pa per request, within [250, 600]~Pa and from an initial value of 500~Pa. The intent is to provide enough pressure for the most open terminal without wasting fan power through excessive throttling.
  \item \textbf{Supply-air temperature.} Zone cooling requests reset the supply-air temperature setpoint within [14, 18]$\,^{\circ}$C. Requests depend on the duration and size of over-temperature and on a hysteresis latch for nearly saturated dampers.
  \item \textbf{Ventilation.} A PI controller adjusts the outdoor-air damper from the highest zone CO$_2$ concentration. Uneven zone loads can leave the least-supplied critical zone near or above the 1000~ppm evaluation limit \cite{xu2007_adaptive_dcv}, so the loop uses a 950~ppm setpoint to provide a 50~ppm control margin, together with a 50~ppm deadband, output limits [0.10, 0.50], anti-windup, and startup reset.
\end{enumerate}

\subsection{Reinforcement learning baseline (TD3)}\label{subsec:td3-setup}
The RL baseline learns through repeated interaction with the emulator: at each five-minute decision, its actor selects seven continuous commands and its critic supplies the learning signal (Table~\ref{tab:concept-map}).

We use twin-delayed deep deterministic policy gradient (TD3), an actor--critic algorithm designed for continuous actions \cite{ref:fujimoto2018_td3}. TD3 trains two critics and uses the smaller target estimate to reduce optimistic value errors. It adds noise to target actions so that the learned value is less affected by narrow action peaks, and it updates the actor less often than the critics to improve stability. During evaluation, the training noise and critics are removed from the control loop; the frozen deterministic actor alone maps observations to actions.

TD3 was selected not only as a performance baseline but also because its deterministic evaluation actor gives each candidate branch a reproducible continuation. Its critics estimate the discounted continuation return in Eq.~\eqref{eq:state-q-definition} without an entropy bonus, making them natural verifier candidates; Section~\ref{subsec:verifier-selection} reports the critic audit and the selection of direct rollout. This methodological fit does not imply that TD3 generally outperforms other RL algorithms.

TD3 receives the 27-dimensional observation in Table~\ref{tab:observation-action-spaces}. The error terms tell the policy how far each zone is from its target; the changes and previous action provide one step of local trend information; and the cyclic time features distinguish, for example, morning startup from late-afternoon shutdown. A running mean and variance (Welford's algorithm) normalize variables with different units, after which extreme normalized values are clipped. A $\tanh$ actor produces an action in $[-1,1]^7$, and a linear map converts it to the physical actuator ranges in Table~\ref{tab:observation-action-spaces}. The action is held for five minutes. Replay transitions are stored only at decision times, with the five one-minute rewards summed.

The actor cannot optimize the constrained problem in Eq.~\eqref{eq:control-problem} directly, so training converts its goals into one scalar \emph{shaped reward}. The design gives positive reward for temperatures near 26$\,^{\circ}$C and subtracts penalties for temperature-band violation, high CO$_2$, severe CO$_2$ violation, and electrical power. Smooth terms provide a learning gradient before a hard threshold is crossed, while the larger violation terms discourage unsafe trade-offs.

The reward is evaluated once per emulator minute. Let $\ell$ denote a one-minute sample, $n=4$ the number of zones, $C^{\max}_{\ell}=\max_i C_{i,\ell}$ the highest zone CO$_2$ concentration, and $[x]_+=\max(x,0)$. Let $d_{i,\ell}=[25-T_{i,\ell}]_+ + [T_{i,\ell}-27]_+$ be the distance from the comfort band, $\sigma(\cdot)$ the logistic function, and $I[\cdot]$ an indicator equal to one when its condition is true. Temperature and CO$_2$ each have a smooth term and a violation term:
\begin{equation}\label{eq:reward-terms}
\begin{aligned}
\phi^{(T)}_{\ell}&=\frac{1}{n}\sum_{i=1}^{n}
\exp\!\left(-\frac{1}{2}\left(\frac{T_{i,\ell}-26}{0.6}\right)^{2}\right),\\
v^{(T)}_{\ell}&=\frac{1}{n}\sum_{i=1}^{n}d_{i,\ell}^{2},\\
\phi^{(C)}_{\ell}&=\sigma\!\left(\frac{C^{\max}_{\ell}-1000}{12}\right),
\\
v^{(C)}_{\ell}&=I\!\left[C^{\max}_{\ell}>1100\right].
\end{aligned}
\end{equation}
The weighted terms are grouped into the temperature, CO$_2$, and electricity reward terms:
\begin{equation}\label{eq:reward}
\begin{aligned}
R^{(T)}_{\ell}
&=\omega_{\mathrm{T}}\phi^{(T)}_{\ell}
-\omega_{\mathrm{B}}v^{(T)}_{\ell}, \\
R^{(C)}_{\ell}
&=-\omega_{\mathrm{C}}\phi^{(C)}_{\ell}
-\omega_{\mathrm{V}}v^{(C)}_{\ell}, \\
R^{(E)}_{\ell}
&=-\omega_{\mathrm{E}}P_{\ell}, \\
r_t
&=\sum_{\ell\in\mathcal{I}_t}m_{\ell}
\left(R^{(T)}_{\ell}+R^{(C)}_{\ell}+R^{(E)}_{\ell}\right).
\end{aligned}
\end{equation}
The weights are $\omega_{\mathrm{T}}=1.0$, $\omega_{\mathrm{B}}=2.0$, $\omega_{\mathrm{C}}=0.5$, $\omega_{\mathrm{V}}=10.0$, and $\omega_{\mathrm{E}}=0.2$ for temperature, temperature-band violation, CO$_2$, severe CO$_2$ violation, and power, respectively. Here, $P_{\ell}$ is total fan, pump, and chiller power in kW; $m_{\ell}$ is one only while the HVAC is scheduled and active; and $\mathcal{I}_t$ contains the five one-minute samples in interval $t$. A higher $r_t$ is better. The logistic argument is clipped to $\pm 60$ to prevent numerical overflow. The same $r_t$ is stored in the TD3 replay buffer and used by the rollout verifier in Eq.~\eqref{eq:state-q-definition}, ensuring that the teacher and verifier value the same defined behavior.

\begin{figure}[H]
\centering
\begin{minipage}[t]{0.48\textwidth}
\centering
\includegraphics[width=\textwidth]{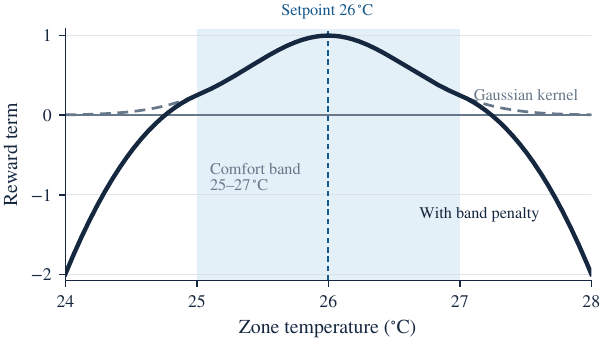}

{\small (a) Temperature reward term $\omega_{\mathrm{T}}\phi^{(T)}_{\ell}-\omega_{\mathrm{B}}v^{(T)}_{\ell}$}
\end{minipage}\hfill
\begin{minipage}[t]{0.48\textwidth}
\centering
\includegraphics[width=\textwidth]{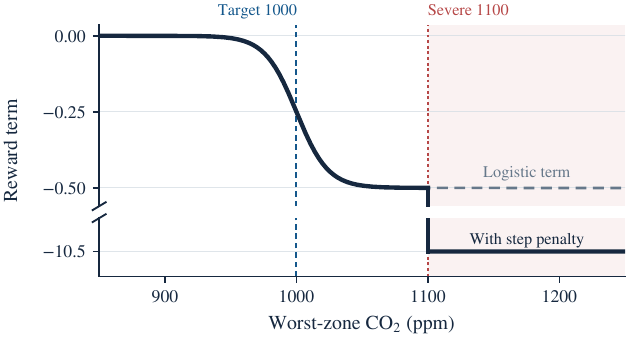}

{\small (b) CO$_2$ reward term $-\,\omega_{\mathrm{C}}\phi^{(C)}_{\ell}-\omega_{\mathrm{V}}v^{(C)}_{\ell}$}
\end{minipage}
\caption{Shaped reward terms used for TD3 training: (a) Gaussian temperature term with a quadratic penalty outside the comfort band; (b) logistic CO$_2$ penalty centered at 1000~ppm, with an additional fixed penalty above 1100~ppm.}
\label{fig:reward_shapes}
\end{figure}

Figure~\ref{fig:reward_shapes} plots these reward terms. Equation~\eqref{eq:reward} is a task-specific approximation of the constrained objective in Eq.~\eqref{eq:control-problem}; it is not mathematically equivalent to that objective. Even an exactly computed rollout can reinforce an undesirable preference if the reward weights encode the wrong trade-off. For this reason, the final evaluation reports energy, temperature, and CO$_2$ separately instead of reporting only RL return.

Training uses July 2024 weather data---one year before the evaluation window---for 10{,}000 episodes; the evaluation phase runs the frozen actor without exploration noise or normalizer updates. Appendix Table~\ref{tab:td3_hparams} gives the complete TD3 configuration. The resulting training trajectory is reported in Section~\ref{subsec:td3-training-results}.

\subsection{LLM controllers}\label{subsec:llm-setup}
We evaluate two reasoning models as controllers using the interface in Section~\ref{subsec:controller-design}. GPT-5 (\texttt{gpt-5-2025-08-07}), accessed through the OpenAI API, is the capability reference: it tests what a high-capability hosted model can do without building-specific weight updates \cite{ref:gpt5_system_card}. \texttt{gpt-oss-20b} is a 20.9B-parameter open-weight model served locally. It represents a model that can be adapted and deployed on premises and is the only model updated by RFT \cite{ref:openai2025_gptoss}. The comparison therefore serves two distinct purposes rather than treating model size or access mode as a controlled variable.

Both models receive exactly the same task description and observation schema. Every five minutes, each model returns one JSON action, which is parsed, clipped to the actuator bounds, and held until the next decision. Both models use medium reasoning effort; \texttt{gpt-oss-20b} additionally uses a sampling temperature of 0 during closed-loop evaluation. These settings govern inference-time computation and output variability, respectively; they neither make the models equivalent nor guarantee physically correct outputs. Appendix~\ref{app:evaluation-prompts} gives the separate settings for transition prediction, and Appendix~\ref{app:control-prompt} gives the full control prompt.

In preliminary runs with only a 1000~ppm target, GPT-5 tended to maintain CO$_2$ near 900~ppm. We therefore added a separate 1050~ppm upper limit during development of the prompt in Appendix~\ref{app:control-prompt}.

Each observation JSON contains the physical information corresponding to Table~\ref{tab:observation-action-spaces}: absolute zone measurements and the fixed targets, from which the listed errors are recoverable, and clock time in place of the sine--cosine encoding. It also carries fixed metadata such as the control interval and actuator bounds. Neither LLM observes the supply-air temperature sensor used inside the rule-based baseline. We log the GPT-5 reasoning summary and the \texttt{gpt-oss-20b} analysis channel for later qualitative analysis.

\subsection{Rollout-verified RFT configuration}\label{subsec:rollout-setup}
The reported run adapts \texttt{gpt-oss-20b} with Dr.~GRPO and LoRA for 200 optimizer steps. This is one training run with a fixed seed, not a hyperparameter search or an estimate over multiple random seeds. Each generated batch receives one policy update with clipping parameter $\epsilon=0.2$, KL coefficient $\beta=0$, and no reference-model penalty. The base-model weights remain frozen, and only the LoRA adapters are updated. Appendix Table~\ref{tab:vav-rft-hparams} lists all model-loading, optimizer, sampling, memory, and software settings.

Rewards follow Eq.~\eqref{eq:softmax-reward} with $\tau=0.05$; no learned critic score or auxiliary reward is used. We chose this low temperature to strongly reinforce the candidate with the highest rollout return. This can place nearly all reward weight on one candidate, so differences among the other candidates have little effect on the update. Each optimizer step contains four same-state groups, one per gradient-accumulation micro-step, and each group contains 16 independently sampled responses. The four micro-steps allow information from four prompts to contribute before one parameter update. This yields 64 generated responses per update and 12{,}800 generations over 200 updates; only valid, non-truncated responses receive rollout scores and participate in the loss. Of these generations, 614 (4.80\%) were invalid or truncated. No actions required clipping. Each group of 16 contained, on average, 15.2 valid responses and approximately 12.6 distinct valid actions; 2.6 valid responses repeated an action already present in the group. Appendix~\ref{app:generation-quality} gives the failure breakdown.

Training ran on one NVIDIA H100 SXM GPU with 80~GB of memory and took 52.42~h. The group size was set to 16, the largest number of responses of up to 4{,}000 tokens that fit in this memory under our training configuration. The budget of 200 optimizer steps matches our previous TES study \cite{shioda2026_tes_rft} and was also constrained by the available computation time. The candidates support within-state comparisons but do not provide exhaustive coverage of the seven-dimensional action space.

The 30 training states are selected solely by time before the run, without using critic values, rollout returns, or model outputs. They comprise 10 hourly time points from 08:20 to 17:20 on each of 21--23 July 2025, taken from three saved TD3 trajectories. The corresponding prompts use the same format as those used for closed-loop control. The training dates do not overlap the 28 July--1 August control-evaluation period. Each valid action is scored by the direct rollout in Section~\ref{subsec:verifier-selection} with $\gamma=0.995$ per decision interval, matching TD3 training and the pilot audit; thus the teacher, the audited critic, and the verifier value the same return. A resumable cache stores rollout scores so that each unique state--action branch is evaluated only once.

Appendix~\ref{app:computational-cost} reports the computational requirements for LLM inference, RFT training, and rollout scoring.

We monitor sampled-action quality through return gaps from the deterministic TD3 action at the same state. Section~\ref{subsec:rft-training-results} defines and reports these metrics. They are not used as training rewards.

\section{Results}\label{sec:results}
\subsection{TD3 training}\label{subsec:td3-training-results}
\begin{figure}[H]
\centering
\begin{minipage}[t]{0.40\textwidth}
\centering
\begin{minipage}[c][0.22\textheight][c]{\linewidth}
\centering
\includegraphics[width=\linewidth,height=0.22\textheight,keepaspectratio]{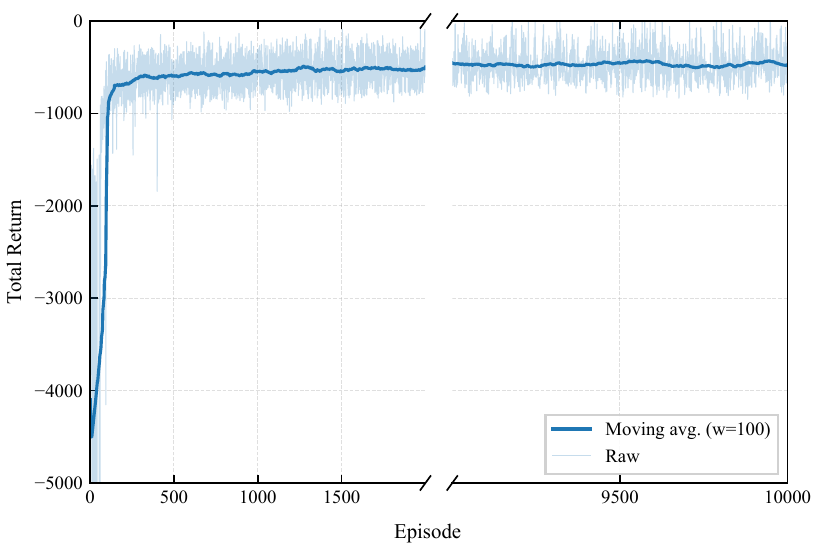}
\end{minipage}
{\small (a) TD3 training curve}
\end{minipage}\hfill
\begin{minipage}[t]{0.575\textwidth}
\centering
\begin{minipage}[c][0.22\textheight][c]{\linewidth}
\centering
\includegraphics[width=\linewidth,height=0.22\textheight,keepaspectratio]{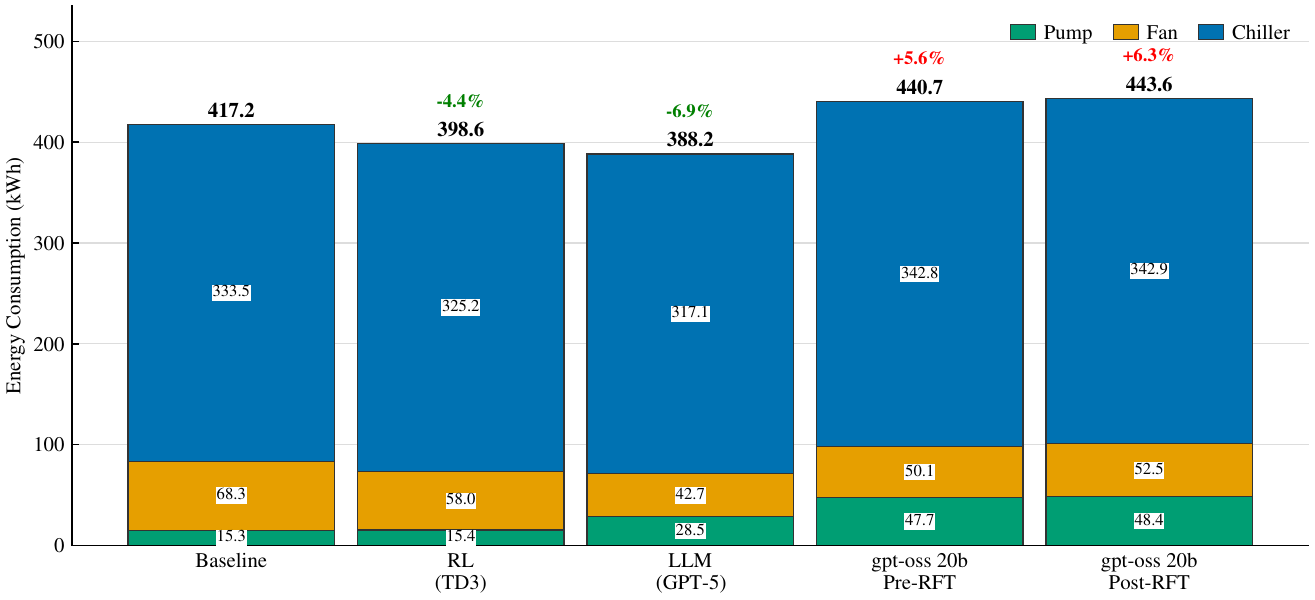}
\end{minipage}
{\small (b) Five-day HVAC electricity use}
\end{minipage}
\caption{TD3 training and equipment electricity use: (a) total episode return versus training episode on the four-zone VAV emulator, with raw returns and a 100-episode moving average; (b) chiller, fan, and pump electricity use for each controller over 28 July--1 August 2025.}
\label{fig:td3_training}
\label{fig:control-energy}
\end{figure}

Figure~\ref{fig:td3_training} shows the TD3 training curve. An episode return is the cumulative shaped reward over one simulated operating episode; a larger value means that the policy performed better under the particular combination of temperature, CO$_2$, and power terms in Eq.~\eqref{eq:reward}. The return rises quickly during the first few hundred episodes and then remains stable through episode 10{,}000. Training therefore stabilized under that reward.

\subsection{Control results}\label{subsec:control-results}
Table~\ref{tab:control-performance} compares the five controllers over four zones, five days, and ten occupied hours per day, for a total of 200 occupied zone-hours. No single column is sufficient to rank a constrained controller. Energy measures efficiency; compliance is the fraction of occupied zone-minutes that satisfy a limit; integrated deviation measures both the magnitude and duration of violations; and the maximum CO$_2$ value exposes the worst event.

All controllers are evaluated against a 1000~ppm threshold. The G36-based sequence uses a 950~ppm setpoint for the reason given in Section~\ref{subsec:g36-setup}; the prompt in Appendix~\ref{app:control-prompt} specifies a 1000~ppm target and a 1050~ppm upper limit. The TD3 reward, also used by the RFT verifier, uses a smooth penalty centered at 1000~ppm and an additional penalty above 1100~ppm. The comparison reflects both the control methods and their chosen CO$_2$ settings. These settings may partly explain GPT-5's lower outdoor-air intake and lower CO$_2$ compliance. TD3, however, used less electricity with more outdoor air. We did not isolate the effect of the CO$_2$ settings.

\begin{figure}[H]
\centering
\includegraphics[width=\textwidth]{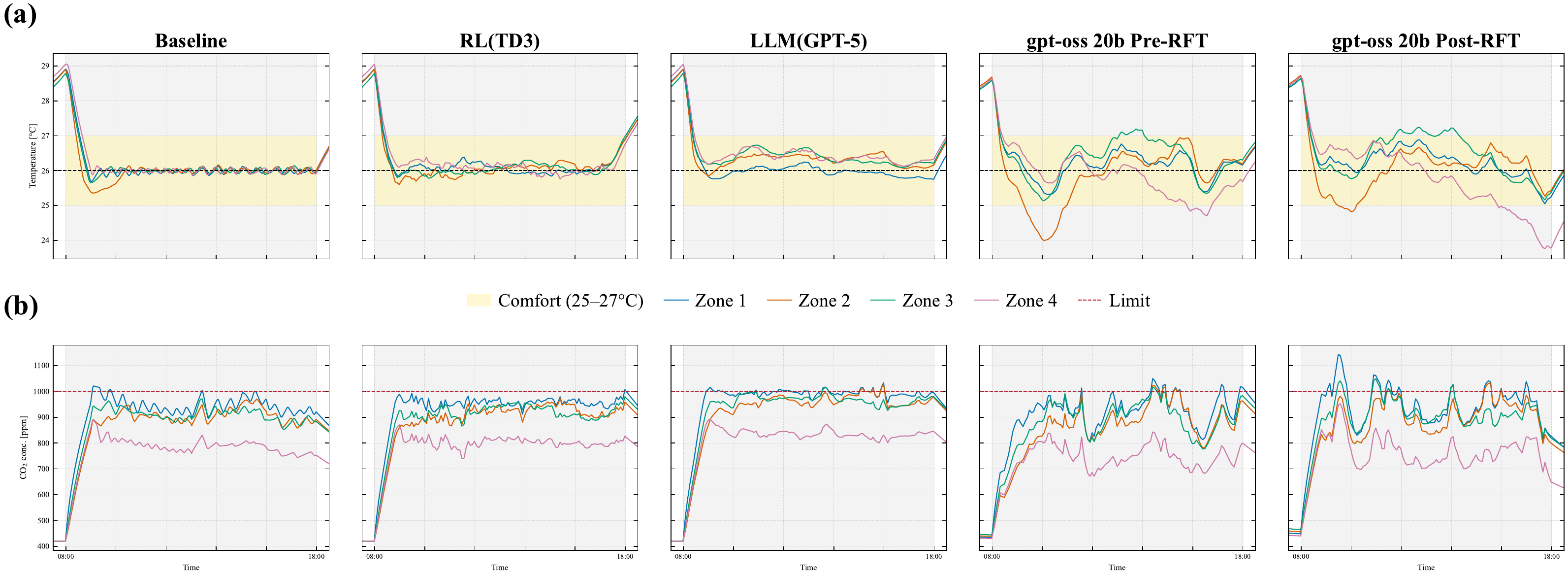}
\caption{Representative occupied-day profiles for the five controllers: (a) zone temperature, with the shaded 25--27$\,^{\circ}$C band; and (b) zone CO$_2$, with the dashed 1000~ppm target.}
\label{fig:control-temperature-co2}
\end{figure}

Electricity is reported in kWh ($1~\mathrm{kWh}=3.6~\mathrm{MJ}$), and CO$_2$ concentration in ppm by volume ($1~\mathrm{ppm}=10^{-6}$ by volume). Compliance rates use every occupied zone-minute. Temperature deviation is the time integral outside 25--27$\,^{\circ}$C: one zone remaining 1$\,^{\circ}$C outside the band for one hour contributes 1$\,^{\circ}$C$\cdot$h. CO$_2$ deviation is defined analogously above 1000~ppm in ppm$\cdot$h. Both are summed across zones, not averaged. Negative $\Delta E$ means lower electricity than the G36-based baseline, whereas positive $\Delta E$ means higher electricity; OA in the table denotes outdoor air. Figure~\ref{fig:control-energy} shows the five-day electricity use by equipment, while Figure~\ref{fig:control-temperature-co2} shows temperature and CO$_2$ profiles for one occupied day. Appendix Figures~\ref{fig:control-detail-reference} and~\ref{fig:control-detail-gptoss} show the full control trajectories.

\begin{table}[H]
\centering
\caption{Electricity use, outdoor-air intake, and temperature and CO$_2$ performance over 28 July--1 August 2025. Compliance and deviation metrics cover 08:00--18:00 across four zones. $\Delta E$ is relative to the G36-based baseline; best values are shown in bold. The GPT-5 row reports the first of three runs.}
\label{tab:control-performance}
\footnotesize
\setlength{\tabcolsep}{2.5pt}
\begin{tabular*}{\textwidth}{@{\extracolsep{\fill}}lrrrrrrrr@{}}
\toprule
Controller &
\makecell{Energy\\(kWh)} &
\makecell{$\Delta E$\\(\%)} &
\makecell{OA intake\\(kg)} &
\makecell{$25\leq T\leq27\,^{\circ}\mathrm{C}$\\(\%)} &
\makecell{Temp. dev.\\($^{\circ}$C$\cdot$h)} &
\makecell{CO$_2\leq1000$~ppm\\(\%)} &
\makecell{CO$_2$ dev.\\(ppm$\cdot$h)} &
\makecell{Max CO$_2$\\(ppm)} \\
\midrule
G36-based & 417.18 & --- & 69{,}378 & 90.10 & 13.81 & 97.29 & 124.83 & 1083 \\
TD3 & 398.63 & $-4.45$ & 73{,}472 & \textbf{97.61} & \textbf{4.22} & \textbf{99.98} & \textbf{0.17} & \textbf{1007} \\
GPT-5 & \textbf{388.23} & \textbf{$-6.94$} & 63{,}963 & 97.38 & 4.64 & 94.95 & 52.82 & 1033 \\
\texttt{gpt-oss} pre-RFT & 440.70 & $+5.64$ & 72{,}124 & 82.96 & 24.31 & 88.23 & 324.14 & 1201 \\
\texttt{gpt-oss} post-RFT & 443.64 & $+6.34$ & 71{,}842 & 79.84 & 25.81 & 87.53 & 350.72 & 1180 \\
\bottomrule
\end{tabular*}
\end{table}

Two results should be distinguished at the outset. First, GPT-5 and TD3 both reduce total electricity relative to the rule-based baseline, but only TD3 improves every reported temperature and CO$_2$ metric. GPT-5's larger energy reduction coincides with lower outdoor-air intake and lower CO$_2$ compliance. Second, RFT does not make \texttt{gpt-oss-20b} competitive with the baseline: the post-RFT model remains higher in energy and worse in temperature and CO$_2$ performance. The following trajectory analysis explains how the controllers reach these totals.

Zone relative humidity (RH) was within 40--70\% for 99.89--99.91\% of occupied zone-minutes across all controllers, showing little difference in this measure under the evaluated conditions (Appendix~\ref{app:control-trajectories}).

The main control table, figures, and period-specific comparisons report the first GPT-5 run. Appendix~\ref{app:gpt5-repeats} reports all three runs.

The actuator interpretations below use the representative 28 July trajectories in Figure~\ref{fig:control-temperature-co2} and Appendix Figures~\ref{fig:control-detail-reference} and~\ref{fig:control-detail-gptoss}.

\smallskip
\begin{sloppypar}
\noindent\textbf{G36-based controller.} On the representative day, the rule-based controller tracks the 26$\,^{\circ}$C setpoint closely after start-up, but this does not minimize electricity use over the five-day evaluation. Despite its 950~ppm ventilation setpoint, five-day CO$_2$ compliance is 97.29\%, with a maximum concentration of 1083~ppm. Outdoor-air intake is higher than for GPT-5. Appendix~\ref{app:g36-final-day} examines the final-day temperature and CO$_2$ deviations and their contribution to the five-day totals.
\end{sloppypar}

\smallskip
\begin{sloppypar}
\noindent\textbf{TD3.} TD3 reduces total electricity by 4.45\% and improves all reported temperature and CO$_2$ indicators relative to G36. Fan and chiller electricity decrease from 68.3 and 333.5~kWh to 58.0 and 325.2~kWh, respectively, while pump electricity increases slightly from 15.3 to 15.4~kWh. Outdoor-air intake increases by approximately 5.9\%, so the energy reduction is not accompanied by reduced ventilation. On the representative day, at least one zone damper remains nearly fully open through most of the occupied period, while fan speed falls as the load decreases toward 18:00. This behavior reduces damper-induced airflow restriction and is consistent with the lower fan use. TD3 also allows temperature to rise modestly near the end of operation while remaining within the comfort band.
\end{sloppypar}

\smallskip
\begin{sloppypar}
\noindent\textbf{GPT-5.} GPT-5 uses the least electricity over five days: 388.23~kWh versus 417.18~kWh for G36, a reduction of 6.94\%. Outdoor-air intake is 63{,}963 versus 69{,}378~kg, approximately 7.8\% lower. CO$_2$ compliance is lower (94.95\% versus 97.29\%), but cumulative exceedance above 1000~ppm is smaller (52.82 versus 124.83~ppm$\cdot$h), as is the maximum concentration (1033 versus 1083~ppm). Fan and chiller electricity decrease from 68.3 and 333.5~kWh to 42.7 and 317.1~kWh, respectively, whereas pump electricity increases from 15.3 to 28.5~kWh. The fan and chiller savings exceed the additional pumping cost. The representative trajectory shows a lower supply-air temperature and a lower, steadier fan command, consistent with this air--water trade-off. GPT-5 therefore has the lowest total electricity use, but is not superior on every performance indicator.
\end{sloppypar}

Across three GPT-5 runs, electricity use ranged from 386.53 to 388.84~kWh, or 6.79--7.35\% below G36. All three runs used less electricity than TD3 (398.63~kWh). Temperature compliance was 97.21--97.51\%, close to TD3's 97.61\%, while CO$_2$ compliance was 92.71--94.95\%, below both G36 and TD3. Total electricity use was similar across runs despite differences in pump and fan use (Appendix~\ref{app:gpt5-repeats}).

GPT-5 also stated short-horizon forecasts before choosing some actions. In the 09:20 example reproduced in Appendix~\ref{app:gpt5-forecast-example}, it first estimated the zone state after five minutes if the commands were unchanged, then described the expected response to its selected action. The forecast was numerically close: it placed zone~1 near 25.8--26.0$\,^{\circ}$C with CO$_2$ below 1000~ppm and expected zone~4 to cool modestly toward 26$\,^{\circ}$C. Five minutes later, the emulator measured 25.82$\,^{\circ}$C and 991~ppm in zone~1, with zone~4 at 26.47$\,^{\circ}$C cooling by less than predicted. This example suggests that a short-term system forecast informed the control decision. Section~\ref{sec:transition-evaluation} tests the same capability directly across controlled counterfactual branches.

\smallskip
\begin{sloppypar}
\noindent\textbf{\texttt{gpt-oss-20b}.} Before RFT, the controller does not maintain stable coordination among temperature control, ventilation, and equipment commands. Its representative trajectory contains large changes in zone dampers, outdoor-air intake, coil-valve position, and fan speed, together with larger temperature and CO$_2$ deviations. The five-day aggregate results show no overall improvement after RFT: electricity increases from 440.70 to 443.64~kWh, both temperature and CO$_2$ compliance decline, and both integrated deviations increase. Maximum CO$_2$ decreases from 1201 to 1180~ppm, but remains above the 1100~ppm severe-violation threshold.
\end{sloppypar}

A qualitative comparison of the reasoning logs also shows little change after RFT. Generated reasoning is not treated as a faithful record of the model's computation, since reasoning models often omit information that influences their answers \cite{chen2025_reasoning_models_faithfulness}; the logs serve here to identify candidate physical misconceptions, which the machine-scored transition test in Section~\ref{sec:transition-evaluation} then checks directly. Appendix~\ref{app:gptoss-reasoning-example} gives representative pre- and post-RFT actions and verbatim excerpts. Before and after training, the model makes local guesses about actuator effects without a stable prediction of the next state or a consistent check against the following observation. The pre-RFT logs illustrate this uncertainty:

\begin{quote}
\small\itshape
``But would that maintain temperature? \textbf{Hard to know}. We can approximate:''\\[\baselineskip]
``but damper open to 0.3 may still deliver enough ventilation? \textbf{Hard to know}.''\\[\baselineskip]
``Will this keep CO$_2$ below 1000? \textbf{Hard to know}.''
\end{quote}

The post-RFT logs still raise the same questions---what a command would do, and whether the result would violate a limit---but the model never produces the prediction those questions require. One post-RFT excerpt names the missing element directly:

\begin{quote}
\small\itshape
``Check if any constraints violation: Temperature zone 3? With OA at 0.05, maybe zone 3 will not get enough ventilation to cool. But coil valve at max maybe enough. \textbf{But we don't have dynamic model. We'll assume it's okay}.''
\end{quote}

In both conditions, the model guesses actuator effects one step at a time, sometimes in the wrong direction, and abandons the constraint check because it cannot produce the forward prediction the check requires. The transition test in Section~\ref{sec:transition-evaluation} examines this failure quantitatively.

\subsection{Rollout-verified RFT}\label{subsec:rft-training-results}
Section~\ref{subsec:verifier-selection} selected direct rollout because the critic's within-state ranking was unreliable. We now test whether this exact rollout signal improves the sampled policy.

\begin{figure}[H]
\centering
\includegraphics[width=\textwidth]{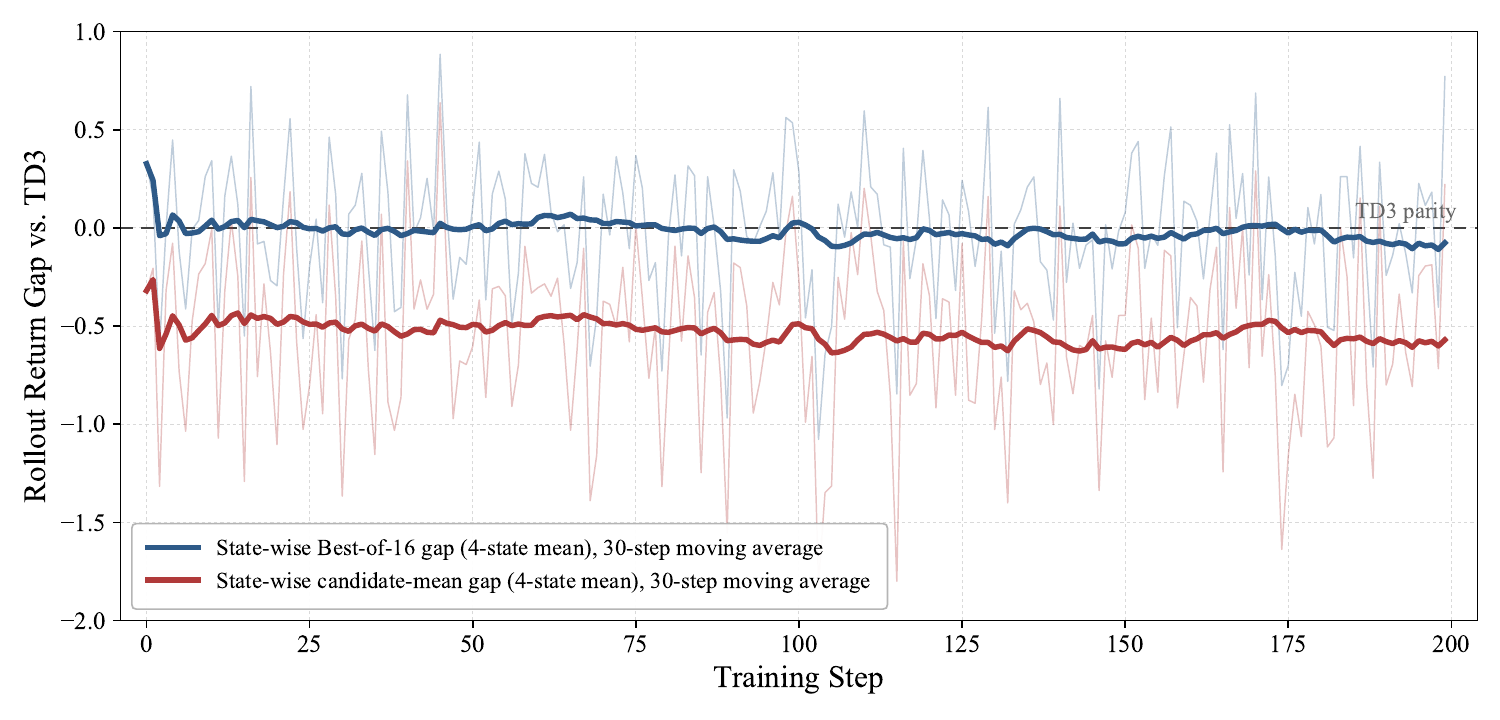}
\caption{Return gaps from the TD3 action over 200 RFT steps. Thin lines show the four-state means, and thick lines show 30-step moving averages. Blue is the best-of-16 gap, red is the candidate-mean return gap, and the dashed line marks equal return to TD3.}
\label{fig:rollout_gap_overlay}
\end{figure}

For the actions of valid responses indexed by $\mathcal{V}$ at state $s$, we measure action quality relative to the deterministic TD3 action. A gap of zero means equal return to the TD3 action under the common TD3 continuation, a positive gap means better return, and a negative gap means worse return:
\begin{equation}\label{eq:gaps}
\begin{aligned}
\Delta_{\mathrm{mean}}(s)
&=\frac{1}{|\mathcal{V}|}\sum_{j\in\mathcal{V}}
G^{\pi_{\mathrm{TD3}}}(s,a^{(j)})\\
&\quad-G^{\pi_{\mathrm{TD3}}}(s,a_{\mathrm{TD3}}),\\
\Delta_{\mathrm{best}}(s)
&=\max_{j\in\mathcal{V}}
G^{\pi_{\mathrm{TD3}}}(s,a^{(j)})\\
&\quad-G^{\pi_{\mathrm{TD3}}}(s,a_{\mathrm{TD3}}).
\end{aligned}
\end{equation}
$\Delta_{\mathrm{mean}}$ is the average quality of the sampled policy, whereas $\Delta_{\mathrm{best}}$ asks whether at least one of the 16 samples contains a useful action that training could reinforce. Together the two gaps distinguish two failure modes. A high best gap with a low mean gap means the model can occasionally propose a good action but does not choose it reliably. Low values for both mean and best indicate that the useful action is rarely present in the sampled group. Invalid generations are excluded. Valid duplicates remain because repeated actions represent probability mass in the policy. Each optimizer step gives equal weight to its four prompt groups.

Figure~\ref{fig:rollout_gap_overlay} shows both gaps during RFT. Individual candidates sometimes outperformed TD3, but the moving average of the best-candidate gap did not stay positive, and the mean sampled-action return remained below TD3 throughout training. In the early (steps 1--67), middle (68--134), and late (135--200) stages, the mean best-candidate gaps were $+0.014$, $-0.049$, and $-0.034$, respectively. The corresponding candidate-average gaps were $-0.488$, $-0.573$, and $-0.542$. Candidate quality therefore varied within groups, but neither measure showed sustained improvement during training. Under these settings, occasionally sampling an action better than TD3 did not lead to a lasting improvement in the sampled policy. These results do not rule out improvement with more candidates or a longer training run.

\section{Testing action-conditioned transition knowledge}\label{sec:transition-evaluation}

A controller needs more than a preference score. To propose a useful new action, it must anticipate at least some local consequences of changing that action; if the anticipated directions of those consequences are grossly wrong, systematic comparison of candidate actions becomes difficult. A predictor that maps the current state and action to the next state is called a \emph{transition model}; when it supports prediction and planning over an environment, it is often called a \emph{world model}, a role language models have been used to fill during planning \cite{hao2023_rap}.

The qualitative logs in Section~\ref{subsec:control-results} suggest a difference in action-conditioned transition knowledge. GPT-5 gave an approximate five-minute forecast before choosing an action, whereas \texttt{gpt-oss-20b} often relied on unsupported guesses and sometimes predicted effects in the wrong direction. Because reported reasoning need not reflect either model's internal computation, we examined the same difference with a separate, machine-scored counterfactual test. The initial test assessed whether the models could express action-conditioned transition knowledge as numerical predictions without generating explanations. GPT-5 used minimal reasoning effort, and thinking was disabled for \texttt{gpt-oss-20b}. All three conditions produced zero reasoning tokens, avoiding differences in generated reasoning length in this comparison. This test assessed direct prediction performance, not the models' full reasoning capability.

The test used 12 states from TD3 trajectories. At each saved state, we varied only the outdoor-air damper, coil valve, or fan speed by $\pm 25\%$ of its allowed range around the TD3 action, while keeping the other six commands fixed. The emulator then produced a low-action and high-action five-minute branch, giving 36 action pairs. Each model received the current observation, including the measured one-minute changes in zone temperature and CO$_2$ immediately before the branch point, and returned JSON predictions of temperature and CO$_2$ in all four zones. Mean absolute error (MAE) measures how far the predicted values are from the emulator values. Strict direction accuracy instead measures whether the model correctly predicts which branch produces the larger next temperature or CO$_2$ value. We exclude temperature differences of $0.02\,^{\circ}$C or less and CO$_2$ differences of 2~ppm or less because their signs are not practically meaningful at this test resolution. For an eligible comparison, a prediction is correct only if its high-minus-low difference has the correct sign and exceeds the same magnitude threshold. Each reported actuator--variable pair aggregates up to 48 comparisons (12 action pairs across four zones); the exclusions remove at most five comparisons per pair. Appendix~\ref{app:evaluation-prompts} gives the prompt and model settings.

\begin{figure}[H]
\centering
\includegraphics[width=\textwidth]{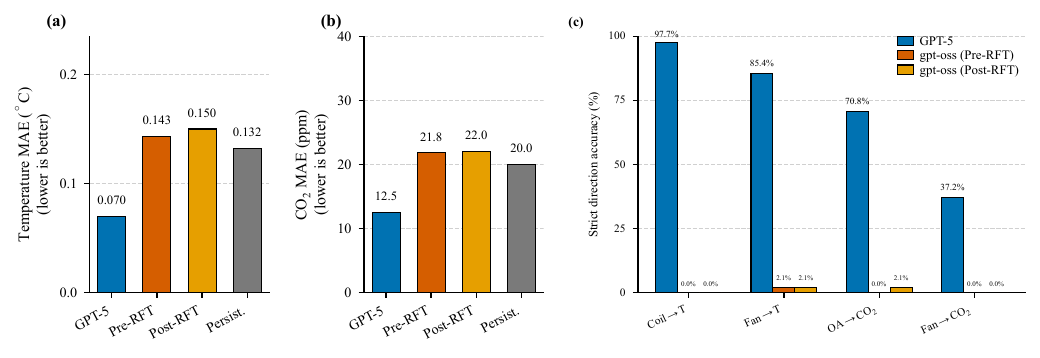}
\caption{Five-minute predictions in the direct-answer setting for 12 states and 36 action pairs. (a,\,b)~Mean absolute error for temperature and CO$_2$; gray bars show the persistence baseline. (c)~Strict direction accuracy for each actuator--variable pair. For direction accuracy, comparisons are excluded when the absolute difference between the two emulator branches is no greater than $0.02\,^{\circ}$C for temperature or 2~ppm for CO$_2$. Pre- and post-RFT refer to \texttt{gpt-oss-20b}.}
\label{fig:transition-evaluation}
\end{figure}

Figure~\ref{fig:transition-evaluation} reports the direct-answer results. The \emph{persistence baseline} simply predicts that the measured temperature and CO$_2$ will remain unchanged after five minutes; outperforming it is a minimal test of whether a model adds useful dynamic information. GPT-5 reached 37.2--97.7\% direction accuracy and had lower mean absolute error than persistence for both variables. \texttt{gpt-oss-20b} reached only 0.0--2.1\% before and after RFT, and its mean errors were higher than persistence. Under this interface, \texttt{gpt-oss-20b} therefore did not express useful local transition predictions, and value-based RFT did not improve them.

In the pre-RFT direct-answer outputs, 169 of 182 errors involved predicted differences at or below the scoring threshold, including 128 numerical ties (Appendix~\ref{app:evaluation-prompts}). With thinking enabled in the generation mode used during RFT, direction accuracy for the same four effects was higher (42.1--60.3\% before RFT and 42.9--58.6\% after RFT). CO$_2$ MAE was lower than that of the persistence baseline, whereas temperature MAE was higher. RFT produced no consistent improvement across prediction metrics (Appendix~\ref{app:evaluation-prompts}).

The absence of consistent gains in transition prediction under either generation setting motivates further study of the training signal: the verifier rewards sampled actions by their long-horizon value but does not provide next-state prediction targets. This result motivates testing transition-prediction SFT before value-based RFT. The emulator can also restart from the same saved state, apply different actions, and automatically score how well a model predicts and plans HVAC behavior.

\section{Discussion}\label{sec:discussion}

The study proceeds in two stages: GPT-5 demonstrates the capability prerequisite, after which rollout-verified RFT tests transfer to a locally deployable open-weight model.

\subsection{Capability of LLM-based VAV control}
\begin{sloppypar}
The five-day evaluation shows an energy--ventilation trade-off rather than uniform superiority of GPT-5. Relative to G36, GPT-5 used less electricity and outdoor air and had lower CO$_2$ compliance, but smaller cumulative CO$_2$ exceedance and a lower peak concentration. TD3 performed best on every temperature and CO$_2$ metric, while GPT-5's temperature performance was close to that of TD3. Compliance is the fraction of occupied zone-minutes within the limit, whereas integrated deviation measures the magnitude and duration of exceedances. These indicators therefore need to be considered together when interpreting energy savings; lower electricity use alone does not establish superior overall control performance.
\end{sloppypar}

\begin{sloppypar}
The final-day deviations of the G36-based controller are consistent with the challenge of maintaining effective tuning across changing load conditions discussed in Section~\ref{sec:introduction}. Appendix~\ref{app:g36-final-day} presents the trajectories and discusses their interpretation without attributing the deviations to a specific parameter.
\end{sloppypar}

\begin{sloppypar}
GPT-5 is not a matched baseline but a capability reference. Given only a textual equipment description and BEMS observations---with no building-specific weight updates, no building-specific system-identification stage, and no explicit numerical online optimizer---this general-purpose reasoning model coordinated seven coupled continuous actuators over the five-day evaluation. It used less total electricity than a tuned high-performance rule sequence, with less outdoor-air intake and lower CO$_2$ compliance. This result has two implications. First, the textual interface is not the limiting factor: from the same observations, GPT-5 both controlled the system and predicted its five-minute responses better than persistence (Section~\ref{sec:transition-evaluation}), so the information content of the interface cannot by itself explain the \texttt{gpt-oss-20b} failures. Second, it reframes the negative transfer result of Section~\ref{subsec:rft-training-results}: what fails there is the current open-weight model under the present training signal, not the feasibility of LLM-based VAV control. Constraint handling still depends on the specific model, prompt, and objective. The open problem is how to develop comparable coordination in a small, locally deployable model that reliably meets operating constraints.
\end{sloppypar}

\smallskip
\begin{sloppypar}
\noindent\textbf{Deployment considerations.} Safety filters have been used to check RL actions in building energy control \cite{wang2025_safe_building_rl}. In practice, an LLM could propose commands under an independent safety supervisor, with a controller validated for the site available as fallback. The supervisor would check sensor validity and freshness, screen commands against equipment and operating limits, and compare commanded and measured actuator responses. Invalid or late responses, communication failures, and detected faults would trigger fallback or equipment protection. Temperature and CO$_2$ monitoring would provide further overrides. LLMs can generate fluent but factually incorrect text \cite{manakul2023_selfcheckgpt}, so a generated explanation should not serve as the safety check. Cybersecurity measures would restrict command access, validate incoming data, and log inputs and actions. These safeguards were not evaluated in this study. Local inference could reduce dependence on external services, keep building data on site, and allow a fixed model version. These potential benefits motivate RFT for a smaller model, but do not establish its control performance or safety.
\end{sloppypar}

\subsection{Verifier accuracy and sampled actions}
Transferring TD3-derived control knowledge through RFT requires three links to work: the model must construct useful action candidates from the prompt, the verifier must rank those candidates correctly, and the optimizer must make the better candidates more probable without damaging other behavior. The critic audit in Section~\ref{subsec:verifier-selection} tests the middle link. Pooled over all candidates, critic estimates correlated with rollout returns at $r=0.9995$, but differences between states dominate this value. In comparisons within each state, as required by group-relative RFT, the critic selected the rollout-best candidate in only 7 of 30 states. Many LLM actions also lay outside the TD3 reference distribution, where critic estimates require extrapolation. Direct rollout avoids this critic error for every sampled action and makes the RFT result easier to interpret, but it does not guarantee the other two links.

The audit also yields a recommendation that applies beyond this study. Wherever a learned value model is proposed as the reward model for group-relative fine-tuning, aggregate correlation with returns is the wrong acceptance test: the quantity to audit is the within-state ranking of actions sampled from the policy that will be trained, checked against ground-truth returns before any fine-tuning budget is spent. Here, the pilot audit used 160 generations and their rollouts and settled the verifier choice; the 30-state audit used 480 generations.

\begin{sloppypar}
We evaluated the pre- and post-RFT controllers in closed-loop operation. Over the five evaluation days, their electricity use was 440.70 and 443.64~kWh, respectively, compared with 417.18~kWh for the G36-based baseline: increases of 5.64\% and 6.34\%. Temperature compliance with the 25--27~$^{\circ}$C band was 82.96\% before RFT and 79.84\% after RFT; CO$_2$ compliance with the 1000~ppm threshold was 88.23\% and 87.53\%, respectively. Each pre/post condition represents one evaluation run, not a repeated-run average. Thus, neither electricity use nor these two compliance rates improved in this evaluation. Appendix~\ref{app:state-distribution} compares the 30 training prompts with the closed-loop observations over all control times.
\end{sloppypar}

\begin{sloppypar}
The distribution comparison shows that the temperature 5th percentile falls from 25.8~$^{\circ}$C in the training prompts to 20.9 and 21.0~$^{\circ}$C in pre- and post-RFT operation, while the median CO$_2$ concentration rises from 918~ppm to 945 and 947~ppm, respectively (Table~\ref{tab:observation-quantiles}). These shifts document differences between the fixed training sample and deployment observations, including after RFT. Figure~\ref{fig:rollout_gap_overlay} also shows no sustained improvement in sampled-action returns on the teacher-derived training states. The results therefore do not show a demonstrated training-state gain that was subsequently lost at deployment. Training returns and closed-loop metrics differ in both continuation policy and performance criterion, and these comparisons do not isolate the effect of distribution shift on deployment performance. Iterative student-policy state collection and candidate rescoring with LLM continuation were not performed and remain untested extensions.
\end{sloppypar}

The first link is candidate construction. In the earlier TES task, exact dynamic-programming values were available for every action in a small discrete set. Even before RFT, the best of ten sampled actions produced a near-optimal schedule; RFT mainly made the model choose such actions more often \cite{shioda2026_tes_rft}. In the present seven-dimensional continuous space, we generate only 16 candidate responses per prompt group. The best VAV action sometimes outperforms TD3, but its moving average does not remain at TD3 performance, and the mean sampled-action return stays below TD3 throughout training. A reliable verifier can rank the sampled actions, but it does not explicitly supply the action-conditioned transition knowledge needed to construct a coordinated action outside the sampled group. The size of the action set is therefore one difference between the two tasks; Section~\ref{subsec:action-space-direction} argues that it is not the only one.

\subsection{Value supervision and transition knowledge}
Even an exact action value is a single number summarizing a trajectory. It is \emph{evaluative} feedback in the classical sense: it reports how well the complete action vector performed under the continuation, not which intermediate prediction or actuator choice produced that return \cite{sutton2018rl}. Equation~\eqref{eq:drgrpo-objective} makes the resulting credit-assignment problem explicit: the verifier scores only the parsed seven-dimensional action, yet every token in response $j$ is optimized under the same completion-level advantage $\hat{A}_j$. An above-average response is reinforced in its entirety, including any incorrect transition claim or poorly chosen actuator component it contains; a below-average response is suppressed in its entirety, including any useful reasoning. A model can therefore receive exact value supervision without learning the local dynamics needed to construct and compare new actions.

This account has a testable implication: if completion-level value supervision does not teach local dynamics, transition-prediction accuracy should not improve during RFT. The counterfactual test in Section~\ref{sec:transition-evaluation} shows this pattern: the direct-answer direction accuracy of \texttt{gpt-oss-20b} remains at 0.0--2.1\% before and after the 200 RFT steps, while unadapted GPT-5 predicts the same branches far better. Under the evaluated protocol, exact scores did not yield consistent improvements in either sampled-action returns or transition prediction.

The transition results are consistent with this explanation, but they do not establish why RFT failed to improve performance. Similar observations can correspond to different hidden states and lead to different transitions, so partial observability may contribute to prediction errors. GPT-5 predicted the same transitions more accurately from the same observations, so the inputs support useful prediction. Because we did not compare predictions with and without access to the full state, we cannot separate the effect of hidden states from limitations in the model's predictive ability. The test therefore measures prediction from the supplied observations, rather than physical knowledge alone. Limited exploration, reward concentration, optimization settings, and model capacity may also have contributed to the RFT outcome; their effects were not isolated.

\subsection{From action values to action-space direction}\label{subsec:action-space-direction}
The verifier reports which sampled action scored better, not which of the seven commands to move or in which direction. Expanding Eq.~\eqref{eq:state-q-definition} by one decision interval makes the missing quantity explicit. With $s'=f(s,a,d_t)$ the state reached after the candidate interval,
\begin{equation}\label{eq:action-space-direction}
\frac{\partial Q^{\mu}_{t,d}(s,a)}{\partial a}
= \frac{\partial r_t(s,a)}{\partial a}
+ \gamma\,\frac{\partial V^{\mu}_{t+1,d}(s')}{\partial s'}\,\frac{\partial s'}{\partial a}.
\end{equation}
The transition-mediated part of this direction therefore factors into a preference over next states and the action-to-state Jacobian $\partial s'/\partial a$, which describes how each actuator locally changes the next state. The rollout verifier evaluates $Q^{\mu}_{t,d}$ exactly at sampled points but supplies neither factor, and the reported procedure computes no gradient in action space. For a fixed sampling policy and budget, group-relative RFT can redistribute probability only over sampled actions; even an exact verifier cannot select a candidate better than the best one available in that set \cite{mroueh2025rlvr}. In the reported continuous-control setting, exact action values therefore do not by themselves guarantee improvement: they rank sampled candidates but do not enlarge policy support or provide a direction for constructing better ones.

Different learning systems obtain this directional content by different routes. Deterministic policy gradients update the actor through an action-differentiable critic \cite{silver2014dpg}. TD3 uses this mechanism with delayed actor updates \cite{ref:fujimoto2018_td3}, supplying $\partial Q/\partial a$ for all seven commands at each actor update. Model-based RL uses the other factor instead: Dreamer trains its actor by backpropagating return gradients through a learned differentiable dynamics model \cite{hafner2020dreamer}, and DayDreamer applies that mechanism to continuous control on physical robots \cite{wu2023_daydreamer}. The present pipeline has neither route. It receives scalar returns only at sampled action vectors, so the second factor must come from the model's own knowledge. Section~\ref{sec:transition-evaluation} measures precisely that: varying one actuator while the other six are held fixed probes the sign of one entry of $\partial s'/\partial a$, and the four bars in Figure~\ref{fig:transition-evaluation}(c) are four such entries. GPT-5 recovers their sign in 37.2--97.7\% of pairs and states the corresponding effects in words before choosing commands (Appendix~\ref{app:gpt5-forecast-example}); \texttt{gpt-oss-20b} recovers them in 0.0--2.1\% before and after RFT in the direct-answer setting. The two models differ in size, post-training, and access mode, so the pairing illustrates this explanation without isolating which property produces the difference.

The earlier TES task did not require the model to infer this Jacobian: its additive storage balance made $\partial s'/\partial a$ fixed, so forward prediction reduced to arithmetic and RFT could reinforce a procedure whose prediction and constraint-checking steps were already executable \cite{shioda2026_tes_rft}. In VAV, the same step depends on the coupled nonlinear response of the air network, cooling coil, and water circuit. GPT-5 executes that step before committing to commands (Appendix~\ref{app:gpt5-forecast-example}); \texttt{gpt-oss-20b} reaches it and abandons it (Appendix~\ref{app:gptoss-reasoning-example}). Value-based RFT may therefore reinforce a procedure without directly supplying the transition knowledge that one of its steps requires. Because the two tasks also differ in action set and verifier, this contrast is not a controlled comparison.

\emph{Transition SFT} would supply this information directly by training on action-conditioned state changes generated through the same save-and-branch interface; related work fine-tunes language models on embodied experiences collected in a simulator \cite{xiang2023_language_world_models}, while another approach separately trains models to predict action preconditions and effects \cite{xie2025_llm_world_model}. A direct next test should compare the base model, RFT only, transition SFT only, and transition SFT followed by RFT, reporting transition error and direction accuracy alongside sampled-action quality and closed-loop control metrics. This comparison would test whether transition knowledge limited the present run and whether subsequent value-based RFT preserves it.

\subsection{Limitations}\label{subsec:limitations}

\smallskip
\noindent\textbf{Exact-score claim.} The rollout return is exact only for the deterministic emulator, saved full state, fixed future disturbances, stated reward, and TD3 continuation used here. Exactness concerns score computation, not whether the reward represents the best control objective. The action-distance threshold used in the critic audit is diagnostic rather than a formal distribution boundary.

\smallskip
\noindent\textbf{Empirical scope.} The controller comparison covers one uncalibrated four-zone emulator in one climate over five consecutive summer weekdays, including days with different occupancy schedules and one cooler, more humid day. Other seasons, buildings, and climates were not evaluated. The reported energy differences are specific to this emulator. Our aim was to evaluate LLM control and rollout-verified RFT, which guided our choice of baselines. G36 provides a rule-based reference for control performance, while TD3 provides both a learned baseline and the continuation policy for RFT. We therefore focused on these two baselines and did not extend the comparison to MPC. GPT-5 and \texttt{gpt-oss-20b} share an interface but are otherwise unmatched, so their difference cannot be attributed to a particular model property or generalized to reasoning models as a class.

\smallskip
\noindent\textbf{Energy--ventilation trade-off.} The evaluation did not isolate the effects of reduced outdoor-air intake from those of other control actions, so their respective contributions to the energy savings remain unresolved.

\smallskip
\noindent\textbf{Negative transfer and transition evidence.} The negative result is limited to one 200-step, fixed-seed RFT run and its reported optimization and sampling settings. We did not repeat the run with other random seeds or vary the candidate-group size, training duration, learning rate, LoRA rank, reward temperature, or KL regularization. The absence of improvement therefore does not distinguish limitations of RFT itself from insufficient exploration, training budget, or other optimization choices; larger candidate groups or longer training may yield different outcomes. The direct-answer and thinking-enabled transition tests showed no consistent improvement after RFT under the evaluated settings.

\smallskip
\begin{sloppypar}
\noindent\textbf{Deployment and generality.} No field experiments were conducted, so this emulator study does not establish safe deployment or cross-building generality. Field validation under model error, uncertain disturbances, faults, and previously unseen building systems remains necessary. Application in other regions would require validation with local weather, occupancy schedules, equipment, and comfort and ventilation requirements.
\end{sloppypar}

\section{Conclusion}\label{sec:conclusion}
\begingroup
\sloppy
This study examined whether a large language model can control a four-zone variable-air-volume system through text and whether rollout-verified reinforcement fine-tuning can improve a smaller model's control. GPT-5 coordinated seven continuous actuators without building-specific weight updates, providing a capability reference for a smaller, locally deployable model. Across three five-day runs, it used 6.8--7.4\% less electricity than a Guideline~36-based controller, with lower outdoor-air intake and lower CO$_2$ compliance. A reinforcement learning controller reduced electricity use by 4.5\% while improving temperature and CO$_2$ performance. GPT-5's electricity savings therefore do not establish better overall control.

The verifier audit showed that high correlation with rollout returns did not ensure correct action rankings. Across 470 candidate actions, estimates from the reinforcement learning controller's critic correlated with rollout returns at $r=0.9995$, but the critic selected the best candidate in only 7 of 30 states. For fine-tuning that compares candidates at the same state, a learned verifier should therefore be evaluated on its rankings within each state.

With direct rollout scores, one 200-step fine-tuning run showed no sustained gain in the returns of sampled actions and did not improve closed-loop control. Analysis of the model's reasoning traces and quantitative tests showed that \texttt{gpt-oss-20b} did not reliably predict how its actions would change temperature and CO$_2$, either before or after fine-tuning.

These findings are limited to one emulator and the evaluated training configuration. They do not establish poor next-state prediction as the cause of the training outcome. An open question is whether the ability to predict action effects is a prerequisite for effective rollout-verified reinforcement fine-tuning in physical control. A next step is to test whether supervised fine-tuning on next-state prediction enables subsequent rollout-verified reinforcement fine-tuning to improve control, across different training settings, buildings, and operating conditions.
\par\endgroup

\section*{CRediT authorship contribution statement}
\textbf{Takumi Shioda:} Conceptualization, Methodology, Software, Validation, Formal analysis, Investigation, Data curation, Writing - original draft, Writing - review \& editing, Visualization.
\textbf{Kohei Terashima:} Writing - review \& editing.
\textbf{Tatsuo Nagai:} Supervision, Writing - review \& editing.

\section*{Funding}
This research did not receive any specific grant from funding agencies in the public, commercial, or not-for-profit sectors.

\section*{Declaration of competing interests}
The authors declare that they have no known competing financial interests or personal relationships that could have appeared to influence the work reported in this paper.

\section*{Data availability}
Data will be made available on request.

\section*{Declaration of generative AI and AI-assisted technologies in the manuscript preparation process}
During the preparation of this work, the authors used OpenAI ChatGPT to improve the English language and readability of the manuscript. After using this tool/service, the authors reviewed and edited the content as needed and take full responsibility for the content of the published article.

\clearpage
\appendix
\counterwithin{equation}{section}
\counterwithin{figure}{section}
\counterwithin{table}{section}
\counterwithin{algocf}{section}
\setcounter{topnumber}{3}
\setcounter{dbltopnumber}{3}
\setcounter{totalnumber}{6}
\renewcommand{\topfraction}{0.9}
\renewcommand{\dbltopfraction}{0.98}
\renewcommand{\textfraction}{0.02}
\renewcommand{\floatpagefraction}{0.85}

\section{Emulator and training configurations}\label{app:configurations}

\subsection{VAV emulator formulation and parameters}\label{app:emulator-parameters}
This appendix gives the component dynamics omitted from Section~\ref{subsec:simulator}. The coil-load and power equations used by the reward are given in the main text as Eqs.~\eqref{eq:coil-load} and \eqref{eq:system-power}.

\begin{figure}[H]
\centering
\begin{minipage}[t]{0.48\textwidth}
\centering
\includegraphics[width=\linewidth]{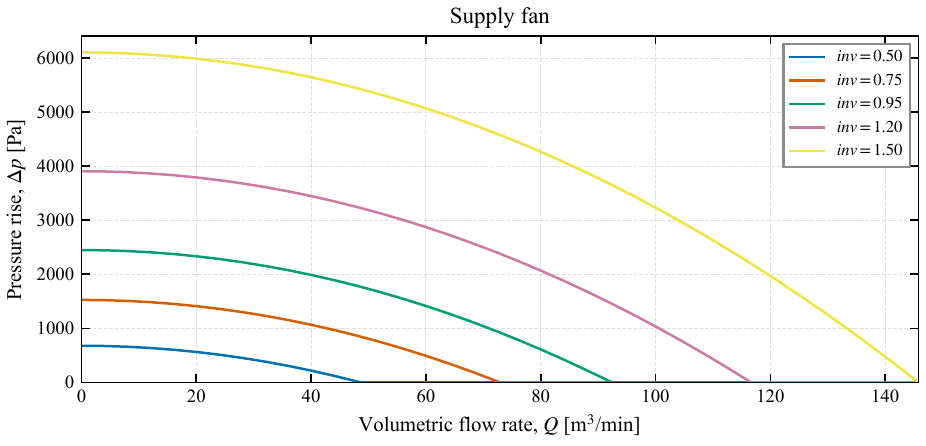}

{\small (a) Supply fan}
\end{minipage}\hfill
\begin{minipage}[t]{0.48\textwidth}
\centering
\includegraphics[width=\linewidth]{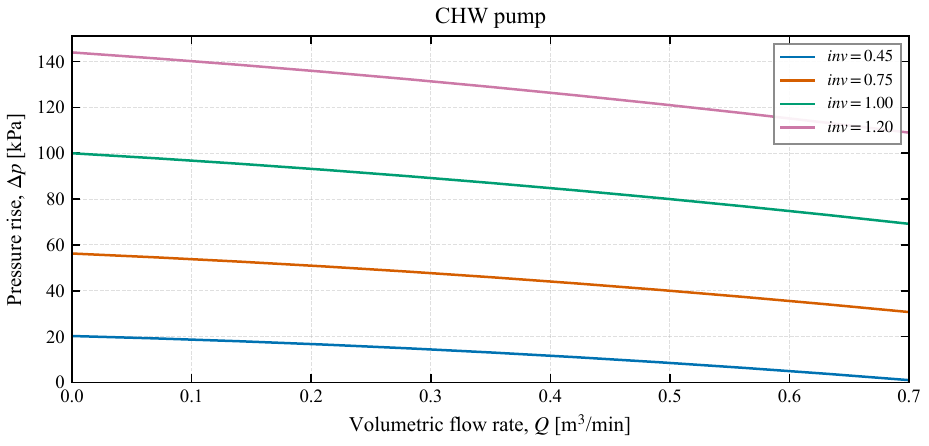}

{\small (b) Chilled-water pump}
\end{minipage}
\caption{Pressure--flow (P--Q) curves of the fan and pump models at several speed ratios.}
\label{fig:pq_curves}
\end{figure}

\paragraph{Air-side network.}
The supply duct is represented as a network of nodes and segments. At each junction $v$, the segment flows satisfy continuity. Each segment $e$ has a Darcy--Weisbach pressure loss:
\begin{equation}\label{eq:air-network}
\begin{gathered}
\sum_{e\in\mathcal{E}^{\mathrm{in}}_v}G_e
=\sum_{e\in\mathcal{E}^{\mathrm{out}}_v}G_e, \\
\Delta P_e
=\frac{\rho}{2}
\left(\frac{G_e}{60A_e}\right)^{\!2}
\left(\frac{f_eL_e}{D_e}+\sum_m\zeta_{e,m}\right)
=K_eG_e^2,
\end{gathered}
\end{equation}
where $G_e$ is in m$^3$/min; $A_e$, $L_e$, and $D_e$ describe the segment geometry; $f_e$ is the friction factor; and $\zeta_{e,m}$ are local-loss coefficients. The coefficients $K_e$ are computed from geometry rather than fitted to data. For zone branch $i$,
\begin{equation}\label{eq:air-flow-conversion}
\dot m_i=\frac{\rho G_i}{60},
\qquad
q_i=\frac{\dot m_i}{\rho}=\frac{G_i}{60}.
\end{equation}
Each branch adds a fixed loss and a damper loss $K_{\mathrm{dmp}}(a_i^{\mathrm{eff}})G_i^2$, with
\begin{equation}\label{eq:damper-floor}
a_i^{\mathrm{eff}}=\max\!\left(a_i^{\mathrm{vav}},a_i^{\mathrm{floor}}\right),
\qquad a_i^{\mathrm{floor}}=0.40.
\end{equation}
The solver updates branch flows and node pressures until Eq.~\eqref{eq:air-network} is satisfied. An outer bisection matches the fan pressure to the network pressure:
\begin{equation}\label{eq:fan-pressure-closure}
\begin{gathered}
\Delta P_{\mathrm{fan}}(G,u)
=\Delta P_0u^2
\left[1-\left(\frac{G}{G_0u}\right)^2\right], \\
0
=\Delta P_{\mathrm{fan}}
-\Delta P_{\mathrm{coil}}
-\Delta P_{\mathrm{return}}
-\Delta P_{\mathrm{network}}.
\end{gathered}
\end{equation}
Figure~\ref{fig:pq_curves} shows the fan and pump pressure--flow characteristics.

\begin{table}[H]
\centering
\caption{Main scalar parameters of the VAV emulator.}
\label{tab:emulator-parameters}
\footnotesize
\begin{tabular}{@{}>{\raggedright\arraybackslash}p{0.27\textwidth}>{\raggedright\arraybackslash}p{0.17\textwidth}>{\raggedright\arraybackslash}p{0.27\textwidth}>{\raggedright\arraybackslash}p{0.17\textwidth}@{}}
\toprule
Item & Setting & Item & Setting \\
\midrule
Simulation time step $\Delta t$ & 60~s & Coil dry--wet boundary surface humidity & 95\% \\
Air density $\rho$ & 1.2~kg/m$^3$ & Coil $\mathrm{UA}$ & 3500~W/K \\
Branch design static pressure & 700~Pa & Chilled-water supply temperature & $7.0\,^{\circ}$C \\
Design damper opening / authority & 0.60 / 0.55 & Coil bypass factor / minimum approach & 0.10 / $1.0\,^{\circ}$C \\
Effective VAV opening floor $a_i^{\mathrm{floor}}$ & 0.40 & Coil-valve filter coefficient $\alpha_v$ & 0.25 \\
Peak fan / motor efficiency & 0.62 / 0.88 & Pump efficiency $\eta_{\mathrm p}$ / chiller COP & 0.80 / 4.0 \\
Outdoor CO$_2$ concentration $C_{\mathrm{oa}}$ & 420~ppm & Zone volume $V_i$ & 250~m$^3$ \\
CO$_2$ generation $g_{\mathrm{CO_2}}$ & 0.0055~L/(s person) & Occupant latent gain & 0.06~kW/person \\
\bottomrule
\end{tabular}
\end{table}

\begin{table}[H]
\centering
\caption{Zone parameters used in the four-zone VAV emulator.}
\label{tab:zone_params}
\small
\begin{tabular}{lrrrrrrrr}
\toprule
Zone & $C_i^{\mathrm{th}}$ [J/K] & $U_i$ [W/K] & $\dot m_{\mathrm{inf},i}$ [kg/s] & $\dot m_{\mathrm{sa},i}^{\mathrm{des}}$ [kg/s] & $a_i^{\mathrm{floor}}$ [--] & $Q_{\mathrm{int},i}^{\mathrm{day}}$ [W] & $Q_{\mathrm{int},i}^{\mathrm{night}}$ [W] & $T_i(t_0)$ [$^{\circ}$C] \\
\midrule
1 & $3.2\times 10^{6}$ & 170 & 0.018 & 0.42 & 0.40 & 3200 & 300 & 28.0 \\
2 & $2.8\times 10^{6}$ & 150 & 0.016 & 0.36 & 0.40 & 2800 & 280 & 27.5 \\
3 & $2.6\times 10^{6}$ & 145 & 0.015 & 0.34 & 0.40 & 2500 & 250 & 27.2 \\
4 & $3.0\times 10^{6}$ & 160 & 0.017 & 0.32 & 0.40 & 3000 & 300 & 28.3 \\
\bottomrule
\end{tabular}
\end{table}

\paragraph{Mixing, coil, and chilled-water circuit.}
Let $\boldsymbol{z}=(h,w,C)$ collect moist-air enthalpy, humidity ratio, and CO$_2$ concentration. The outdoor-air command is converted to mass flow subject to the ventilation minimum:
\begin{equation}\label{eq:outdoor-air-flow}
\begin{gathered}
\dot m_{\mathrm{oa}}
=\operatorname{clip}\!\left(
a^{\mathrm{oa}}\dot m_{\mathrm{a}},
\dot m_{\mathrm{oa}}^{\min},
\dot m_{\mathrm{a}}
\right), \\
\alpha_{\mathrm{oa}}=\frac{\dot m_{\mathrm{oa}}}{\dot m_{\mathrm{a}}}.
\end{gathered}
\end{equation}
Return air is the flow-weighted mixture of the zone exhausts, and mixed air combines the return and outdoor streams:
\begin{equation}\label{eq:air-mixing}
\begin{gathered}
\boldsymbol{z}_{\mathrm{ra}}
=\frac{\sum_{i=1}^{n}\dot m_{\mathrm{r},i}\boldsymbol{z}_i}
{\sum_{i=1}^{n}\dot m_{\mathrm{r},i}}, \\
\boldsymbol{z}_{\mathrm{ma}}
=\alpha_{\mathrm{oa}}\boldsymbol{z}_{\mathrm{oa}}
+(1-\alpha_{\mathrm{oa}})\boldsymbol{z}_{\mathrm{ra}}, \\
T_{\mathrm{ma}}
=\frac{h_{\mathrm{ma}}-r_0w_{\mathrm{ma}}}
{c_{pa}+c_{pv}w_{\mathrm{ma}}}.
\end{gathered}
\end{equation}
Here $\dot m_{\mathrm{r},i}$ is the return-air mass flow from zone $i$, and $\alpha_{\mathrm{oa}}$ is the outdoor-air mass fraction produced by the linked outdoor-, return-, and exhaust-air dampers.

The cooling coil is a water--air heat exchanger solved with effectiveness--NTU relations. The model switches between dry and wet operation according to the coil-surface humidity. Its sensible and latent loads form $Q_{\mathrm{tot}}$ in Eq.~\eqref{eq:coil-load}. On the water side, a first-order filter is applied to the valve command. Pump speed follows the square root of the filtered opening, and the chilled-water flow satisfies the pump--circuit pressure balance:
\begin{equation}\label{eq:chw-circuit}
\begin{gathered}
\bar v_{t+1}
=(1-\alpha_v)\bar v_t+\alpha_vv^{\mathrm{coil}}_{k(t)},
\qquad \alpha_v=0.25, \\
u_{\mathrm{p},t}
=u_{\mathrm{p}}^{\min}
+\left(u_{\mathrm{p}}^{\max}-u_{\mathrm{p}}^{\min}\right)
\sqrt{\bar v_t}, \\
0
=\Delta P_{\mathrm{pump}}(\dot V_{\mathrm{chw},t},u_{\mathrm{p},t})
-\Delta P_{\mathrm{circuit}}(\dot V_{\mathrm{chw},t},\bar v_t).
\end{gathered}
\end{equation}
The flow balance is solved by bisection. Equation~\eqref{eq:chw-circuit} links the coil-valve command to water flow and, through Eqs.~\eqref{eq:coil-load} and \eqref{eq:system-power}, to the supply-air state and total power.

\paragraph{Zone balances.}
Each zone is a well-mixed lumped-capacity volume with temperature, CO$_2$, and humidity-ratio states. The sensible heat balance is
\begin{equation}\label{eq:zone-temperature}
\begin{aligned}
C_i^{\mathrm{th}}\frac{dT_i}{dt}
&= U_i\,(T_{\mathrm{oa}}-T_i)\\
&\quad+1000\,c_{pa}\,\dot m_{\mathrm{inf},i}
  (T_{\mathrm{oa}}-T_i)\\
&\quad+1000\,c_{pa}\,\dot m_{\mathrm{sa},i}
  (T_{\mathrm{sa}}-T_i)
+ Q_{\mathrm{int},i}.
\end{aligned}
\end{equation}
The occupancy-dependent CO$_2$ source and concentration balance are
\begin{equation}\label{eq:zone-co2}
\begin{gathered}
G_{\mathrm{CO_2},i}=10^{-3}N_i g_{\mathrm{CO_2}}
\quad [\mathrm{m^3/s}], \\
\frac{dC_i}{dt}
= \frac{q_{\mathrm{sa},i}}{V_i}\,(C_{\mathrm{sa}}-C_i)
+ \frac{q_{\mathrm{inf},i}}{V_i}\,(C_{\mathrm{oa}}-C_i)
+ \frac{G_{\mathrm{CO_2},i}}{V_i}\times 10^{6}.
\end{gathered}
\end{equation}
The humidity-ratio balance is
\begin{equation}\label{eq:zone-humidity}
\begin{aligned}
\rho V_i\frac{dw_i}{dt}
&=\dot m_{\mathrm{sa},i}(w_{\mathrm{sa}}-w_i)\\
&\quad+\dot m_{\mathrm{inf},i}(w_{\mathrm{oa}}-w_i)
+\dot m_{w,i}^{\mathrm{int}},\\
\dot m_{w,i}^{\mathrm{int}}
&=\frac{Q_{\mathrm{lat},i}^{\mathrm{int}}}{r_0}.
\end{aligned}
\end{equation}
These balances use explicit Euler integration, $\xi_{i,t+1}=\xi_{i,t}+\Delta t\,\dot\xi_{i,t}$ for $\xi_i\in\{T_i,w_i,C_i\}$. The design mass flow $\dot m_{\mathrm{sa},i}^{\mathrm{des}}$ sets the branch resistance; it is not a hard flow limit. The pressure-network solution determines the actual $\dot m_{\mathrm{sa},i}$.

\paragraph{Sequential computation.}
The minute-level transition follows a fixed sequence:
\begin{equation}\label{eq:emulator-sequence}
\begin{aligned}
(\mathbf{G}_t,\mathbf{p}_t)
&=\mathcal{N}(x_t,a_{k(t)}), \\
\boldsymbol{z}_{\mathrm{ma},t}
&=\mathcal{M}(\mathbf{G}_t,x_t,d_t), \\
(\boldsymbol{z}_{\mathrm{sa},t},P_{\mathrm{fan},t},
P_{\mathrm{pump},t},P_{\mathrm{ch},t})
&=\mathcal{E}(\boldsymbol{z}_{\mathrm{ma},t},\mathbf{G}_t,x_t,a_{k(t)}), \\
x_{t+1}
&=\mathcal{U}_{\Delta t}(x_t,\boldsymbol{z}_{\mathrm{sa},t},
\mathbf{G}_t,d_t).
\end{aligned}
\end{equation}
Here, $\mathcal{N}$ solves the air network, $\mathcal{M}$ computes the return and mixed-air states, $\mathcal{E}$ solves the fan, chilled-water circuit, and coil, and $\mathcal{U}_{\Delta t}$ updates the zone and internal states. The operating schedule and weather enter through $d_t$, and the five-minute action $a_{k(t)}$ remains fixed during the one-minute updates.

Table~\ref{tab:emulator-parameters} lists the main scalar parameters. Table~\ref{tab:zone_params} gives the values for each zone.

\subsection{TD3 configuration}\label{app:td3-configuration}
Table~\ref{tab:td3_hparams} lists the TD3 training hyperparameters.

\noindent\begin{minipage}{\columnwidth}
\castblcaption
\sffamily\small
\captionof{table}{Hyperparameters used to train the TD3 teacher on the four-zone VAV emulator.}
\label{tab:td3_hparams}
\centering
\footnotesize
\begin{tabular}{@{}p{0.55\columnwidth}p{0.38\columnwidth}@{}}
\toprule
Item & Setting \\
\midrule
Training episodes & 10{,}000 \\
Discount factor $\gamma$ & 0.995 \\
Actor / critic learning rate & $3\times 10^{-4}$ / $3\times 10^{-4}$ \\
Target soft-update rate $\tau_{\mathrm{soft}}$ & 0.005 \\
Batch size / replay buffer & 512 / 200{,}000 \\
Warm-up steps & 5{,}000 \\
Target policy smoothing (std / clip) & 0.2 / 0.5 \\
Policy update delay & 2 \\
Exploration noise (std, $\tanh$ space) & 0.1 \\
Actor / critic hidden layers & (256, 256, 128) \\
Observation normalization clip & 5.0 \\
\bottomrule
\end{tabular}
\end{minipage}
\medskip

\subsection{RFT and LoRA configuration}\label{app:rft-configuration}
Table~\ref{tab:vav-rft-hparams} lists the model-loading, optimizer, sampling, memory, and software settings for the reported run. Settings not listed in the run script use the defaults of the reported library versions.

\begin{table}[H]
\centering
\caption{RFT and LoRA configuration for the reported VAV run.}
\label{tab:vav-rft-hparams}
\footnotesize
\begingroup
\renewcommand{\arraystretch}{1.08}
\begin{tabular}{@{}p{0.34\textwidth}p{0.60\textwidth}@{}}
\toprule
Item & Setting \\
\midrule
Base model & \texttt{openai/gpt-oss-20b} \\
Model loading / arithmetic & MXFP4 checkpoint dequantized to bfloat16 / bfloat16 training \\
Attention kernel & FlashAttention~3 via \texttt{kernels-community/vllm-flash-attn3} \\
Trainable parameters & LoRA adapters only; base-model weights frozen \\
LoRA targets & PEFT \texttt{all-linear} (linear and Conv1D modules; output head excluded) \\
LoRA rank $r$ / scaling $\alpha$ & $4$ / $8$ ($\alpha/r=2$) \\
LoRA dropout / bias & $0$ / none \\
Policy objective & Dr.~GRPO with fixed $L_{\max}$; Liger fused loss \\
Reward normalization & Group-mean centering; no division by reward standard deviation \\
Policy iterations / clip $\epsilon$ / KL $\beta$ & $1$ / $0.2$ / $0$ (no reference-model penalty) \\
Optimizer & Fused PyTorch AdamW \\
Learning rate / schedule / warm-up & $5\times10^{-5}$ / linear decay / 0 steps \\
AdamW $(\beta_1,\beta_2)$ / $\varepsilon$ / weight decay & $(0.9,0.999)$ / $10^{-8}$ / $0$ \\
Maximum gradient norm & $1.0$ \\
Optimizer steps & $200$ \\
Group size $G$ / prompts per micro-step & $16$ / $1$ \\
Per-device sequence batch / accumulation $K$ & $16$ / $4$ micro-steps \\
Effective responses per optimizer step & $64$ (four prompt groups $\times$ 16 responses) \\
Unique prompts / total generated responses & $30$ / $12{,}800$ \\
Maximum prompt / completion length & $1{,}400$ / $4{,}000$ tokens \\
Sampling & Stochastic; temperature $1.0$; top-$p=1.0$; no top-$k$ or min-$p$ filter; repetition penalty $1.0$ \\
Verifier reward & Direct rollout; softmax temperature $\tau=0.05$ \\
Verifier discount $\gamma$ & $0.995$ per decision interval (matches TD3 training) \\
Invalid / truncated outputs & Invalid or unparsable responses masked; completions truncated at $L_{\max}$ masked \\
Generation / memory settings & Transformers generation (no vLLM); gradient checkpointing; Liger loss \\
Random seed / checkpoint interval & $42$ / 20 optimizer steps \\
Software & PyTorch 2.8.0; Transformers 4.57.1; TRL 0.23.1; PEFT 0.17.1 \\
\bottomrule
\end{tabular}
\endgroup
\end{table}

\clearpage
\subsection{Training algorithm with rollout verification}\label{app:rft-algorithm}
Algorithm~\ref{alg:vav_rollout_rft} gives the complete training procedure.

\begin{algorithm}[H]
\footnotesize
\SetAlgoLined
\DontPrintSemicolon
\caption{Rollout-based RFT for VAV control}
\label{alg:vav_rollout_rft}
\KwIn{Prompt-state dataset $\mathcal{D}=\{(x_s,s,t,d_{t:T})\}$, deterministic emulator $F$, frozen TD3 actor $\mu$, policy $\pi_\theta$, action parser $g(\cdot)$, action bounds $\mathcal{A}$, discount $\gamma$, reward temperature $\tau$, group size $G$, accumulation length $K$, completion limit $L_{\max}$, clip $\epsilon$, optimizer steps $N$}
\KwOut{Fine-tuned VAV policy $\pi_\theta$}

Create a reproducible prompt order from $\mathcal{D}$\;
\For{$n \leftarrow 1$ \KwTo $N$}{
  Clear accumulated gradients\;
  \For{$m \leftarrow 1$ \KwTo $K$}{
    Sample one prompt-state tuple $(x_s,s,t,d_{t:T})$\;
    Generate $G$ responses $y^{(1)},\ldots,y^{(G)}$ for $x_s$ with the current LLM policy $\pi_\theta$\;
    Initialize the valid index set $\mathcal{V}\leftarrow\varnothing$\;
    \For{$j \leftarrow 1$ \KwTo $G$}{
      Parse $a^{(j)}\leftarrow g(y^{(j)})$\;
      \eIf{response $y^{(j)}$ is truncated or its parsed action $a^{(j)}$ is invalid}{
        Mask response $j$ from optimization\;
      }{
        Clip $a^{(j)}$ to $\mathcal{A}$\;
        Restore $F$ to $s$, apply $a^{(j)}$ for interval $t$, then follow $\mu$ from $t+1$ to $T$ under $d_{t:T}$\;
        Compute $R_j\leftarrow\sum_{q=0}^{T-t}\gamma^q r_{t+q}$ and set $\mathcal{V}\leftarrow\mathcal{V}\cup\{j\}$\;
      }
    }
    \If{$\mathcal{V}\neq\varnothing$}{
      Convert valid returns to $\tilde r_j=\exp(R_j/\tau)\big/\sum_{\ell\in\mathcal{V}}\exp(R_\ell/\tau)$\;
      Compute centered advantages $\hat A_j=\tilde r_j-|\mathcal{V}|^{-1}\sum_{\ell\in\mathcal{V}}\tilde r_\ell$\;
      Accumulate gradients for one Dr.~GRPO micro-step over the valid responses using Eq.~\eqref{eq:drgrpo-objective}\;
    }
  }
  Apply one optimizer step to update only the LoRA parameters of $\pi_\theta$\;
}
\end{algorithm}

\clearpage
\section{Training and deployment observation distributions}\label{app:state-distribution}
\begin{figure}[H]
\centering
\includegraphics[width=\textwidth]{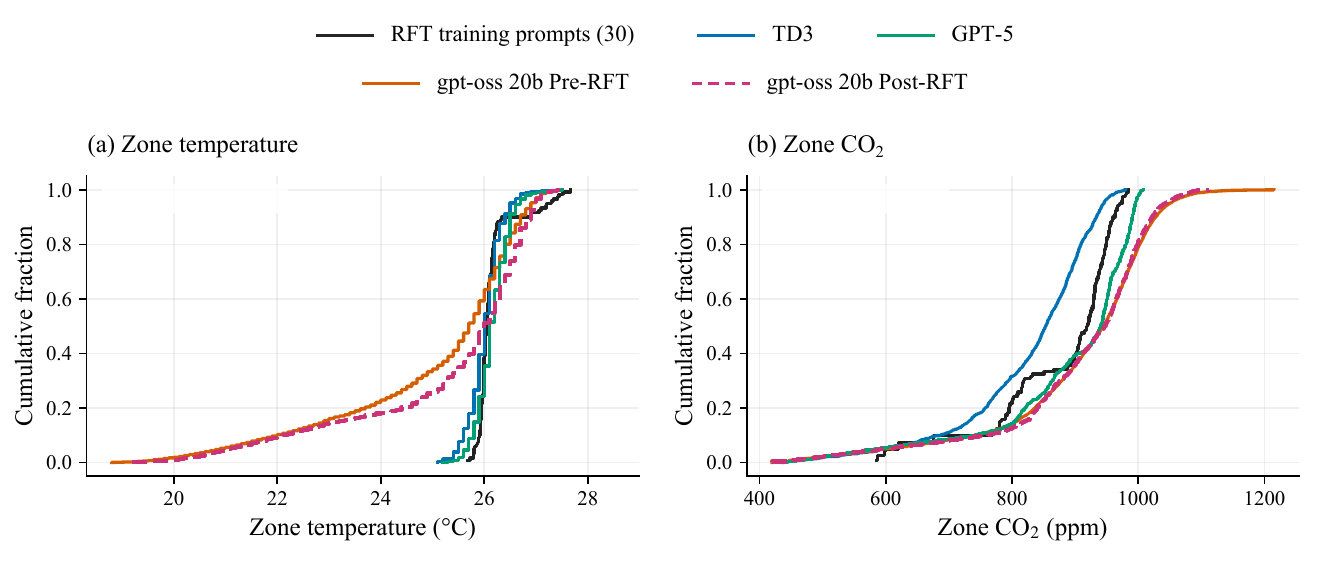}
\caption{Empirical cumulative distributions of (a) zone temperature and (b) zone CO$_2$ concentration. The black curve represents the 30 RFT training prompts; the other curves represent TD3, GPT-5, and pre- and post-RFT gpt-oss-20b during closed-loop evaluation. Observations from the four zones are pooled, and each curve is normalized by its own number of observations. All control times are included.}
\label{fig:training-deployment-distributions}
\end{figure}

Figure~\ref{fig:training-deployment-distributions} compares empirical cumulative distributions of zone temperature and CO$_2$ concentration in the 30 prompts used for RFT with those observed during closed-loop controller evaluation. All control times are included, and observations from the four zones are pooled for each curve.
Table~\ref{tab:observation-quantiles} reports the 5th, 50th, and 95th percentiles, defined as the smallest observation value at which the empirical cumulative fraction reaches the specified probability. These marginal summaries describe differences in the pooled observations; they neither characterize the full joint state distribution nor establish a causal effect on control performance.

\par\medskip\noindent\begin{minipage}{\textwidth}
\castblcaption
\centering
\captionof{table}{Quantiles of zone observations in the 30 RFT training prompts and during closed-loop evaluation. All control times and four zones are pooled, as in Figure~\ref{fig:training-deployment-distributions}.}
\label{tab:observation-quantiles}
\begin{tabular}{lrrrrrr}
\toprule
 & \multicolumn{3}{c}{Temperature ($^{\circ}$C)} & \multicolumn{3}{c}{CO$_2$ (ppm)} \\
\midrule
Condition & P5 & Median & P95 & P5 & Median & P95 \\
\midrule
RFT training prompts (30) & 25.8 & 26.0 & 27.2 & 599 & 918 & 971 \\
TD3 & 25.5 & 26.0 & 26.5 & 590 & 853 & 946 \\
GPT-5 & 25.7 & 26.1 & 26.7 & 583 & 941 & 996 \\
gpt-oss-20b Pre-RFT & 20.9 & 25.7 & 26.9 & 594 & 945 & 1050 \\
gpt-oss-20b Post-RFT & 21.0 & 26.0 & 27.0 & 591 & 947 & 1043 \\
\bottomrule
\end{tabular}
\end{minipage}\par\medskip

\clearpage
\section{Detailed control trajectories}\label{app:control-trajectories}
\begin{figure}[H]
\centering
\includegraphics[width=\textwidth,height=0.85\textheight,keepaspectratio]{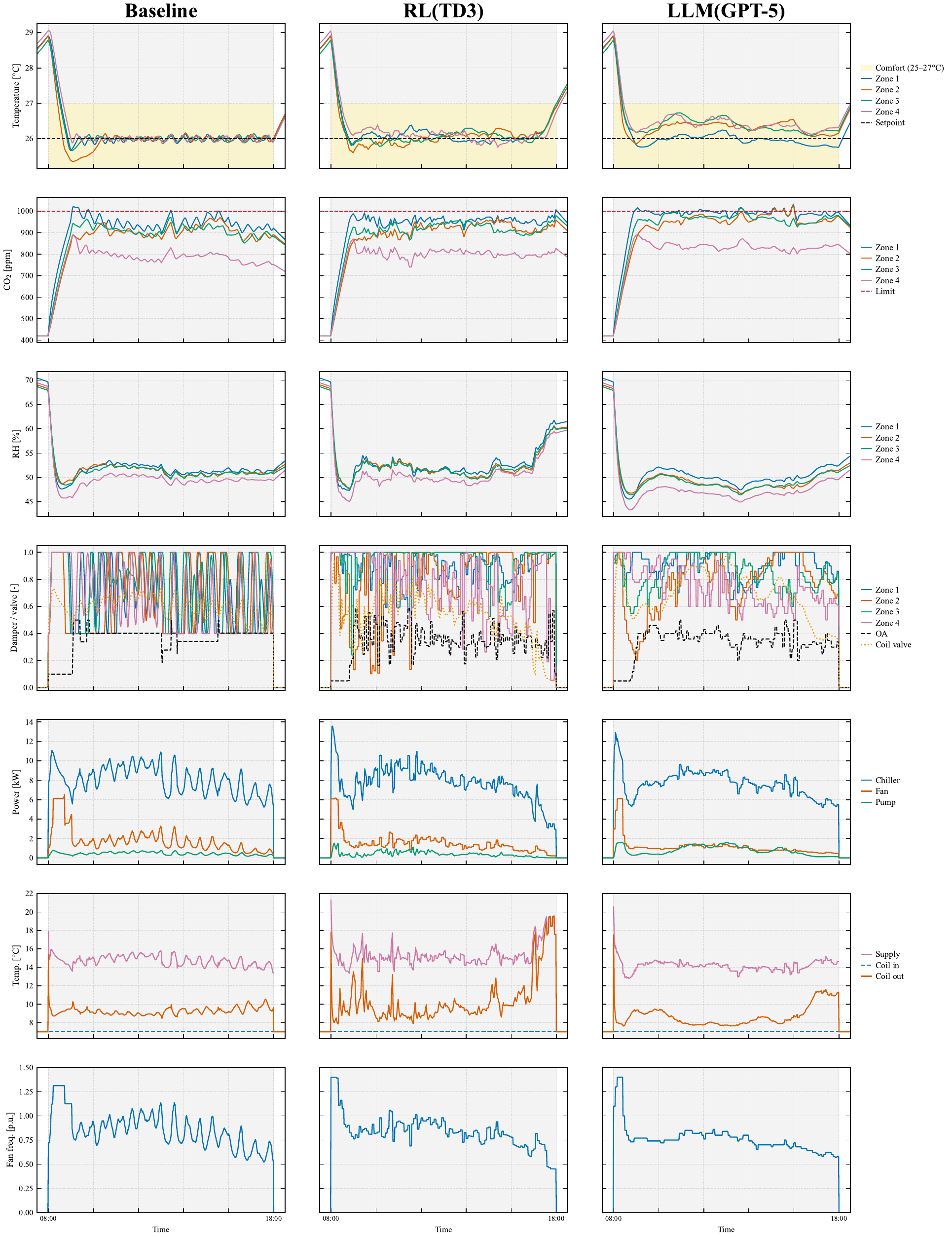}
\caption{Control trajectories on 28 July 2025 for the G36-based baseline, TD3, and GPT-5. From top to bottom, the rows show zone temperature, zone CO$_2$, zone relative humidity, damper and valve commands, equipment power, supply-air and coil-water temperatures, and fan speed.}
\label{fig:control-detail-reference}
\end{figure}

\begin{figure}[H]
\centering
\includegraphics[width=\textwidth,height=0.85\textheight,keepaspectratio]{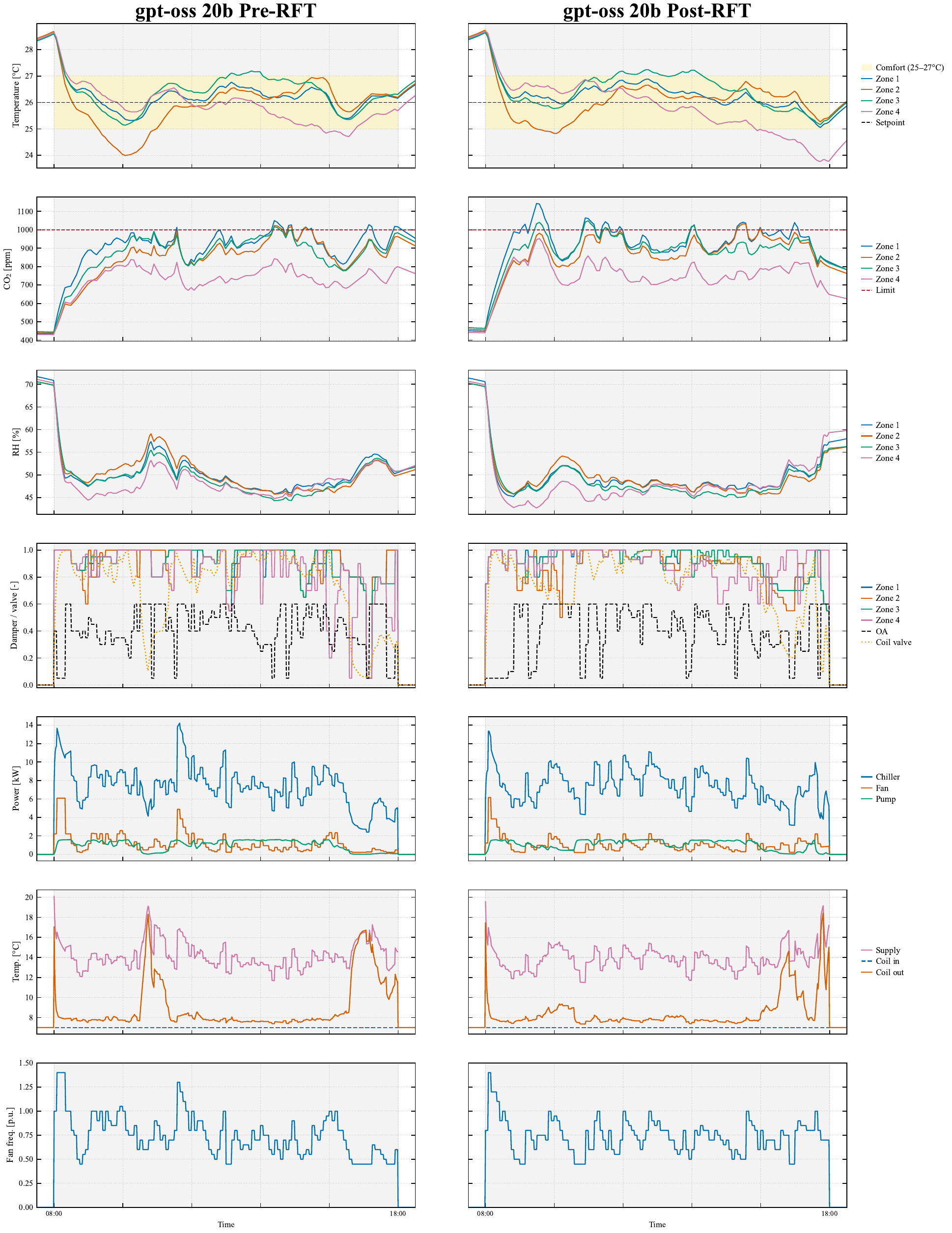}
\caption{Control trajectories on 28 July 2025 for \texttt{gpt-oss-20b} before and after RFT. The rows show the same signals as in Figure~\ref{fig:control-detail-reference}.}
\label{fig:control-detail-gptoss}
\end{figure}

\noindent\begin{minipage}{\textwidth}
We assess humidity using the fraction of occupied zone-minutes with relative humidity (RH) between 40\% and 70\%. This range is specified in Japan's building environmental hygiene standards \cite{mhlw_building_environment_criteria} and recommended in UK Health and Safety Executive guidance \cite{hse_sbs_hsg132}. This is an additional performance measure; humidity is not included in the training reward. Table~\ref{tab:humidity-performance} covers 08:00--18:00 over five days and four zones, giving 12{,}000 zone-minutes per controller. The GPT-5 results are from the first run. RH was within this range for 99.89--99.91\% of the evaluated zone-minutes. Figures~\ref{fig:control-detail-reference} and~\ref{fig:control-detail-gptoss} include zone RH trajectories for 28 July.
\end{minipage}

\par\medskip\noindent\begin{minipage}{\textwidth}
\centering
\captionof{table}{Fraction of occupied zone-minutes with $40\%\leq\mathrm{RH}\leq70\%$ over the five-day evaluation for each controller.}
\label{tab:humidity-performance}
\begin{tabular}{lr}
\toprule
Controller & Within range (\%) \\
\midrule
G36-based & 99.91 \\
TD3 & 99.91 \\
GPT-5 & 99.89 \\
gpt-oss-20b Pre-RFT & 99.90 \\
gpt-oss-20b Post-RFT & 99.91 \\
\bottomrule
\end{tabular}
\end{minipage}

\clearpage

\clearpage

\clearpage
\section{Representative GPT-5 short-horizon forecast}\label{app:gpt5-forecast-example}
The following excerpt is the GPT-5 control example at 09:20 on 29 July 2025 used in the qualitative analysis. The surrounding response text is omitted, as marked by the ellipses; the excerpt itself is reproduced unchanged.

\promptrole{GPT-5 API response at 09:20}
\begin{lstlisting}[style=promptblock]
......
5-min forecasts if unchanged: Z1 temp ~25.85C, CO2 ~996.5 ppm; Z2/Z3 remain ~26.1C; Z4 ~26.5C.
Control strategy for next 5 minutes:
- Prioritize cooling Zone 4 toward 26C by reallocating supply air rather than raising fan speed to save energy.
- Maintain or slightly improve CO2 safety margin, especially for Zone 1, with minimal OA increase to avoid chiller load jump.
- Avoid overcooling Zones 1-3; keep fan power steady; modest coil adjustment to offset a slight OA rise and assist Z4 cooling.
Chosen adjustments:
- Zone dampers: increase Z4 to 0.90 to deliver more cooling; slightly trim Z2 to 0.55 and Z3 to 0.62 to redistribute flow; keep Z1 at 0.80 to preserve ventilation for its near-limit CO2 while not increasing cooling too much.
- OA damper: small increase from 0.40 to 0.42 to add CO2 margin with limited energy impact given hot outdoor air.
- Coil valve: modest increase from 0.51 to 0.53 to maintain supply air temperature with the added OA and support Z4 cooling.
- Fan speed: hold at 0.78 to minimize fan energy; damper redistribution handles zone needs.
Expected outcomes over next 5 min:
- Zone 4 temperature should trend down modestly (~0.1-0.2C), moving toward 26C.
- Zones 2 and 3 remain near 26.1C; minor warming acceptable and within band.
- Zone 1 stays near 25.8-26.0C; CO2 remains below 1000 ppm with a slightly improved margin.
......
\end{lstlisting}

At 09:25, after the selected action, the emulator returned zone temperatures of 25.82, 26.15, 26.15, and 26.47$\,^{\circ}$C and CO$_2$ concentrations of 991, 907, 955, and 858~ppm. These values agree with the stated ranges and directions overall, although Zone~4 cooled less than the predicted 0.1--0.2$\,^{\circ}$C. This is one qualitative example; Section~\ref{sec:transition-evaluation} provides the controlled quantitative test.

\section{Representative \texttt{gpt-oss} reasoning before and after RFT}\label{app:gptoss-reasoning-example}
Table~\ref{tab:gptoss-reasoning-actions} and the excerpts below show one control step before RFT and one after RFT. They come from different trajectories and are not a matched-state test. They are included because both display the failure pattern discussed in Section~\ref{subsec:control-results}. The excerpts are verbatim except for the bracketed omissions.

\begin{center}
\begin{minipage}{\columnwidth}
\castblcaption
\sffamily\small
\captionof{table}{Representative \texttt{gpt-oss-20b} actions before and after RFT. Command rows show previous $\rightarrow$ selected values.}
\label{tab:gptoss-reasoning-actions}
\centering
\scriptsize
\setlength{\tabcolsep}{3pt}
\begin{tabular}{@{}>{\raggedright\arraybackslash}p{0.20\linewidth}>{\raggedright\arraybackslash}p{0.30\linewidth}>{\raggedright\arraybackslash}p{0.30\linewidth}@{}}
\toprule
Variable & Pre-RFT & Post-RFT \\
\midrule
Observed state & \makecell[l]{$T_{\max}=27.3\,^{\circ}$C\\$C_{\max}=1000$~ppm} & \makecell[l]{$T_{\max}=27.3\,^{\circ}$C\\$C_{\max}<1000$~ppm} \\
Zone dampers & $[1,1,1,1]$ unchanged & $[.95,.95,1,.95]$ unchanged \\
OA damper & $0.55\rightarrow0.05$ & $0.60\rightarrow0.05$ \\
Coil valve & $0.95\rightarrow1.00$ & $1.00\rightarrow1.00$ \\
Fan speed & $0.65\rightarrow0.75$ & $0.70\rightarrow0.60$ \\
\bottomrule
\end{tabular}
\end{minipage}
\end{center}

\promptrole{Pre-RFT excerpt}
\begin{lstlisting}[style=promptblock]
Current CO2: Zone1=1000.0, zone2=952.0, zone3=974.0, zone4=773.0. They are all below 1000 ppm except zone1 exactly 1000. So we can reduce OA to minimal. Let's set oa_damper near lower bound 0.05. But is that too low? It may cause CO2 to rise. Let's estimate CO2: CO2 change rates are negative, meaning decreasing. So CO2 is trending down. So safe to reduce OA. So set oa_damper to 0.05 (lower bound).
[... omitted ...]
Perhaps fan speed 0.75 gives 0.422. We'll go 0.75.
[... omitted ...]
Check CO2: zone1 1000 ppm, CO2 trend -0.4 ppm/min for 5 min = -2 ppm, so stays 999.6. So fine. So safe.
\end{lstlisting}

\promptrole{Post-RFT excerpt}
\begin{lstlisting}[style=promptblock]
Better approach: reduce fan speed from 0.7 to 0.6. Keep zone dampers at 0.95 for 1,2,4. Zone 3 at 1.0.

But we also have OA damper; we can reduce OA to 0.05 to reduce heating, but that reduces heat removal. But coil valve 1.0 will provide cooling. So we set oa_damper 0.05.
[... omitted ...]
But is 0.05 too low causing high CO2? Current CO2 all below 1000 and decreasing; so fine. Also not exceed 1050. So it's okay.

Check if any constraints violation: Temperature zone 3? With OA at 0.05, maybe zone 3 will not get enough ventilation to cool. But coil valve at max maybe enough. But we don't have dynamic model. We'll assume it's okay.
\end{lstlisting}

Both examples extrapolate CO$_2$ trends measured under the previous outdoor-air command across a large change to the lower bound, without predicting CO$_2$ under the new action. In the post-RFT example, Zone~3 is already above the comfort band, its damper and the coil valve are at their upper bounds, and the model still lowers fan speed without estimating the resulting temperature. The statement ``we don't have dynamic model. We'll assume it's okay'' is explicit about what is missing: the constraint check is attempted and then abandoned because the model cannot produce the required prediction. Because the states differ, these examples cannot isolate the effect of RFT. They show that the same pattern---the check is attempted, the prediction is not available---remains after training, consistent with the control scores in Table~\ref{tab:control-performance} and the transition test in Section~\ref{sec:transition-evaluation}.

\section{Prompt for VAV control}\label{app:control-prompt}
Both LLM controllers used the same three-part prompt. At each decision, the placeholders were filled with the zone names, number of zones, and current observation.

\promptrole{System message}
\begin{lstlisting}[style=promptblockcompact]
You are a supervisory decision-making agent for a multi-zone VAV HVAC system.
The controlled zones are: {zone_list}.
Select actuator setpoints every 5 minutes.
The goal is to keep all zones comfortable while minimising electrical power.

Follow these physics-based objectives and constraints:
- Comfort: target 26C with an acceptable band of 25-27C for every zone. Prefer values close to 26C.
- CO2/ventilation: minimise outdoor-air intake while keeping each zone's CO2 <= 1000 ppm; it must never exceed 1050 ppm.
- Electrical power to minimise (use these relations for trade-offs):
    Fan power: P_fan ~ P_fan_ref * (f/1.0)^3, where 'fan_speed' f is relative speed (1.0 = nominal).
    Pump power (water-side via coil): P_pump ~ P_pump_ref * (w/1.0)^3, where w is proportional to water flow (use 'coil_valve' as a proxy).
    Increasing water flow is usually cheaper than increasing fan speed for the same load.
    Total power to minimise: P_total = P_fan + P_pump (+ chiller/compressor if applicable). Reducing outdoor air also reduces coil/chiller power.
- Static-pressure reset logic: keep at least one zone damper at 100% open; minimising throttling loss lowers the required static pressure and reduces fan power.
- Avoid abrupt actuator changes unless a constraint would be violated.
\end{lstlisting}

\promptrole{Developer message}
\begin{lstlisting}[style=promptblockcompact]
Developer instructions:
- Formatting rules: respond with a single JSON object.
- Use keys 'thought_process' (string) and 'action' (object).
- The 'action' object must include:
  - 'zone_dampers' (list of length {n})
  - 'oa_damper'
  - 'coil_valve'
  - 'fan_speed'
- Provide the thought_process as explicit step-by-step reasoning.
- Return numeric values as floats; clip them to the provided bounds before finalising.
\end{lstlisting}

\promptrole{User message}
\begin{lstlisting}[style=promptblockcompact]
Observation for the next 5-minute HVAC control interval:
{observation_json}
Choose actuator commands that:
- keep all zones within 25-27C (target 26C),
- keep CO2 <= 1000 ppm (never exceed 1050 ppm),
- minimise outdoor-air intake, and
- minimise total electrical power P_total
  using the cubic scaling of fan/pump power described above.
\end{lstlisting}

\section{Prompt and output format for predicting VAV system responses}\label{app:evaluation-prompts}
\subsection{Direct-answer setting}
Each request included the current observation, previous action, actuator bounds, varied actuator, two candidate actions, and five-minute horizon:

\begin{lstlisting}[style=promptblockcompact]
{"observation": {...}, "varied_actuator": "<family>",
 "action_low": {...}, "action_high": {...}, "horizon_minutes": 5}
\end{lstlisting}

The same instructions were used in all three direct-answer conditions:

\begin{lstlisting}[style=promptblockcompact,breakindent=0pt,breakautoindent=false,breakatwhitespace=true]
You are evaluating an action-conditioned HVAC transition model. The input contains one current observation and two actuator actions that differ in one named actuator. Predict the observable state exactly five minutes after each action begins. Predict both cases separately. Return only the requested structured values. Do not provide analysis, rationale, control advice, or an alternative action. Make best numerical predictions even when uncertain and preserve zone order (Zones 1, 2, 3, 4).

For both the low and high cases, make use of the observation fields temp_delta_c_per_min and co2_delta_ppm_per_min, which represent the measured one-minute changes immediately preceding the start of either new action.
\end{lstlisting}

The output schema required four temperature and four CO$_2$ predictions for each action, an \texttt{abstain} flag, and a confidence score from 0 to 1. GPT-5 used minimal reasoning effort. For \texttt{gpt-oss-20b}, thinking was disabled, with temperature 0 and seed 0. All three conditions---GPT-5, \texttt{gpt-oss-20b} before RFT, and \texttt{gpt-oss-20b} after RFT---produced zero reasoning tokens.

\subsection{Error analysis of direct-answer predictions}
We re-scored the saved pre-RFT outputs for the four actuator--variable pairs in Figure~\ref{fig:transition-evaluation}(c). All 36 responses contained valid numerical predictions, so the low accuracy did not result from parsing failures. All numerical predictions were scored, including the 29 responses with \texttt{abstain=true}.

\par\medskip\noindent\begin{minipage}{\columnwidth}
\castblcaption
\centering
\captionof{table}{Pre-RFT direct-answer error counts. Small differences include predicted ties and differences whose magnitude does not exceed 0.02~$^{\circ}$C or 2~ppm. Only comparisons with emulator differences above these thresholds are included.}
\label{tab:transition-error-counts}
\small
\setlength{\tabcolsep}{4pt}
\begin{tabular}{@{}lrrrr@{}}
\toprule
Effect & Eligible & Correct & Small & Opposite \\
\midrule
Coil $\rightarrow T$ & 44 & 0 & 40 & 4 \\
Fan $\rightarrow T$ & 48 & 1 & 39 & 8 \\
OA $\rightarrow$ CO$_2$ & 48 & 0 & 48 & 0 \\
Fan $\rightarrow$ CO$_2$ & 43 & 0 & 42 & 1 \\
\midrule
Total & 183 & 1 & 169 & 13 \\
\bottomrule
\end{tabular}
\end{minipage}\par\medskip

Of the 182 errors, 169 (92.9\%) had predicted differences at or below the threshold, including 128 numerical ties (absolute difference below $10^{-9}$). The remaining 13 predicted the opposite direction with a difference above the threshold. A tie between the two action predictions does not necessarily mean persistence: it means that the model predicted the same future value for both actions. Because ties and small predicted differences count as errors, this score is not a binary direction-choice task with a 50\% chance baseline.

For example, increasing the outdoor-air damper command from 0.147 to 0.422 at one state lowered the simulated five-minute zone~1 CO$_2$ concentration by 63.00~ppm, but the model predicted 970.15~ppm for both actions. At another state, increasing fan speed from 0.660 to 1.135 lowered zone~1 temperature by 0.206~$^{\circ}$C, but both predictions were 26.20~$^{\circ}$C. In a further case, increasing fan speed lowered zone~1 CO$_2$ by 53.65~ppm, but both predictions were 963.70~ppm. An opposite-direction error also occurred: increasing fan speed from 0.631 to 1.106 lowered zone~3 temperature by 0.172~$^{\circ}$C, whereas the predicted values rose from 26.20 to 26.40~$^{\circ}$C. These diagnostic examples were selected for the largest absolute emulator effect within the corresponding effect and error category, rather than sampled at random.

As a scoring check, counting only the sign of nonzero predicted differences, without requiring their magnitudes to exceed the practical thresholds, gave 0/44, 3/48, 8/48, and 2/43 correct comparisons for coil--temperature, fan--temperature, outdoor-air--CO$_2$, and fan--CO$_2$, respectively. Thus, the poor performance was not explained by the magnitude threshold alone. These observations describe the outputs under the direct-answer setting; they do not identify the internal cause of the errors.

\subsection{Transition prediction with thinking enabled}
We also evaluated the pre- and post-RFT models with thinking enabled, using temperature 0 and medium reasoning effort. The test used the same 12 states, 36 action pairs, and scoring thresholds as the direct-answer test. It examines prediction with the thinking-enabled generation mode used during RFT. Table~\ref{tab:transition-thinking} reports one evaluation set per model. CO$_2$ MAE was lower than the persistence baseline (20.00~ppm), while temperature MAE was higher. Direction accuracy increased for some effects and decreased for others after RFT; these single evaluations do not estimate run-to-run variability.

\par\medskip\noindent\begin{minipage}{\columnwidth}
\castblcaption
\centering
\captionof{table}{Five-minute transition prediction with thinking enabled for \texttt{gpt-oss-20b}, before and after RFT. Results use 12 states and 36 action pairs, with one evaluation set per model, temperature 0, and medium reasoning effort.}
\label{tab:transition-thinking}
\small
\begin{tabular}{@{}lrr@{}}
\toprule
Metric & Pre-RFT & Post-RFT \\
\midrule
Temperature MAE ($^{\circ}$C) & 0.231 & 0.241 \\
CO$_2$ MAE (ppm) & 17.42 & 18.21 \\
Coil $\rightarrow T$ accuracy (\%) & 42.1 & 48.4 \\
Fan $\rightarrow T$ accuracy (\%) & 46.1 & 42.9 \\
OA $\rightarrow$ CO$_2$ accuracy (\%) & 43.5 & 50.2 \\
Fan $\rightarrow$ CO$_2$ accuracy (\%) & 60.3 & 58.6 \\
OA $\rightarrow T$ accuracy (\%) & 33.5 & 32.6 \\
\bottomrule
\end{tabular}
\end{minipage}\par\medskip

\clearpage
\section{Final-day control and period-specific comparisons}\label{app:g36-final-day}
\begin{figure}[H]
\centering
\includegraphics[width=0.9\textwidth]{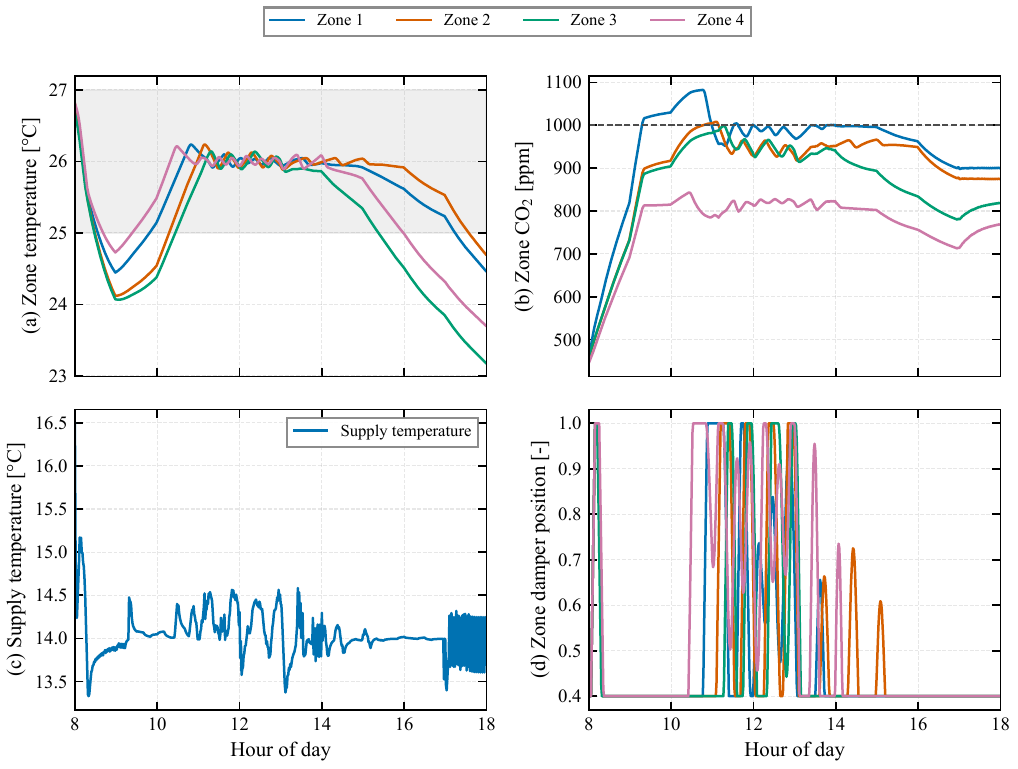}
\caption{G36-based control during occupied hours on 1 August 2025. (a) Zone temperatures, with the 25--27~$^{\circ}$C band shaded. (b) Zone CO$_2$ concentrations, with the 1000~ppm threshold dashed. (c) Logged supply-air temperature. (d) Zone damper positions. Colors identify the same zones in panels (a), (b), and (d).}
\label{fig:g36-final-day}
\end{figure}

Figure~\ref{fig:g36-final-day} shows the G36-based controller on 1 August. Temperature compliance was 71.25\%, the minimum zone temperature was 23.18~$^{\circ}$C, and integrated temperature deviation was 6.95~$^{\circ}$C$\cdot$h. CO$_2$ compliance was 94.00\%, with cumulative exceedance of 79.22~ppm$\cdot$h and a maximum of 1083~ppm. This day contributed approximately 63.5\% of the G36-based controller's five-day cumulative CO$_2$ exceedance; zone~1 accounted for 77.57~ppm$\cdot$h of the daily total. Thus, the five-day comparison with GPT-5 reflects a concentrated final-day exceedance as well as the frequency of violations across the evaluation.

The final day had a lower outdoor temperature and higher humidity than on 28--30 July, together with a different occupancy schedule (Figure~\ref{fig:scenario}). Under these conditions, the G36-based controller exhibited both overcooling and CO$_2$ exceedance concentrated in zone~1. These observations indicate that the performance of this implementation varied with operating conditions and suggest that its control settings may require reassessment under different loads. The present comparison does not isolate the effects of weather, occupancy, or individual control parameters, and does not establish a general limitation of Guideline~36.

\clearpage
\subsection{Energy and CO$_2$ comparison by evaluation period}
Table~\ref{tab:period-energy} reports electricity use and CO$_2$ performance from the one-minute trajectories. Electricity covers all hours; CO$_2$ metrics cover 08:00--18:00 across four zones. Excluding 1 August, GPT-5 used 6.27\% less electricity than G36, but had lower CO$_2$ compliance (93.85\% versus 98.11\%) and greater cumulative exceedance (52.31 versus 45.61~ppm$\cdot$h). Thus, the energy--ventilation trade-off remained when the final day was excluded. The five-day results remain the primary comparison. This supplementary breakdown compares G36, TD3, and GPT-5.

\par\medskip\noindent\begin{minipage}{\textwidth}
\castblcaption
\captionof{table}{Electricity use and CO$_2$ performance by evaluation period. $\Delta E$ is relative to G36 over the same period; CO$_2$ compliance and exceedance use a 1000~ppm threshold.}
\label{tab:period-energy}
\centering
\setlength{\tabcolsep}{5pt}
\renewcommand{\arraystretch}{1.08}
\begin{tabular*}{\textwidth}{@{\extracolsep{\fill}}llrrrr@{}}
\toprule
Period & Controller & \makecell{Energy\\(kWh)} & $\Delta E$ (\%) & \makecell{CO$_2$ compliance\\(\%)} & \makecell{CO$_2$ exceedance\\(ppm$\cdot$h)} \\
\midrule
28--31 July & G36-based & 359.44 & --- & 98.11 & 45.61 \\
 & TD3 & 345.40 & -3.91 & 99.98 & 0.17 \\
 & GPT-5 & 336.91 & -6.27 & 93.85 & 52.31 \\
\midrule
1 August & G36-based & 57.75 & --- & 94.00 & 79.22 \\
 & TD3 & 53.23 & -7.83 & 100.00 & 0.00 \\
 & GPT-5 & 51.32 & -11.14 & 99.33 & 0.51 \\
\midrule
All five days & G36-based & 417.18 & --- & 97.29 & 124.83 \\
 & TD3 & 398.63 & -4.45 & 99.98 & 0.17 \\
 & GPT-5 & 388.23 & -6.94 & 94.95 & 52.82 \\
\bottomrule
\end{tabular*}
\end{minipage}\par\medskip

\subsection{GPT-5 control on the final day}
On 1 August, GPT-5 used 51.32~kWh, maintained all zone temperatures within 25--27~$^{\circ}$C during occupied hours, and achieved CO$_2$ compliance of 99.33\% (Table~\ref{tab:final-day-comfort}). Figure~\ref{fig:gpt5-final-day} shows the first-run trajectory under the final-day conditions.

\par\medskip\noindent\begin{minipage}{\textwidth}
\castblcaption
\captionof{table}{Temperature and CO$_2$ performance on 1 August, evaluated over 08:00--18:00 across four zones.}
\label{tab:final-day-comfort}
\centering
\begin{tabular}{lrrrrr}
\toprule
Controller & \makecell{Temperature\\compliance (\%)} & \makecell{Temperature deviation\\($^{\circ}$C$\cdot$h)} & \makecell{CO$_2$\\compliance (\%)} & \makecell{CO$_2$ exceedance\\(ppm$\cdot$h)} & \makecell{Peak CO$_2$\\(ppm)} \\
\midrule
G36-based & 71.25 & 6.95 & 94.00 & 79.22 & 1083 \\
TD3 & 100.00 & 0.00 & 100.00 & 0.00 & 980 \\
GPT-5 & 100.00 & 0.00 & 99.33 & 0.51 & 1004 \\
\bottomrule
\end{tabular}
\end{minipage}\par\medskip

\par\smallskip\noindent\begin{minipage}{\textwidth}
\centering
\includegraphics[width=0.50\textwidth]{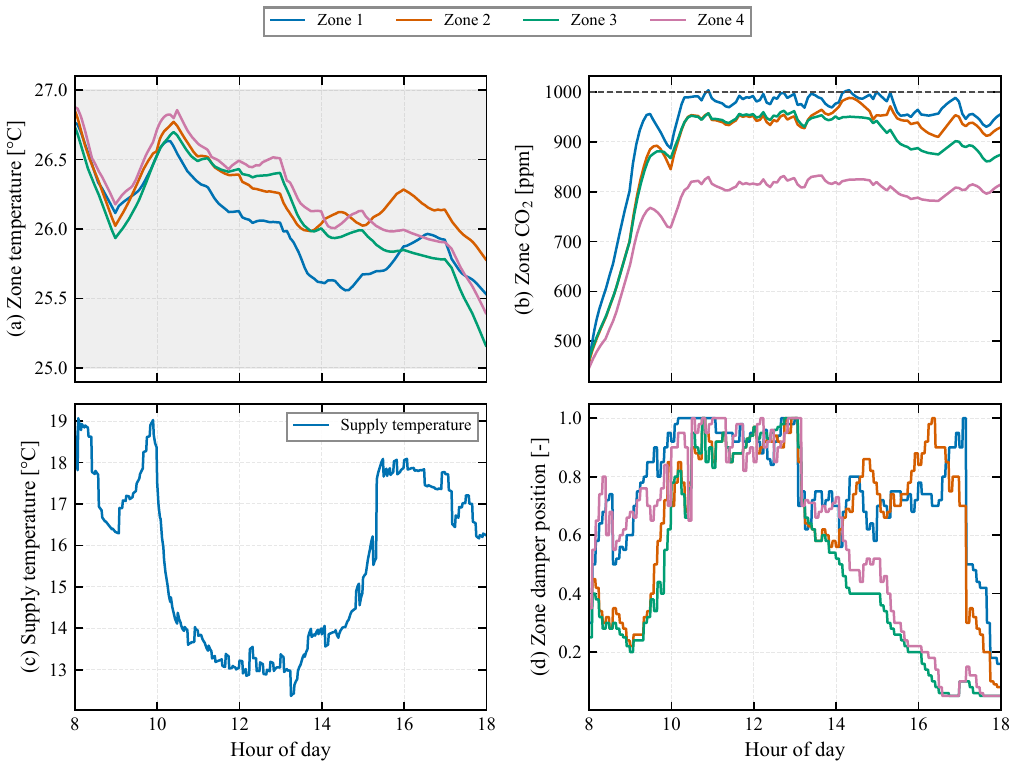}
\captionof{figure}{GPT-5 control during occupied hours on 1 August 2025, using the same panel definitions and zone colors as Figure~\ref{fig:g36-final-day}.}
\label{fig:gpt5-final-day}
\end{minipage}

\clearpage
\section{Candidate-selection loss in the critic audit}\label{app:critic-selection-loss}
\begin{figure}[H]
\centering
\includegraphics[width=\textwidth]{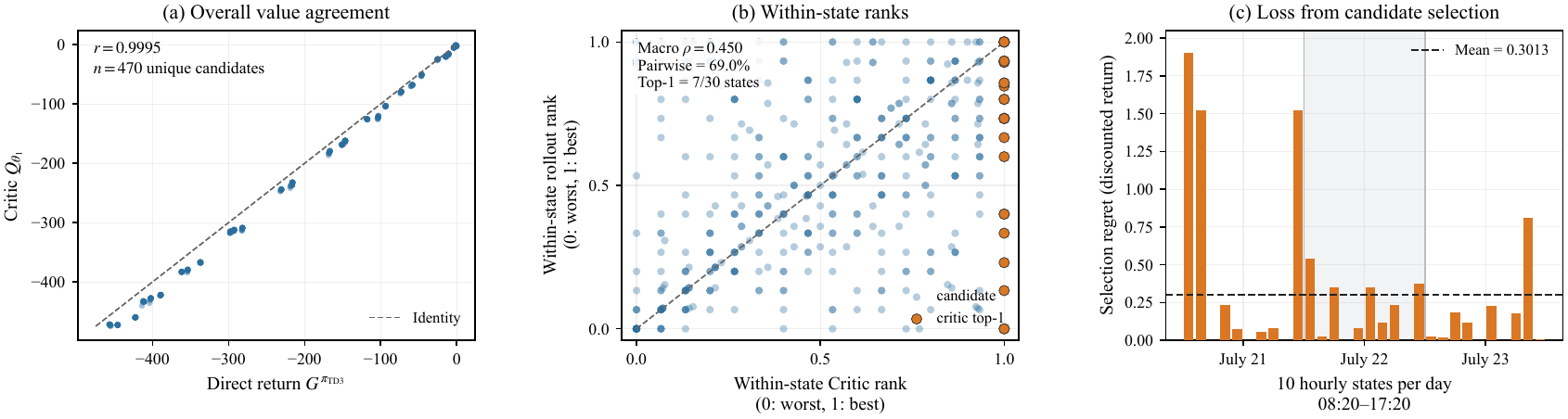}
\caption{Supplementary critic diagnostics for the same 470 unique candidates. (a) Pooled critic--return correlation. (b) Within-state ranks, with critic-selected candidates in orange. (c) Selection regret for each of the 30 states, grouped by day; the dashed line marks the mean. States within each day share a trajectory.}
\label{fig:critic-selection-loss}
\end{figure}

The 30-state audit evaluated 470 unique candidates from three daily trajectories. For each state, selection regret is the largest direct return among its candidates minus the direct return of the candidate selected by the critic; zero regret means that the critic selected a rollout-best candidate. Mean regret was $0.3013$ in discounted-return units. Figure~\ref{fig:critic-selection-loss} shows the pooled agreement, the within-state ranks, and the regret for each state. The pooled correlation of $r=0.9995$ in panel (a) uses the same 470 candidates as the within-state metrics.

\begin{figure}[H]
\centering
\includegraphics[width=0.45\textwidth]{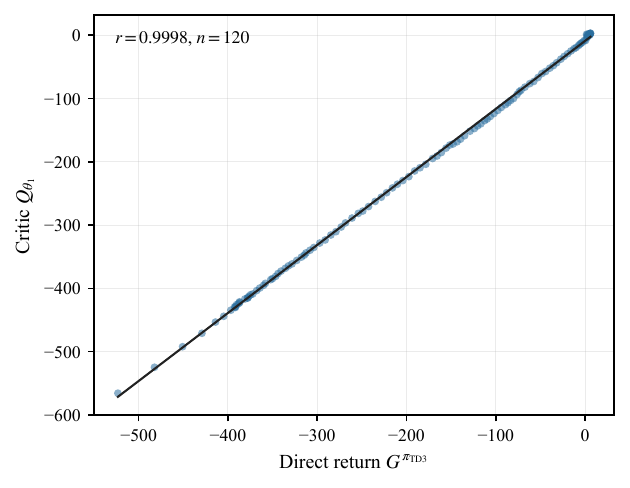}
\caption{Critic estimates and direct returns along a 120-decision TD3 evaluation trajectory. The main audit compares pooled and within-state agreement on the same 470 candidates.}
\label{fig:critic-td3-trajectory}
\end{figure}

Along a 120-decision TD3 evaluation trajectory, critic estimates and direct returns had a Pearson correlation of $r=0.9998$ (Figure~\ref{fig:critic-td3-trajectory}). This trajectory is separate from the 470-candidate audit used in the main figure.

\clearpage
\section{Repeated GPT-5 closed-loop evaluations}\label{app:gpt5-repeats}
GPT-5 was evaluated in three independent five-day runs with the same prompt, model settings, weather, occupancy, and initial state. The main control comparisons use the first run. Tables~\ref{tab:gpt5-repeats} and~\ref{tab:gpt5-repeat-energy} report all three runs. All used less electricity than G36 and TD3, but had lower CO$_2$ compliance. These repeats show how closed-loop results vary across runs. They do not test repeated API outputs at the same state. The evidence is limited to three runs under the evaluated conditions.

\par\medskip\noindent\begin{minipage}{\textwidth}
\centering
\captionof{table}{Five-day GPT-5 results across three independent runs. $\Delta E$ is relative to G36. Compliance and deviation metrics use the same occupied-hour definitions as Table~\ref{tab:control-performance}.}
\label{tab:gpt5-repeats}
\begin{tabular}{lrrr}
\toprule
Metric & Baseline run & Repeat 1 & Repeat 2 \\
\midrule
Electricity (kWh) & 388.23 & 386.53 & 388.84 \\
$\Delta E$ (\%) & $-6.94$ & $-7.35$ & $-6.79$ \\
Outdoor-air intake (kg) & 63{,}963 & 63{,}572 & 64{,}003 \\
Temperature compliance (\%) & 97.38 & 97.51 & 97.21 \\
Temperature deviation ($^{\circ}$C$\cdot$h) & 4.64 & 4.39 & 4.91 \\
CO$_2$ compliance (\%) & 94.95 & 93.62 & 92.71 \\
CO$_2$ exceedance (ppm$\cdot$h) & 52.82 & 59.99 & 57.00 \\
Maximum CO$_2$ (ppm) & 1033 & 1026 & 1022 \\
\bottomrule
\end{tabular}
\end{minipage}

\par\medskip\noindent\begin{minipage}{\textwidth}
\centering
\captionof{table}{Equipment electricity use across the three GPT-5 runs. Rounded component values may not sum to the totals in Table~\ref{tab:gpt5-repeats}.}
\label{tab:gpt5-repeat-energy}
\begin{tabular}{lrrr}
\toprule
Electricity (kWh) & Baseline run & Repeat 1 & Repeat 2 \\
\midrule
Pump & 28.5 & 23.3 & 36.4 \\
Fan & 42.7 & 47.1 & 35.1 \\
Chiller & 317.1 & 316.1 & 317.3 \\
\bottomrule
\end{tabular}
\end{minipage}

\clearpage
\section{Computational requirements}\label{app:computational-cost}
Table~\ref{tab:compute-models} summarizes training and inference costs. GPT-5 response times were measured during the five-day control evaluation. Latency is the elapsed API-call time, including network overhead. All recorded response times were below the 300~s control interval; the maximum was 94.46~s. The estimated API cost was US\$22.45 per five-day run, based on prices of US\$1.25 per million input tokens and US\$10 per million output tokens. This estimate excludes the two repeat runs.

\par\medskip\noindent\begin{minipage}{\textwidth}
\centering
\captionof{table}{Training and inference requirements. The gpt-oss value is generation throughput, not the elapsed time per control decision.}
\label{tab:compute-models}
\begin{tabular}{p{0.22\textwidth}p{0.31\textwidth}p{0.37\textwidth}}
\toprule
Task & Hardware and setting & Time or usage \\
\midrule
RFT training & NVIDIA H100 SXM, 80 GB & 52.42 h \\
RFT responses & Generated / scored by direct rollout & 12{,}800 / 12{,}186 \\
gpt-oss-20b inference & NVIDIA H100, MXFP4 (4-bit) & Approximately 45 tokens/s \\
GPT-5 inference & Hosted API & Mean 45.05 s/response; median 43.32 s; 95th percentile 66.16 s \\
GPT-5 API cost & Five-day estimate per run & US\$22.45 \\
\bottomrule
\end{tabular}
\end{minipage}\par\medskip

The 12{,}186 scored responses include cache reuse and are not a count of newly executed simulator rollouts.

Table~\ref{tab:compute-verifier} compares scoring costs on an Apple M4 CPU with 32 GiB memory, PyTorch 2.9.0, and one PyTorch thread. Both methods used the same 30 states and one candidate per state. Critic scoring was repeated 100 times per state and direct rollout scoring three times. Model loading was excluded. Critic timing includes input preparation, both critic networks, and output retrieval. Rollout timing covers candidate application, TD3 continuation to 18:00, and return calculation and checks; it excludes preparation of the branch state. These timings are separate from the H100 training run and do not measure its verifier overhead. Critic scoring was about 9{,}500 times faster, but the audit showed that it did not reliably rank candidates within a state.

\par\medskip\noindent\begin{minipage}{\textwidth}
\centering
\captionof{table}{Mean critic and direct-rollout scoring time per candidate on an Apple M4 CPU with one PyTorch thread. Both methods use the same 30 states and one candidate per state, with 100 repetitions for the critic and three for direct rollouts.}
\label{tab:compute-verifier}
\begin{tabular}{lr}
\toprule
Scoring method & Mean time per candidate \\
\midrule
Critic & 0.0923 ms \\
Direct rollout & 0.8811 s \\
\bottomrule
\end{tabular}
\end{minipage}\par\medskip

Direct rollout cost increases with the number of candidates and the remaining simulation steps. Larger buildings may also require more computation per step, more action variables, and longer prompts. Caching repeated state--action pairs can reduce repeated work, and evaluating independent candidates in parallel can reduce elapsed time, but neither removes these scaling costs. We did not benchmark larger buildings or longer horizons. The H100 inference result also does not establish the response time on a typical local PC.

\clearpage
\section{RFT generation quality}\label{app:generation-quality}
Of 12{,}800 generated responses, 614 (4.80\%) were invalid or truncated, leaving 12{,}186 valid responses for scoring. Table~\ref{tab:generation-failures} reports the failure breakdown. Its percentages use the 614 failed responses as the denominator. Invalid and truncated responses were excluded from rollout scoring and the training loss.

No actions required clipping. Across the 800 groups of 16 generated responses, the mean number of valid responses was 15.2 (12{,}186/800). Duplicates are counted as copies beyond the first occurrence of an action within a group. There were 2.6 such copies per group on average, leaving approximately 12.6 distinct valid actions. Valid duplicates remained in the loss because repeated actions reflect the probability assigned to them by the policy. They shared the same rollout score, and repeated state--action pairs reused cached scores. The 12{,}186 scored responses therefore do not represent 12{,}186 newly executed rollouts. Using the rounded mean duplicate count gives an approximate upper estimate of 10{,}100 new candidate rollouts ($12{,}186-800\times2.6$); cache reuse across groups would reduce this further. This estimate is not a measured execution count.

\par\medskip\noindent\begin{minipage}{\textwidth}
\centering
\captionof{table}{Failure types among the 614 invalid or truncated responses out of 12{,}800 RFT generations. Percentages are relative to the 614 failed responses.}
\label{tab:generation-failures}
\begin{tabular}{lr}
\toprule
Failure type & Share of failed responses (\%) \\
\midrule
Truncated at the 4{,}000-token limit & 55.2 \\
Invalid JSON escape & 13.8 \\
Extra characters after JSON & 13.4 \\
JSON syntax error & 12.7 \\
Control characters in JSON & 2.4 \\
Missing action object & 2.0 \\
Unterminated JSON string & 0.5 \\
\bottomrule
\end{tabular}
\end{minipage}


\begin{thebibliography}{52}
\footnotesize

\bibitem{iea2026_buildings}
International Energy Agency,
\newblock Energy Efficiency Policy Toolkit: Buildings,
\newblock IEA, Paris, 2026.
\newblock \url{https://www.iea.org/reports/energy-efficiency-policy-toolkit/buildings} (accessed 24 September 2026).

\bibitem{jung2019_human_hvac_review}
W.~Jung and F.~Jazizadeh,
\newblock Human-in-the-loop HVAC operations: A quantitative review on occupancy, comfort, and energy-efficiency dimensions,
\newblock Appl. Energy 239 (2019) 1471--1508.
\newblock \url{https://doi.org/10.1016/j.apenergy.2019.01.070}.

\bibitem{shim2014_fan_control_vav}
G.~Shim, L.~Song, and G.~Wang,
\newblock Comparison of different fan control strategies on a variable air volume systems through simulations and experiments,
\newblock Build. Environ. 72 (2014) 212--22.
\newblock \url{https://doi.org/10.1016/j.buildenv.2013.11.003}.

\bibitem{lu2022_co2_dcv_critical_review}
X.~Lu, Z.~Pang, Y.~Fu, and Z.~O'Neill,
\newblock The nexus of the indoor {CO$_2$} concentration and ventilation demands underlying {CO$_2$}-based demand-controlled ventilation in commercial buildings: A critical review,
\newblock Build. Environ. 218 (2022) 109116.
\newblock \url{https://doi.org/10.1016/j.buildenv.2022.109116}.

\bibitem{xu2007_adaptive_dcv}
X.~Xu and S.~Wang,
\newblock An adaptive demand-controlled ventilation strategy with zone temperature reset for multi-zone air-conditioning systems,
\newblock Indoor Built Environ. 16(5) (2007) 426--37.
\newblock \url{https://doi.org/10.1177/1420326X07082744}.

\bibitem{ref:shi2025_hybrid_mas}
S.~Shi, S.~Miyata, and Y.~Akashi,
\newblock A hybrid multi-agent distributed optimal control strategy of multizone {VAV} systems for edge computing in smart buildings,
\newblock Energy Build. 345 (2025) 116089.
\newblock \url{https://doi.org/10.1016/j.enbuild.2025.116089}.

\bibitem{ahn2001_chw_quadratic_control}
B.~C. Ahn and J.~W. Mitchell,
\newblock Optimal control development for chilled water plants using a quadratic representation,
\newblock Energy Build. 33 (2001) 371--8.
\newblock \url{https://doi.org/10.1016/S0378-7788(00)00119-5}.

\bibitem{jette1998_pi_dual_duct}
I.~Jett{\'e}, M.~Zaheer-Uddin, and P.~Fazio,
\newblock {PI}-control of dual duct systems: Manual tuning and control loop interaction,
\newblock Energy Convers. Manag. 39(14) (1998) 1471--82.
\newblock \url{https://doi.org/10.1016/S0196-8904(98)00020-X}.

\bibitem{fiducioso2019_safe_pid_tuning}
M.~Fiducioso, S.~Curi, B.~Schumacher, M.~Gwerder, and A.~Krause,
\newblock Safe contextual Bayesian optimization for sustainable room temperature {PID} control tuning,
\newblock in: Proceedings of the Twenty-Eighth International Joint Conference on Artificial Intelligence ({IJCAI} 2019), 2019, pp. 5850--6.
\newblock \url{https://doi.org/10.24963/ijcai.2019/811}.

\bibitem{behrooz2018_fcm_hvac_review}
F.~Behrooz, N.~Mariun, M.~H. Marhaban, M.~A. Mohd~Radzi, and A.~R. Ramli,
\newblock Review of control techniques for {HVAC} systems---nonlinearity approaches based on fuzzy cognitive maps,
\newblock Energies 11 (2018) 495.
\newblock \url{https://doi.org/10.3390/en11030495}.

\bibitem{yamamoto2021_shase_vav_vwv_co2_part1}
S.~Yamamoto, S.~Miyata, Y.~Akashi, T.~Sawachi, and M.~Momota,
\newblock Energy-saving effect of {VAV}, {VWV} and {CO$_2$} concentration control of air conditioning system considering automatic control logic and parameters, Part 1---Simulation construction and energy saving effect of {VAV}/{VWV} control,
\newblock Trans. Soc. Heat. Air Cond. Sanit. Eng. Jpn. 46(293) (2021) 23--32 (in Japanese).
\newblock \url{https://doi.org/10.18948/shase.46.293_23}.

\bibitem{ref:ashrae_g36_2024}
ASHRAE,
\newblock ASHRAE Guideline 36-2024: High-Performance Sequences of Operation for {HVAC} Systems,
\newblock ASHRAE, Peachtree Corners, GA, 2024.

\bibitem{ref:drgona2020_mpc_buildings}
J.~Drgo{\v{n}}a, J.~Arroyo, I.~C. Figueroa, D.~Blum, K.~Arendt, D.~Kim, et al.,
\newblock All you need to know about model predictive control for buildings,
\newblock Annu. Rev. Control 50 (2020) 190--232.
\newblock \url{https://doi.org/10.1016/j.arcontrol.2020.09.001}.

\bibitem{ref:saloux2025_mpc_field_implementations}
E.~Saloux, J.~A. Candanedo, C.~Vallianos, N.~Morovat, and K.~Zhang,
\newblock From theory to practice: A critical review of model predictive control field implementations in the built environment,
\newblock Appl. Energy 393 (2025) 126091.
\newblock \url{https://doi.org/10.1016/j.apenergy.2025.126091}.

\bibitem{ref:wang2020_rl_building_controls}
Z.~Wang and T.~Hong,
\newblock Reinforcement learning for building controls: The opportunities and challenges,
\newblock Appl. Energy 269 (2020) 115036.
\newblock \url{https://doi.org/10.1016/j.apenergy.2020.115036}.

\bibitem{ref:savino2025_drl_g36}
S.~Savino, G.~Razzano, M.~Pagone, C.~Novara, and A.~Capozzoli,
\newblock Deploying deep reinforcement learning for low-level {HVAC} control in multi-zone buildings: A comparative study with {ASHRAE G36} sequences,
\newblock Energy Build. 348 (2025) 116456.
\newblock \url{https://doi.org/10.1016/j.enbuild.2025.116456}.

\bibitem{ref:heidari2025_trustworthy_heating}
A.~Heidari, L.~Girardin, C.~Dorsaz, and F.~Mar{\'e}chal,
\newblock A trustworthy reinforcement learning framework for autonomous control of a large-scale complex heating system: Simulation and field implementation,
\newblock Appl. Energy 378 (2025) 124815.
\newblock \url{https://doi.org/10.1016/j.apenergy.2024.124815}.

\bibitem{ref:an2024_verifiable_hvac}
Z.~An, X.~Ding, and W.~Du,
\newblock Go beyond black-box policies: Rethinking the design of learning agent for interpretable and verifiable {HVAC} control,
\newblock in: Proceedings of the 61st {ACM/IEEE} Design Automation Conference ({DAC} 2024), 2024, Article 86, pp. 1--6.
\newblock \url{https://doi.org/10.1145/3649329.3656234}.

\bibitem{ref:nist_ai_optimized_controls}
National Institute of Standards and Technology,
\newblock {AI}-optimized building controls,
\newblock \url{https://www.nist.gov/programs-projects/ai-optimized-building-controls} (accessed 2026-07-24).

\bibitem{ref:doe_emcs_workforce_roundtable}
U.S. Department of Energy,
\newblock Energy management and control systems workforce development roundtable,
\newblock Building Technologies Office, April 2023,
\newblock \url{https://www.energy.gov/sites/default/files/2023-04/bto-emcs-workforce-roundtable-040423.pdf} (accessed 2026-07-24).

\bibitem{vaswani2017_attention}
A.~Vaswani, N.~Shazeer, N.~Parmar, J.~Uszkoreit, L.~Jones, A.~N. Gomez, et al.,
\newblock Attention is all you need,
\newblock in: Advances in Neural Information Processing Systems 30 (NeurIPS 2017), 2017, pp. 5998--6008.

\bibitem{brown2020_language_models}
T.~B. Brown, B.~Mann, N.~Ryder, M.~Subbiah, J.~D. Kaplan, P.~Dhariwal, et al.,
\newblock Language models are few-shot learners,
\newblock in: Advances in Neural Information Processing Systems 33 (NeurIPS 2020), 2020, pp. 1877--1901.

\bibitem{ref:deepseekr1_2025}
D.~Guo, D.~Yang, H.~Zhang, J.~Song, P.~Wang, Q.~Zhu, et al.,
\newblock {DeepSeek-R1} incentivizes reasoning in {LLMs} through reinforcement learning,
\newblock Nature 645 (2025) 633--8.
\newblock \url{https://doi.org/10.1038/s41586-025-09422-z}.

\bibitem{ref:zhang2024_llm_interpretable_control}
L.~Zhang and Z.~Chen,
\newblock Large language model-based interpretable machine learning control in building energy systems,
\newblock Energy Build. 313 (2024) 114278.
\newblock \url{https://doi.org/10.1016/j.enbuild.2024.114278}.

\bibitem{deepseekmath2024}
Z.~Shao, P.~Wang, Q.~Zhu, R.~Xu, J.~Song, X.~Bi, et al.,
\newblock {DeepSeekMath}: Pushing the limits of mathematical reasoning in open language models,
\newblock arXiv preprint arXiv:2402.03300 (2024).
\newblock \url{https://doi.org/10.48550/arXiv.2402.03300}.

\bibitem{ref:mirshekali2025_llm_energy_systems_review}
H.~Mirshekali, M.~R. Shadi, F.~Ghanadi~Ladani, and H.~R. Shaker,
\newblock A review of large language models for energy systems: Applications, challenges, and future prospects,
\newblock IEEE Access 13 (2025) 163162--88.
\newblock \url{https://doi.org/10.1109/ACCESS.2025.3610994}.

\bibitem{qaisar2026_llm_occupancy}
I.~Qaisar, K.~Sun, and Q.~Zhao,
\newblock Exploring large language models for indoor occupancy measurement in smart office buildings,
\newblock Build. Environ. 287 (2026) 113860.
\newblock \url{https://doi.org/10.1016/j.buildenv.2025.113860}.

\bibitem{qaisar2025_occupancy_centric_control}
I.~Qaisar, W.~Liang, K.~Sun, T.~Xing, and Q.~Zhao,
\newblock An experimental comparative study of energy saving based on occupancy-centric control in smart buildings,
\newblock Build. Environ. 268 (2025) 112322.
\newblock \url{https://doi.org/10.1016/j.buildenv.2024.112322}.

\bibitem{qaisar2026_occmamba}
I.~Qaisar, K.~Sun, D.~Zhong, X.~Yang, X.~Miao, and Q.~Zhao,
\newblock {OCC-Mamba}: A {Mamba}-based deep learning approach for indoor occupancy prediction,
\newblock Build. Environ. 289 (2026) 114085.
\newblock \url{https://doi.org/10.1016/j.buildenv.2025.114085}.

\bibitem{ref:song2023_pretrained_llm_industrial_control}
L.~Song, C.~Zhang, L.~Zhao, and J.~Bian,
\newblock Pre-trained large language models for industrial control,
\newblock arXiv preprint arXiv:2308.03028 (2023).
\newblock \url{https://doi.org/10.48550/arXiv.2308.03028}.

\bibitem{ref:ahn2023_chatgpt_hvac}
K.~U. Ahn, D.-W. Kim, H.~M. Cho, and C.-U. Chae,
\newblock Alternative approaches to {HVAC} control of Chat Generative Pre-Trained Transformer ({ChatGPT}) for autonomous building system operations,
\newblock Buildings 13(11) (2023) 2680.
\newblock \url{https://doi.org/10.3390/buildings13112680}.

\bibitem{ref:sawada2025_agentic_ai_dce}
T.~Sawada, M.~Mizuno, T.~Hasegawa, K.~Yokoyama, and M.~Kono,
\newblock Office-in-the-loop: An investigation into agentic {AI} for advanced building {HVAC} control systems,
\newblock Data-Centric Eng. 6 (2025) e31.
\newblock \url{https://doi.org/10.1017/dce.2025.10010}.

\bibitem{ref:bhatt2026_thermollm}
K.~Bhatt, X.~Lin, M.~Amos, F.~D. Salim, and W.~Hu,
\newblock {ThermoLLM}: Thermodynamics-aware {HVAC} control with spatial-semantic knowledge graph,
\newblock arXiv preprint arXiv:2606.22911 (2026).
\newblock \url{https://doi.org/10.48550/arXiv.2606.22911}.

\bibitem{ref:ko2025_darlin}
Y.-D. Ko and R.~K. Jain,
\newblock {DARLIN}: Domain-guided augmented retrieval for {LLM}-based interpretable {HVAC} control,
\newblock in: Proceedings of the 12th ACM International Conference on Systems for Energy-Efficient Buildings, Cities, and Transportation (BuildSys '25), 2025, pp. 440--3.
\newblock \url{https://doi.org/10.1145/3736425.3772322}.

\bibitem{ref:zhong2026_hierarchical_llm_rl}
D.~Zhong, T.~Xing, K.~Sun, X.~Yang, H.~Huang, I.~Qaisar, et al.,
\newblock Hierarchical control framework integrating {LLMs} with {RL} for decarbonized {HVAC} operation,
\newblock arXiv preprint arXiv:2603.26050 (2026).
\newblock \url{https://doi.org/10.48550/arXiv.2603.26050}.

\bibitem{liu2025r1zero}
Z.~Liu, C.~Chen, W.~Li, P.~Qi, T.~Pang, C.~Du, et al.,
\newblock Understanding {R1-Zero}-like training: A critical perspective,
\newblock arXiv preprint arXiv:2503.20783 (2025).
\newblock \url{https://doi.org/10.48550/arXiv.2503.20783}.

\bibitem{shioda2026_tes_rft}
T.~Shioda, K.~Terashima, and T.~Nagai,
\newblock Verifier-based reinforcement fine-tuning of reasoning models for thermal energy storage control,
\newblock arXiv preprint arXiv:2607.12856 (2026).
\newblock \url{https://doi.org/10.48550/arXiv.2607.12856}.

\bibitem{mroueh2025rlvr}
Y.~Mroueh,
\newblock Reinforcement learning with verifiable rewards: {GRPO}'s effective loss, dynamics, and success amplification,
\newblock arXiv preprint arXiv:2503.06639 (2025).
\newblock \url{https://doi.org/10.48550/arXiv.2503.06639}.

\bibitem{ichter2023_saycan}
B.~Ichter, A.~Brohan, Y.~Chebotar, C.~Finn, K.~Hausman, A.~Herzog, et al.,
\newblock Do as I can, not as I say: Grounding language in robotic affordances,
\newblock in: Proceedings of the 6th Conference on Robot Learning, PMLR 205, 2023, pp. 287--318.


\bibitem{lora2022}
E.~J. Hu, Y.~Shen, P.~Wallis, Z.~Allen-Zhu, Y.~Li, S.~Wang, et al.,
\newblock {LoRA}: Low-rank adaptation of large language models,
\newblock in: International Conference on Learning Representations (ICLR 2022), 2022.

\bibitem{ref:fujimoto2019_bcq}
S.~Fujimoto, D.~Meger, and D.~Precup,
\newblock Off-policy deep reinforcement learning without exploration,
\newblock in: Proceedings of the 36th International Conference on Machine Learning (ICML 2019), PMLR 97, 2019, pp. 2052--62.

\bibitem{ref:kumar2020_cql}
A.~Kumar, A.~Zhou, G.~Tucker, and S.~Levine,
\newblock Conservative Q-learning for offline reinforcement learning,
\newblock in: Advances in Neural Information Processing Systems 33 (NeurIPS 2020), 2020, pp. 1179--91.

\bibitem{tesauro1997_rollout}
G.~Tesauro and G.~R. Galperin,
\newblock On-line policy improvement using Monte-Carlo search,
\newblock in: Advances in Neural Information Processing Systems 9 (NIPS 1996), 1997, pp. 1068--74.

\bibitem{sutton2018rl}
R.~S. Sutton and A.~G. Barto,
\newblock Reinforcement Learning: An Introduction,
\newblock second ed., MIT Press, Cambridge, MA, 2018.

\bibitem{mhlw_building_environment_criteria}
Ministry of Health, Labour and Welfare, Japan,
\newblock Building environmental hygiene management standards (in Japanese),
\newblock \url{https://www.mhlw.go.jp/bunya/kenkou/seikatsu-eisei10/index.html} (accessed 2026-07-23).

\bibitem{miyata2023_phyvac}
S.~Miyata, S.~Shi, Y.~Akashi, and T.~Sawachi,
\newblock Phyvac: A {Python} module for highly flexible {HVAC} system simulation, and fault dataset generation as an application example,
\newblock in: Proceedings of the 18th IBPSA Conference (Building Simulation 2023), Shanghai, China, 2023, pp. 915--22.
\newblock \url{https://doi.org/10.26868/25222708.2023.1370}.

\bibitem{jma_tokyo_10min_2025}
Japan Meteorological Agency,
\newblock Past Weather Data Search: 10-minute observations at Tokyo station 47662, 28 July--1 August 2025,
\newblock \url{https://www.data.jma.go.jp/stats/etrn/} (accessed 2026-07-18).

\bibitem{ref:fujimoto2018_td3}
S.~Fujimoto, H.~van Hoof, and D.~Meger,
\newblock Addressing function approximation error in actor-critic methods,
\newblock in: Proceedings of the 35th International Conference on Machine Learning (ICML 2018), PMLR 80, 2018, pp. 1587--96.

\bibitem{ref:gpt5_system_card}
{OpenAI},
\newblock {GPT-5} system card,
\newblock 2025, \url{https://openai.com/index/gpt-5-system-card/} (accessed 2026-07-30).

\bibitem{ref:openai2025_gptoss}
{OpenAI},
\newblock gpt-oss-120b \& gpt-oss-20b model card,
\newblock arXiv preprint arXiv:2508.10925 (2025).
\newblock \url{https://doi.org/10.48550/arXiv.2508.10925}.

\bibitem{chen2025_reasoning_models_faithfulness}
Y.~Chen, J.~Benton, A.~Radhakrishnan, J.~Uesato, C.~Denison, J.~Schulman, et al.,
\newblock Reasoning models don't always say what they think,
\newblock arXiv preprint arXiv:2505.05410 (2025).
\newblock \url{https://doi.org/10.48550/arXiv.2505.05410}.

\bibitem{hao2023_rap}
S.~Hao, Y.~Gu, H.~Ma, J.~Hong, Z.~Wang, D.~Wang, et al.,
\newblock Reasoning with language model is planning with world model,
\newblock in: Proceedings of the 2023 Conference on Empirical Methods in Natural Language Processing, 2023, pp. 8154--73.
\newblock \url{https://doi.org/10.18653/v1/2023.emnlp-main.507}.

\bibitem{wang2025_safe_building_rl}
X.~Wang, P.~Wang, R.~Huang, X.~Zhu, J.~Arroyo, and N.~Li,
\newblock Safe deep reinforcement learning for building energy management,
\newblock Applied Energy 377 (2025) 124328.
\newblock \url{https://doi.org/10.1016/j.apenergy.2024.124328}.

\bibitem{manakul2023_selfcheckgpt}
P.~Manakul, A.~Liusie, and M.~J.~F.~Gales,
\newblock {SelfCheckGPT}: Zero-resource black-box hallucination detection for generative large language models,
\newblock in: Proceedings of the 2023 Conference on Empirical Methods in Natural Language Processing, 2023, pp. 9004--9017.
\newblock \url{https://doi.org/10.18653/v1/2023.emnlp-main.557}.

\bibitem{silver2014dpg}
D.~Silver, G.~Lever, N.~Heess, T.~Degris, D.~Wierstra, and M.~Riedmiller,
\newblock Deterministic policy gradient algorithms,
\newblock in: Proceedings of the 31st International Conference on Machine Learning (ICML 2014), PMLR 32, 2014, pp. 387--95.

\bibitem{hafner2020dreamer}
D.~Hafner, T.~Lillicrap, J.~Ba, and M.~Norouzi,
\newblock Dream to control: Learning behaviors by latent imagination,
\newblock in: International Conference on Learning Representations (ICLR 2020), 2020.

\bibitem{wu2023_daydreamer}
P.~Wu, A.~Escontrela, D.~Hafner, P.~Abbeel, and K.~Goldberg,
\newblock {DayDreamer}: World models for physical robot learning,
\newblock in: Proceedings of the 6th Conference on Robot Learning (CoRL 2022), PMLR 205, 2023, pp. 2226--40.

\bibitem{xiang2023_language_world_models}
J.~Xiang, T.~Tao, Y.~Gu, T.~Shu, Z.~Wang, Z.~Yang, et al.,
\newblock Language models meet world models: Embodied experiences enhance language models,
\newblock in: Advances in Neural Information Processing Systems 36 (NeurIPS 2023), 2023, pp. 75392--412.
\newblock \url{https://doi.org/10.52202/075280-3295}.

\bibitem{xie2025_llm_world_model}
K.~Xie, I.~Yang, J.~Gunerli, and M.~Riedl,
\newblock Making large language models into world models with precondition and effect knowledge,
\newblock in: Proceedings of the 31st International Conference on Computational Linguistics, 2025, pp. 7532--45.

\bibitem{hse_sbs_hsg132}
Health and Safety Executive,
\newblock How to deal with sick building syndrome (SBS), HSG132, paragraph 48,
\newblock \url{https://www.hse.gov.uk/pubns/priced/hsg132.pdf} (accessed 2026-09-24).

\end{thebibliography}
\end{document}